\documentclass[10pt,twocolumn,letterpaper]{article}

\usepackage[pagenumbers]{iccv} 

\usepackage{epsfig}
\usepackage{graphicx}
\usepackage{amsmath}
\usepackage{amssymb}

\usepackage[x11names, table]{xcolor}

\usepackage{stmaryrd}

\usepackage{pgfplots}
\pgfplotsset{compat=1.17}

\usepackage{colortbl}
\usepackage{booktabs}

\usepackage{multirow}
\usepackage{rotating}
\usepackage{array}
\newcolumntype{S}{>{\centering\arraybackslash}m{1.5cm}}
\newcolumntype{L}{>{\centering\arraybackslash}m{5cm}}
\usepackage{rotating, 
            makecell}
\usepackage{tabularray}
\usepackage{adjustbox}

\usepackage[T1]{fontenc}

\usepackage{lipsum}

\usepackage{nccmath}

\usepackage{mathtools}

\usepackage{siunitx}

\usepackage{amsthm}
\usepackage{tcolorbox}
\makeatletter
\DeclareRobustCommand\onedot{\futurelet\@let@token\@onedot}
\def\@onedot{\ifx\@let@token.\else.\null\fi\xspace}

\def\eg{\emph{e.g}\onedot} 
\def\ie{\emph{i.e}\onedot} 
\def\cf{\emph{cf}\onedot} \def\Cf{\emph{Cf}\onedot}
\def\etc{\emph{etc}\onedot} 
 
\def\etal{\emph{et al}\onedot}
\makeatother

\newcommand{\papertitle}{Discontinuity-aware Normal Integration for Generic Central Camera Models\xspace}

\definecolor{abl_color_1}{rgb}{0.1216, 0.4667, 0.7059} 
\definecolor{abl_color_2}{rgb}{0.6824, 0.7804, 0.9098} 
\definecolor{abl_color_3}{rgb}{0.0902, 0.7451, 0.8118} 
\definecolor{abl_color_4}{rgb}{0.1725, 0.6275, 0.1725} 
\definecolor{abl_color_5}{rgb}{0.7373, 0.8412, 0.1333} 
\definecolor{abl_color_6}{rgb}{0.92, 0.92, 0.0} 
\definecolor{abl_color_7}{rgb}{1.0000, 0.7333, 0.4706} 
\definecolor{abl_color_8}{rgb}{1.0000, 0.4980, 0.0549} 
\definecolor{abl_color_9}{rgb}{0.8902, 0.4667, 0.7608} 
\definecolor{abl_color_10}{rgb}{0.8392, 0.1529, 0.1569} 
\definecolor{abl_color_11}{rgb}{0.5804, 0.4039, 0.7412} 
\definecolor{abl_color_12}{rgb}{0.5490, 0.3373, 0.2941} 
\definecolor{abl_color_13}{rgb}{0.4980, 0.4980, 0.4980} 

\newtcolorbox{propositionbox}[2]{colback=gray!10, colframe=gray!50, 
  fonttitle=\bfseries, title=Proposition~#1:~#2}

\newenvironment{argument}[1][Argument]{\par\noindent\textit{#1.} }{\par}

\newlength{\savearrayrulewidth} 

\definecolor{iccvblue}{rgb}{0.21,0.49,0.74}
\usepackage[pagebackref,breaklinks,colorlinks,allcolors=iccvblue]{hyperref}

\def\paperID{11244} 
\def\confName{ICCV}
\def\confYear{2025}

\title{\papertitle}

\author{
\begin{tabular*}{\textwidth}{@{\extracolsep{\fill}}lcccr}
Francesco Milano\textsuperscript{1$\star$} & Manuel López-Antequera\textsuperscript{2} & Naina Dhingra\textsuperscript{2} & Roland Siegwart\textsuperscript{1} & Robert Thiel\textsuperscript{2}
\end{tabular*}
\\[5pt]${}^1$ETH Zurich, ${}^2$Meta
}
\begin{document}
\maketitle
\iftoggle{iccvfinal}{%
\renewcommand{\thefootnote}{\fnsymbol{footnote}}
\footnotetext{$^\star$Work mainly performed during an internship at Meta.}
\renewcommand{\thefootnote}{\arabic{footnote}} 
}{%
}

\label{sec:abstract}
\begin{abstract}
Recovering a 3D surface from its surface normal map, a problem known as normal integration, is a
key component for
photometric shape reconstruction techniques such as shape-from-shading and photometric stereo.
The
vast
majority of existing approaches for normal
integration
handle only implicitly the presence of depth discontinuities and are limited to orthographic or ideal pinhole
cameras.
In this paper, we propose a novel formulation
that
allows
modeling discontinuities
explicitly
and handling generic central 
cameras.
Our
key idea is
based on a local planarity assumption,
that we model
through constraints between surface normals and ray directions.
Compared to existing methods,
our
approach
more
accurately
approximates
the relation between depth and surface
normals, achieves state-of-the-art results on the standard
normal integration benchmark, and is the first to
directly handle
generic central camera models.
\end{abstract}
\section{Introduction\label{sec:introduction}}
The problem of reconstructing a 3D surface from its surface normal map, also known as \emph{normal integration},
has long been studied in computer vision.
Its importance lies in its several applications for shape reconstruction,
in particular
as a necessary step
to recover the surface from the output of
photometric stereo~\cite{Woodham1980PhotometricStereo} or
shape-from-shading~\cite{Horn1975ShapeFromShading} techniques, which
estimate
normals from
image shading.

Classically, normal integration has been studied predominantly under the assumption that the surface to be reconstructed is smooth~\cite{Queau2018NormalIntegrationSurvey}.
This assumption, however,
breaks in the presence of depth discontinuities, which naturally arise due to
occlusions.
While a number of methods for discontinuity-preserving integration have been proposed,
these
tend to
introduce
simplifying assumptions on the statistics of the discontinuities~\cite{Badri2014RobustSurfaceReconstruction, Queau2018VariationMethodsNormalIntegration}
or model
their magnitude only implicitly~\cite{Agrawal2005AnAlgebraicApproach, Cao2022BiNI, Wu2006VisibleSurfaceReconstruction}.
Moreover,
the vast majority of the
existing methods for discontinuity-preserving normal integration
tackle the case of
orthographic
projection~\cite{Badri2014RobustSurfaceReconstruction, Queau2018VariationMethodsNormalIntegration, Xie2019ARobustDGPBased}; an 
exception is represented by
the recently proposed methods of BiNI~\cite{Cao2022BiNI} and of Kim~\etal~\cite{Kim2024DiscontinuityPreserving}, which allow handling
normals
observed by an
ideal pinhole camera,
which
more closely resembles
real-world scenarios.

All the leading
methods
are
derived from
partial differential equations
(PDEs) which
relate normals to the depth map describing the surface, and typically
base
their formulation
on
functionals that discretely approximate these PDEs~\cite{Horn1986Poisson, Queau2018NormalIntegrationSurvey, Cao2022BiNI, Kim2024DiscontinuityPreserving}. 
In this work, we propose a novel formulation
not derived from differential constraints, but
based
instead
on the simple assumption that the surface is composed of local planes
separated by discontinuities. We 
model this assumption through
conditions between the surface normal and
the \emph{ray direction} associated
with
each
pixel. We show experimentally that this
results in a more accurate approximation of
the ground-truth relation between depth and normals.
Additionally,
by relying on ray directions,
our approach is to the best of our knowledge the first
to
directly handle
generic central camera models, thereby extending the case of an ideal pinhole.
Furthermore, 
our mathematical formulation explicitly takes discontinuities into
account.

In order to recover both the depth map and the discontinuity values,
we
adopt an iterative optimization
process
based on
the
bilateral
weighting scheme
of
BiNI. In particular, we adapt their semi-smooth assumption to our
formulation
and extend its optimization scheme to iteratively estimate depth and discontinuities.
We additionally provide important novel insights on
the optimization convergence,
in light of our formulation.
Experimental results show that our method
captures discontinuities more accurately than existing methods and
sets a new
state of the art
in
the standard normal integration benchmark~\cite{Shi2016DiLiGenT}.
We
provide extensive ablations on the hyperparameters of our method and
further demonstrate it
on normal maps from non-ideal pinhole cameras
and real-world
data, showing effective surface reconstruction also under these conditions.

In summary, our main contribution is a novel formulation for discontinuity-aware normal integration based on a local planarity assumption and ray directions, that: \textit{(i)} more accurately describes the relation between depth and surface normals, \textit{(ii)} achieves state-of-the-art results on the standard normal integration benchmark, and \textit{(iii)} shows for the first time direct applicability to generic central camera models.
\section{Related work\label{sec:related_work}}
In the following Section, we briefly review
the main
existing approaches for normal
integration.
For a more extensive summary, we refer the reader to the surveys~\cite{Queau2018NormalIntegrationSurvey, Queau2018VariationMethodsNormalIntegration}.

The majority of normal integration methods proposed in the literature are derived from discrete approximations to PDEs relating depth and surface normals. One category of approaches, pioneered by Horn and Brooks~\cite{Horn1986Poisson}, 
are based on
constraints between the partial derivatives of the depth and the gradient field computed from the normal map~\cite{Badri2014RobustSurfaceReconstruction, Durou2009IntegratingNormalFieldSurface, Harker2008LeastSquareSurfaceReconstruction, Agrawal2006WhatIsTheRange, Queau2018VariationMethodsNormalIntegration}. More recently, an alternative differential formulation has been 
proposed by Zhu and Smith~\cite{Zhu2020LeastSquaresSurfaceReconstruction} and later extended by Cao~\etal~\cite{Cao2022BiNI}
that instead enforces an orthogonality constraint between the normals and the tangent plane to the surface, showing improved numerical stability. Our
method
is derived from
a similar orthogonality constraint, but
proposes
a more general formulation
that is applicable to generic central camera models
and
explicitly takes discontinuities into account.

To handle depth discontinuities, two main categories of approaches have been proposed that extend the PDE-based formulations above.
One category
of methods modify their functionals with robust estimators that reduce the effect of large residuals~\cite{Durou2009IntegratingNormalFieldSurface, Badri2014RobustSurfaceReconstruction, Queau2015EdgePreservingIntegrationNormalField}.
Another line of
approaches instead introduce weights in the terms of the PDEs. Among
these, single-analysis methods
use weights defined before the optimization based on error residuals or input gradients~\cite{Karacali2003ReconstructingDiscontinuous, Agrawal2005AnAlgebraicApproach, Wang2012DetectingDiscontinuitiesSurfaceReconstruction, Xie2019ARobustDGPBased, Fraile2006CombinatorialSurfaceIntegration}. Since the weights are kept fixed, these approaches might fail to correct wrong discontinuities during the optimization. To address this issue, alternative approaches have been proposed that iteratively 
optimize
the weights.
Typically, this is achieved
by
alternatively updating
depth and 
parameters controlling the location of the 
discontinuities~\cite{Agrawal2006WhatIsTheRange, Wu2006VisibleSurfaceReconstruction, Queau2018VariationMethodsNormalIntegration}.
Recently, Cao~\etal~\cite{Cao2022BiNI} significantly advanced the state of the art by proposing an iterative weight-update approach based on the assumption that the target surface is one-sided differentiable. At each iteration, the terms in its functional are scaled by relatively
weighting
the
residuals on the two sides of each point, resulting in effective discontinuity preservation for the first time also for the perspective, ideal pinhole case.
Kim~\etal~\cite{Kim2024DiscontinuityPreserving}
later proposed
to explicitly model
gradients across
discontinuities through auxiliary
edges,
showing more accurate detection of small discontinuities.
In our
approach,
we adopt the bilateral weighting scheme of~\cite{Cao2022BiNI} and extend it to our formulation, which
explicitly models discontinuities and handles generic central cameras.
\section{Discontinuity-aware normal integration\label{sec:method}}
Formally, the objective of normal integration is to recover
a surface, in the form of a depth map, from a single-view
per-pixel normal map and
known camera parameters.
Our method tackles this problem by
explicitly
modeling surface discontinuities while solving for the unknown depth values.
Additionally,
unlike previous methods
that are designed
for
orthographic and pinhole cameras,
our approach
allows
modeling the
broader
category of central
cameras.

In the following Section, we first derive the general formulation of our method for discontinuity-aware surface normal integration
for
arbitrary
central cameras
(\cref{sec:method_general_formulation}).
We then
describe the general
optimization
framework
to
estimate
solutions from our proposed formulation (\cref{sec:method_solving_for_depth_and_discontinuities}).
Finally, in \Cref{sec:method_bilateral_formulation} we 
provide
specific
details on
how we perform
the optimization
and
retrieve discontinuities
by
extending
and generalizing
the bilateral
assumption
of~\cite{Cao2022BiNI}.

\subsection{Proposed formulation\label{sec:method_general_formulation}}
Let us consider a generic central camera, that is, any camera that models a 
\emph{central projection}~\cite{Hartley2004MultipleViewGeometry}, and let
us denote with \hbox{$\boldsymbol{\tau}: \boldsymbol{u}\in\mathbb{R}^2 \mapsto \boldsymbol{\tau}(\boldsymbol{u})=\left(\tau_x(\boldsymbol{u}), \tau_y(\boldsymbol{u}), 1\right)^\mathsf{T}\in\mathbb{R}^3$} the mapping from a point $\mathbf{u}=\left(u, v\right)^\mathsf{T}$ on its image plane to its corresponding \emph{ray direction vector} $\boldsymbol{\tau}(\boldsymbol{u})$. The elements $\tau_x(\boldsymbol{u})$ and $\tau_y(\boldsymbol{u})$ represent the tangent of the viewing angle, corresponding to the ray passing through $\boldsymbol{u}$, respectively along the $x$ and $y$ axes of the camera coordinate frame. For a generic point $\boldsymbol{p}(\boldsymbol{u})$ along the ray, with camera coordinates $\left(x(\boldsymbol{u}), y(\boldsymbol{u}), z(\boldsymbol{u})\right)^\mathsf{T}\in\mathbb{R}^3$, these can be expressed as $\tau_x(\boldsymbol{u})=\frac{x(\boldsymbol{u})}{z(\boldsymbol{u})}$ and $\tau_y(\boldsymbol{u})=\frac{y(\boldsymbol{u})}{z(\boldsymbol{u})}$. In the specific case of a pinhole camera with focal lengths $f_x$ and $f_y$ and principal point $(c_x, c_y)$, the mapping $\boldsymbol{\tau}$ is affine in the image coordinates and can be written as $\boldsymbol{\tau}(\boldsymbol{u})=\left(\frac{u-c_x}{f_x}, \frac{v-c_y}{f_y}, 1\right)^\mathsf{T}$.

When the camera observes a
fully-opaque
surface, each ray that intersects the surface is in one-to-one correspondence both with the visible
3D
point \hbox{$\boldsymbol{p} = \left(x, y, z\right)^\mathsf{T}\in\mathbb{R}^3$} at which it intersects the surface and with the
normal vector \hbox{$\boldsymbol{n}(\boldsymbol{p})=\left(n_x, n_y, n_z\right)^\mathsf{T}\in\mathcal{S}^2\subset\mathbb{R}^3$}
at that point.
It follows that it is possible to define injective mappings
from image coordinates to visible surface points and 
normal vectors.

\begin{figure}[!t]
    \centering
    \includegraphics[height=0.55\linewidth]{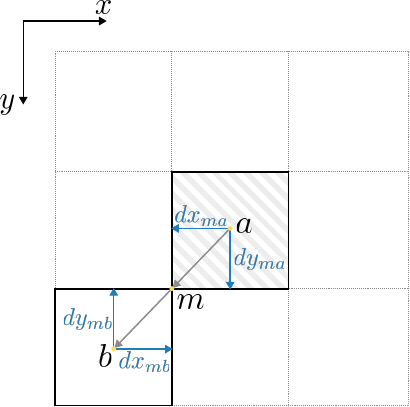}
    \caption{\textbf{Visualization of our local planarity assumption on the image plane.} For each pair of neighboring pixels $a$ and $b$, a subpixel $m$ is selected on the line segment connecting $a$ and $b$, here chosen to be equidistant from the pixel centers. Along both the directions $m\rightarrow b$ and $m\rightarrow a$, the surface is assumed to be locally planar, with a discontinuity at the location of $m$.}
\label{fig:surface_normal_integration_scheme_2d}
\vspace{-10pt}
\end{figure}

\begin{figure}[!t]
    \centering
    \includegraphics[height=0.55\linewidth]{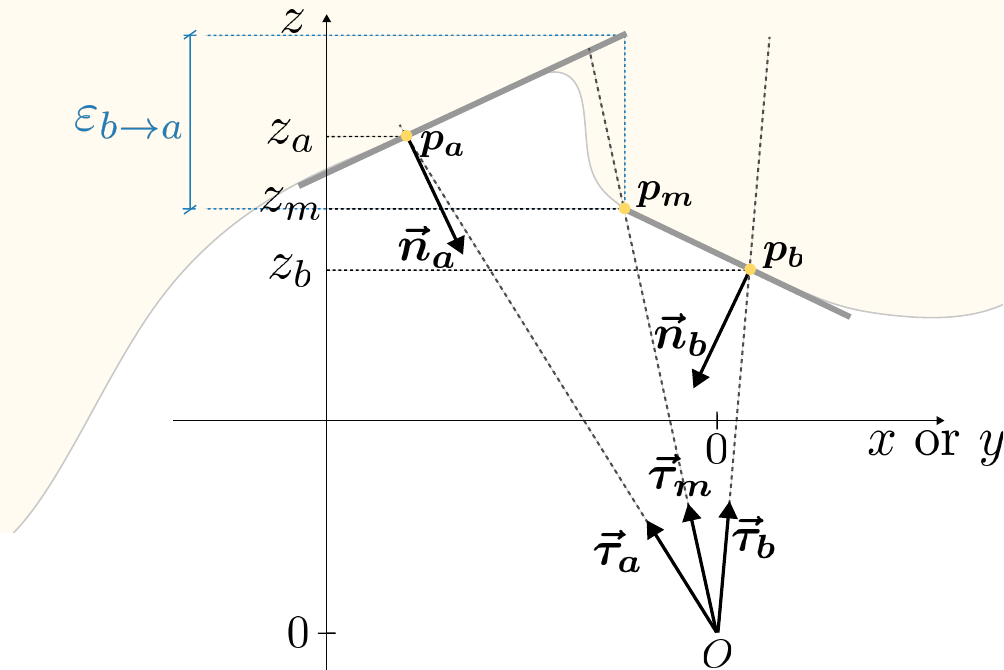}
    \caption{\textbf{Visualization of our local planarity assumption in 3D.} We assume that the surface can be modeled as piecewise-planar and define the plane endpoints as the intersection
    between the surface and the
    rays
    from the camera center of projection $O$ along the ray directions
    $\boldsymbol{\tau_a}$, $\boldsymbol{\tau_b}$, and $\boldsymbol{\tau_m}$.
    We explicitly model a discontinuity $\varepsilon_{b\rightarrow a}$ along the $z$ camera axis in
    line
    with
    $\boldsymbol{p_m}$.}
    \label{fig:surface_normal_integration_scheme_3d}
    \vspace{-10pt}
\end{figure}

Our general formulation for
normal integration,
makes use of:
(i) a local planarity approximation
to handle
pixel discretization,
(ii)
explicit discontinuity modelling, and
(iii)
the general definition of ray direction vectors.
In particular, let $a$ and $b$ be two neighboring pixels in the input normal map, with corresponding image coordinates $\boldsymbol{u_a}=\left(u_a, v_a\right)^\mathsf{T}$ and $\boldsymbol{u_b}=\left(u_b, v_b\right)^\mathsf{T}$. In our
main
experiments, we define neighborhood based on
$4-$connectivity,
although other connectivities
can
also be considered.
Furthermore, let $m$ denote a subpixel location along the line segment connecting pixel $a$ and $b$ on the image plane (\cref{fig:surface_normal_integration_scheme_2d}) and let $\boldsymbol{\tau_i}=\left(\tau_{x_i}, \tau_{y_i}, 1\right)^\mathsf{T}$, $\boldsymbol{p_i}=\left(x_i, y_i, z_i\right)^\mathsf{T}$, and $\boldsymbol{n_i}=\left(n_{ix}, n_{iy}, n_{iz}\right)^\mathsf{T}$ denote respectively the ray direction vector, the unknown visible surface point, and the known normal vector corresponding to (sub)pixel $i\in\{a, b, m\}$.
Our method assumes that
at the location of
both
$a$ and
$b$
the surface can be locally approximated by a plane segment perpendicular to the normal vector. More precisely, as illustrated in \cref{fig:surface_normal_integration_scheme_3d}, we assume that
the point $\boldsymbol{p_m}$ can be found at the intersection between the ray $\boldsymbol{\tau_m}$ and the plane tangent to the surface at $\boldsymbol{p_b}$.
To model a depth discontinuity at $\boldsymbol{p_m}$,
we
further assume that
the
plane tangent
to the surface
at point $\boldsymbol{p_a}$
can be intersected by
moving from $\boldsymbol{p_m}$
by
$\varepsilon_{b\rightarrow a}$
units
along the
positive
direction of the $z$ camera axis.

The assumptions described above can be modeled through the following system of $6$ independent equations, where we define
$dx_{ij} := x_i - x_j$, $dy_{ij} := y_i - y_j$, $dz_{ij} := z_i - z_j$, with $i, j\in\{a, b, m\}$, and use a right-hand convention for camera coordinates ($x$, $y$, and $z$ axes pointing respectively to the right, bottom, and front):
\begin{equation}
{
    \begin{cases}
    \tau_{x_m} = \frac{x_b+dx_{mb}}{z_b+dz_{mb}}\\[1ex]
    \tau_{y_m} = \frac{y_b+dy_{mb}}{z_b+dz_{mb}}\\[1ex]
    \tau_{x_a} = \frac{x_b+dx_{mb}-dx_{ma}}{z_b+dz_{mb}-dz_{ma}}\\[1ex]
    \tau_{y_a} = \frac{y_b+dy_{mb}-dy_{ma}}{z_b+dz_{mb}-dz_{ma}}\\[1ex]
    \medmath{n_{bx}\cdot dx_{mb}+n_{by}\cdot dy_{mb}+n_{bz}\cdot dz_{mb}=0}\\[1ex]
    \medmath{n_{ax}\cdot dx_{ma}+n_{ay}\cdot dy_{ma}+n_{az}\cdot (dz_{ma}+\varepsilon_{b\rightarrow a})=0}\\[1ex]
    \end{cases}
}
    \label{eq:system_equations_ours}
\end{equation}

The first four equations in the system follow from the definition of ray direction vector, while the last two model the perpendicularity constraint between the two plane segments and the normal vectors $\boldsymbol{n_a}$, $\boldsymbol{n_b}$, taking into account the depth discontinuity $\varepsilon_{b\rightarrow a}$.

Solving~\eqref{eq:system_equations_ours} for $dx_{ma}$, $dx_{mb}$, $dy_{ma}$, $dy_{mb}$, $dz_{ma}$, $dz_{mb}$, and plugging the solutions back into the definitions of these quantities yields the following
condition on
$z_a$, $z_b$, and $\varepsilon_{b\rightarrow a}$:
\begin{equation}
    z_a = \omega_{\varepsilon_a}\cdot\varepsilon_{b\rightarrow a} + \omega_{b\rightarrow a}\cdot z_b,
    \label{eq:ours_formulation}
\end{equation}
where
\begin{align}
    \begin{split}
        \omega_{\varepsilon_a}&=\frac{n_{a_z}}{\boldsymbol{n_a}^\mathsf{T} \boldsymbol{\tau_a}}\\
        \omega_{b\rightarrow a}&=\frac{\boldsymbol{n_a}^\mathsf{T}\boldsymbol{\tau_m}\cdot\boldsymbol{n_b}^\mathsf{T} \boldsymbol{\tau_b}}{\boldsymbol{n_a}^\mathsf{T} \boldsymbol{\tau_a}\cdot\boldsymbol{n_b}^\mathsf{T} \boldsymbol{\tau_m}}.
        \label{eq:ours_coefficient_definition}
    \end{split}
\end{align}
We provide a full derivation of~\eqref{eq:ours_coefficient_definition} in \cref{sec_suppl:derivation_our_formulation}.

It should be noted that all the quantities in~\eqref{eq:ours_coefficient_definition} are known by hypothesis, except for $\boldsymbol{\tau_m}$ (or equivalently the location of the subpixel $m$ on the image plane),
the choice of which controls the local planarity approximation (\cref{fig:surface_normal_integration_scheme_3d}).
In our main experiments, we assume for simplicity that
$\boldsymbol{\tau_m} = (\boldsymbol{\tau_a} + \boldsymbol{\tau_b}) / 2$, which for mappings $\boldsymbol{\tau}$ that are affine in the image coordinates corresponds to $m$ having image coordinates given by the average of the pixel coordinates of $a$ and $b$, \ie, $\boldsymbol{u_m} = (\boldsymbol{u_a} + \boldsymbol{u_b}) / 2$.
However,
alternative
choices for obtaining $\boldsymbol{\tau_m}$ are possible, including through linear interpolation $\boldsymbol{\tau_m} = \boldsymbol{\tau_a} + \lambda_m(\boldsymbol{\tau_b} - \boldsymbol{\tau_a})$, with $\lambda_m\in[0, 1]$ (of which the above is a special case, with $\lambda_m=0.5$ for all pixel pairs).
We refer the reader to \cref{sec_suppl:ablation_lambda_m} in the Supplementary Material for a more detailed analysis and for ablations on the choice of $\boldsymbol{\tau_m}$.

We furthermore note that the coefficients in~\eqref{eq:ours_coefficient_definition} depend on
terms of the form $\boldsymbol{n}^\mathsf{T}\boldsymbol{\tau}$, which relate surface normals to ray directions through a dot product. This dot product relationship has previously been studied in the
literature,
famously by Marr~\cite{Marr1977AnalysisOccludingContour} and more recently by Bae and Davison~\cite{Bae2024DSINE}. As previously noted
in these works,
a necessary condition for a surface point to be visible is that the angle between its corresponding ray direction vector and surface normal vector is greater than $90^\circ$, \ie, $\boldsymbol{n}^\mathsf{T}\boldsymbol{\tau} < 0$, with equality being attained in the limit of the point lying on an occluding boundary. It follows that, assuming valid surface normals, the terms $\boldsymbol{n_a}^\mathsf{T}\boldsymbol{\tau_a}$ and $\boldsymbol{n_b}^\mathsf{T}\boldsymbol{\tau_b}$ in~\eqref{eq:ours_coefficient_definition} are strictly 
negative. On the other hand,
the terms $\boldsymbol{n_a}^\mathsf{T}\boldsymbol{\tau_m}$ and $\boldsymbol{n_b}^\mathsf{T}\boldsymbol{\tau_m}$ are negative if the points of intersection between the ray direction $\boldsymbol{\tau_m}$ and the two local planes containing $\boldsymbol{p_a}$ and $\boldsymbol{p_b}$, respectively, are visible by the camera when approximating the surface as local planes.
As we discuss more in detail in
\cref{sec_suppl:analysis_positivity_log_term},
when choosing $\boldsymbol{\tau_m}$ to linearly interpolate $\boldsymbol{\tau_a}$ and $\boldsymbol{\tau_b}$ the latter condition is fulfilled for all but very specific corner cases, and is always verified in practice for $\boldsymbol{\tau_m} = (\boldsymbol{\tau_a} + \boldsymbol{\tau_b}) / 2$.
From~\eqref{eq:ours_coefficient_definition}, it follows that under these settings the $\omega_{b\rightarrow a}$ terms are always positive. While this condition is not strictly necessary, it allows a convenient reformulation of~\eqref{eq:ours_formulation}, as detailed in the next Section.

\subsection{General solution framework\label{sec:method_solving_for_depth_and_discontinuities}}

Similarly to previous methods~\cite{Cao2022BiNI, Kim2024DiscontinuityPreserving},
our formulation allows
estimating
the unknown depth values
by
solving
a
least-squares optimization
problem.
In particular, the set of conditions~\eqref{eq:ours_formulation} for all valid choices of
neighboring pixels
$(a, b)$, $(a, c)$, \etc
can be rewritten in the form of a system of linear equations $\mathbf{Az}=\mathbf{b}$, where
\begin{align}
    \begin{split}
    \mathbf{A} &= \begin{bmatrix}
    1 & -\omega_{b\rightarrow a} & 0 & \cdots\\
    1 & 0 & -\omega_{c\rightarrow a} & \cdots\\
    \vdots & \vdots & \vdots & \ddots\\
    -\omega_{a\rightarrow b} & 1 & 0 & \cdots\\
    \vdots & \vdots & \vdots & \ddots\\
    \end{bmatrix},\\
    \mathbf{z}&=\begin{bmatrix}
        z_a\\z_b\\z_c\\\vdots
    \end{bmatrix},\ \mathrm{and}\  \mathbf{b}=\begin{bmatrix}
        \omega_{\varepsilon_a}\cdot\varepsilon_{b\rightarrow a}\\
        \omega_{\varepsilon_a}\cdot\varepsilon_{c\rightarrow a}\\
        \vdots\\
        \omega_{\varepsilon_b}\cdot\varepsilon_{a\rightarrow b}\\
        \vdots
    \end{bmatrix}.
    \end{split}
    \label{eq:system_equations_optimization_terms}
\end{align}

The optimization problem then reads as:
\begin{equation}
    \min_\mathbf{z} \left(\mathbf{Az}-\mathbf{b}\right)^\mathsf{T}\mathbf{W}\left(\mathbf{Az}-\mathbf{b}\right),
    \label{eq:general_optimization_problem}
\end{equation}
where $\mathbf{W}$ is an optional diagonal matrix that can assign different weights to the equations. The unknown depth values $\mathbf{z}$
can then be found by applying an iterative conjugate gradient method~\cite{Hestenes1952ConjugateGradient, Cao2022BiNI} on the normal equation of~\eqref{eq:general_optimization_problem}, $\mathbf{A}^\mathsf{T}\mathbf{WAz} = \mathbf{A}^\mathsf{T}\mathbf{Wb}$.
However, since our formulation explicitly takes discontinuities into account, the term $\mathbf{b}$ in~\eqref{eq:general_optimization_problem} depends on the values $\varepsilon_{b\rightarrow a}$, $\varepsilon_{c\rightarrow a}$,
\etc.
We
note that if the ground-truth values of
these quantities
-- hence of the term $\mathbf{b}$ -- were known, the conditions expressed by the system $\mathbf{Az}=\mathbf{b}$ would model \emph{exactly} the 
relationship between the
ground-truth
depth values at the different pixels, and as we show in \Cref{sec:benchmark_experiments_pinhole}, the optimization would be able to recover the ground-truth depth values with
close-to-perfect
accuracy. Since, however, the ground-truth values for
$\varepsilon_{b\rightarrow a}$
are unknown,
in our optimization we
not only iteratively
update the depth values, but also
optimize the term $\mathbf{b}$,
so that it progressively models the ground-truth
term
more closely.

More in detail,
upon
initialization we assume the surface to be smooth everywhere, thereby setting all discontinuities $\varepsilon_{b\rightarrow a}$ to $0$. As a consequence, the system of equations~\eqref{eq:system_equations_optimization_terms} is initially homogeneous; knowing that
depth values are positive and following a common practice in the literature~\cite{Queau2018NormalIntegrationSurvey, Cao2022BiNI}, we
therefore
introduce the change of variable $\tilde{z} := \log{z}$. To allow rewriting~\eqref{eq:ours_formulation} as a condition on $\tilde{z}_a$ and $\tilde{z}_b$, we additionally express the discontinuity values as \emph{relative} discontinuities, by introducing
the terms
$\alpha_{b\rightarrow a} \coloneq \varepsilon_{b\rightarrow a}/z_b$, so that
\begin{equation}
    \varepsilon_{b\rightarrow a} = \alpha_{b\rightarrow a}\cdot z_b.
    \label{eq:alpha_a_from_b_definition}
\end{equation}
Using~\eqref{eq:alpha_a_from_b_definition} and applying the logarithm to both sides of~\eqref{eq:ours_formulation},
we can rewrite our condition~\eqref{eq:ours_formulation} as
\begin{equation}
\tilde{z}_a^{(t)} = \log\left(\omega_{\varepsilon_a}\cdot\alpha_{b\rightarrow a}^{(t)} + \omega_{b\rightarrow a}\right) + \tilde{z}_b^{(t)},
\label{eq:ours_formulation_log}
\end{equation}
where we additionally use the superscript ${}^{(t)}$ to indicate that the variables are evaluated at iteration $t$ of the optimization.

We note that
upon
initialization the terms inside the logarithm in~\eqref{eq:ours_formulation_log} coincide with $\omega_{b\rightarrow a}$, having set $\varepsilon_{b\rightarrow a}^{(0)} = 0$ and therefore $\alpha_{b\rightarrow a}^{(0)} = 0$.
As noted in~\Cref{sec:method}, when choosing the subpixel locations $m$ to interpolate between $a$ and $b$, the terms $\omega_{b\rightarrow a}$ are positive, which ensures that the logarithm is always defined.
Similarly to~\cite{Cao2022BiNI, Kim2024DiscontinuityPreserving}, we initialize the log-depth values
$\tilde{z}_a^{(t)}, \tilde{z}_b^{(t)}$, \etc to $0$,
which corresponds to a planar surface of unit depth. At each
iteration $t$,
we
first
optimize the log-depth values
using the system of equations $\mathbf{\tilde{A}\tilde{z}} = \mathbf{\tilde{b}}$ that can be derived from~\eqref{eq:ours_formulation_log}
with
the same procedure used to write~\eqref{eq:system_equations_optimization_terms} 
from~\eqref{eq:ours_formulation};
then, we
update the terms $\alpha_{b\rightarrow a}^{(t)}$, as we detail in the next Section.

\subsection{Discontinuity-aware bilateral formulation\label{sec:method_bilateral_formulation}}
In order to guide the optimization of the log-depth values as well as to iteratively update the discontinuity values, we adopt the
semi-smooth
assumption of BiNI~\cite{Cao2022BiNI}, which we extend to our formulation and briefly summarize below.

For a pinhole camera with focal
lengths $f_x$ and $f_y$
and principal point $(c_x, c_y)$, BiNI makes use of the following discrete PDE, here expressed in our notation:
\begin{equation}
    \gamma_{b\rightarrow a}(\tilde{z}_a - \tilde{z}_b) = \delta_{b\rightarrow a},
    \label{eq:bini_formulation}
\end{equation}
where neighboring pixels $b$ are defined according to $4$-connectivity, and the terms $\gamma_{b\rightarrow a}$ and  $\delta_{b\rightarrow a}$ are defined as:
\begin{align}
    \gamma_{b\rightarrow a} = n_{ax}(u_a - c_x) + n_{ay}(v_a-c_y)+n_{az}f,
    \label{eq:gamma_a_from_b_bini_formulation_1}
\end{align}
with $f = f_x,\ \delta_{b\rightarrow a} = \pm n_{ax}$ for neighboring pixels $b$ s.t. $(u_b, v_b) = (u_a \pm 1, v_a)$ and $f = f_y,\ \delta_{b\rightarrow a} = \pm n_{ay}$ for neighboring pixels $b$ s.t. $(u_b, v_b) = (u_a, v_a \pm 1)$.
Their method then assumes the surface to be semi-smooth, that is, to contain at most one-sided discontinuities. This assumption is modeled
by weighting
each equation~\eqref{eq:bini_formulation},
at each optimization iteration $t$,
by a term
\begin{equation}
    w_{b\rightarrow a}^{\mathrm{BiNI}(t)}=\sigma_k\left(\left(\mathrm{res}_{-b\rightarrow a}^{(t)}\right)^2 - \left(\mathrm{res}_{b\rightarrow a}^{(t)}\right)^2\right),
    \label{eq:bini_weight_definition}
\end{equation}
where
\hbox{$\mathrm{res}_{b\rightarrow a}^{(t)}\coloneq\gamma_{b\rightarrow a}\left(\tilde{z}_a^{(t)} - \tilde{z}_b^{(t)}\right)$}
is a residual that encodes the extent to which the surface is discontinuous between $\boldsymbol{p_b}$ and $\boldsymbol{p_a}$,
$\sigma_k(\cdot)$ is the sigmoid function $\sigma_k(x)=\left(1+e^{-kx}\right)^{-1}$,
and where we denote with $-b$ the neighbor of $a$ opposite to $b$, \ie s.t. \hbox{$\boldsymbol{u_{-b}} - \boldsymbol{u_a} = -\left(\boldsymbol{u_b} - \boldsymbol{u_a}\right)$}. By properties of the sigmoid function, $w_{b\rightarrow a}^{\mathrm{BiNI}(t)}\in[0, 1]$ and $w_{-b\rightarrow a}^{\mathrm{BiNI}(t)} = 1 - w_{b\rightarrow a}^{\mathrm{BiNI}(t)}$,
with
$w_{b\rightarrow a}^{\mathrm{BiNI}(t)}\approx 0$ 
indicating
that the estimated surface is discontinuous between $\boldsymbol{p_b}$ and $\boldsymbol{p_a}$ but continuous between $\boldsymbol{p_{-b}}$ and $\boldsymbol{p_a}$, 
and
$w_{b\rightarrow a}^{\mathrm{BiNI}(t)}\approx w_{-b\rightarrow a}^{\mathrm{BiNI}(t)}\approx 0.5$
that the surface is continuous on both sides.

We note that, up to the multiplicative constant $\gamma_{b\rightarrow a}$, the
formulation of BiNI~\eqref{eq:bini_formulation}
has the same functional form as our formulation. While~\eqref{eq:bini_formulation} could be rewritten as
$\tilde{z}_a - \tilde{z_b} = \delta_{b\rightarrow a}/\gamma_{b\rightarrow a}$, as noted in Sec.~2 of the Supplementary of BiNI~\cite{Cao2022BiNI_supplementary} the factor $\gamma_{b\rightarrow a}$ proves to be crucial to improving
their
numerical stability during optimization. We empirically verify that the same holds true
for
our formulation,
and
we
therefore
rewrite
our
formulation as follows,
by multiplying both sides of~\eqref{eq:ours_formulation_log} by $\gamma_{b\rightarrow a}$ and rearranging:
\begin{equation}
    \gamma_{b\rightarrow a}(\tilde{z}_a - \tilde{z}_b) = \gamma_{b\rightarrow a}\log\left(\omega_{b\rightarrow a}+\omega_{\varepsilon_a}\cdot\alpha_{b\rightarrow a}\right).
    \label{eq:ours_formulation_log_with_bini_factor}
\end{equation}
Following BiNI, we furthermore define our weighting matrix $\mathbf{W}$ based 
on~\eqref{eq:bini_weight_definition}.
Importantly,
we additionally note that
$\gamma_{b\rightarrow a}$ can be rewritten (up to the differences between $f_x$ and $f_y$)
as
\begin{equation}
    \gamma_{b\rightarrow a} = f\cdot\boldsymbol{n_a}^\mathsf{T} \boldsymbol{\tau_a},
    \label{eq:gamma_a_from_b_bini_formulation_2}
\end{equation}
which for generic central cameras we generalize
as
\begin{equation}
    \gamma_{b\rightarrow a} = {\left\|\boldsymbol{u_b}-\boldsymbol{u_a}\right\|} / {\left\|\boldsymbol{\tau_b}-\boldsymbol{\tau_a}\right\|}\cdot\boldsymbol{n_a}^\mathsf{T}\boldsymbol{\tau_a}.
    \label{eq:gamma_a_from_b_formulation_general_central_cameras}
\end{equation}
In light of this observation, we 
present
a thorough analysis of the impact of the terms in~\eqref{eq:gamma_a_from_b_formulation_general_central_cameras} in \Cref{sec_suppl:analysis_gamma_factor},
providing
important novel findings about their effect on convergence
and shedding light on the
role
of $\gamma_{b\rightarrow a}$
in the
optimization.

\begin{figure}[t!]
    \centering
    \begin{tikzpicture}[scale=0.8]
        \begin{axis}[
            width=\linewidth,
            height=0.6\linewidth,
            grid=both,
            grid style={dashed, gray!30},
            xlabel={$w_{b\rightarrow a}^{\mathrm{BiNI}^{(t-1)}}$},
            ylabel={$\beta_{b\rightarrow a}^{(t)}$},
            xmax=1.08,
            xmin=0.1,
            ymax=1.05,
            ymin=-0.05,
            xlabel style={below right},
            axis y line=left,
            axis x line=middle,
            every axis x label/.style={at={(current axis.right of origin)},right=15mm,below=-1mm},
            every axis y label/.style={at={(current axis.north west)},above=8.5mm,left=0.5mm},
            enlargelimits=true,
            ticklabel style={font=\small},
            label style={font=\small},
            title style={font=\small},
            legend pos=south east
        ]
            \addplot[domain=0:1, samples=200, thick, iccvblue] {1. / (1 + exp(50*(x - 0.25)))};
        \end{axis}
    \end{tikzpicture}
    \caption{\textbf{Discontinuity activation term~\eqref{eq:beta_definition_ours} for} $q=50.0$ \textbf{and} $\rho=0.25$. The term $\beta_{b\rightarrow a}^{(t)}$
    progressively 
    incorporates
    discontinuities
    in our formulation,
    which
    correspond to
    $w_{b\rightarrow a}^{\mathrm{BiNI}^{(t-1)}} < 0.5$.
    }
    \vspace{-10pt}
    \label{fig:beta_a_from_b}
\end{figure}
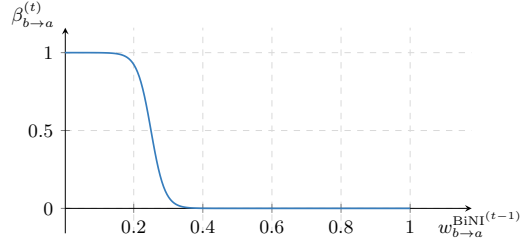
While the above procedure allows optimizing the log-depth values, as
noted
in \cref{sec:method_solving_for_depth_and_discontinuities} we would like to
additionally update our discontinuity terms $\alpha_{b\rightarrow a}^{(t)}$, to model discontinuities with increasingly higher accuracy. To this purpose,
we invert~\eqref{eq:ours_formulation_log} to derive the following
update scheme:
\begin{equation}
    \alpha_{b\rightarrow a}^{(t+1)} \gets 
    \left(\exp\left({\tilde{z}_a^{(t)}} - {\tilde{z}_b^{(t)}}\right) - \omega_{b\rightarrow a}\right)/\omega_{\varepsilon_a},
    \label{eq:update_scheme_alpha}
\end{equation}
with $\alpha_{b\rightarrow a}^{(0)}=0$ for all valid pairs $(a, b)$.
However, applying this update to all
pairs
would cause the optimization to converge in one iteration to a suboptimal solution, since its objective~\eqref{eq:general_optimization_problem} would evaluate
to $0$.
To avoid this, 
we introduce
an additional term
$\beta_{b\rightarrow a}^{(t)}\in[0, 1]$
in~\eqref{eq:ours_formulation_log_with_bini_factor},
which
selectively activates the discontinuity terms,
as follows:
\begin{equation}
\medmath{\gamma_{b\rightarrow a}\left(\tilde{z}_a^{(t)} -\tilde{z}_b^{(t)}\right) = \gamma_{b\rightarrow a}\log\left(\omega_{b\rightarrow a}+\omega_{\varepsilon_a}\cdot\alpha_{b\rightarrow a}^{(t)}\cdot\beta_{b\rightarrow a}^{(t)}\right)}.
    \label{eq:ours_relationship_equation_t}
\end{equation}

The rationale for $\beta_{b\rightarrow a}^{(t)}$ is the following:
If the surface is estimated to be continuous between pixels $a$ and $b$, one can approximate $\alpha_{b\rightarrow a}\approx 0$,
so
the influence of the discontinuity term should be negligible.
On the other hand, the more the surface is estimated to be discontinuous between $a$ and $b$, the more
the term
$\alpha_{b\rightarrow a}$
would
increase the accuracy
of~\eqref{eq:ours_formulation_log_with_bini_factor}, and thus
the more it should be taken into account.
We note that the weights $w_{b\rightarrow a}^{\mathrm{BiNI}}$ naturally model this relationship.
Indeed,
as the optimization identifies with increasing confidence that a discontinuity is present between 
$a$ and $b$
the corresponding term $w_{b\rightarrow a}^{\mathrm{BiNI}}$ increasingly approaches $0$. Viceversa, if the optimization identifies the surface to be continuous between $a$ and $b$, or at least equally discontinuous in the directions of the two opposite pixels $b$ and $-b$, the term $w_{b\rightarrow a}^{\mathrm{BiNI}}$ is greater or equal than $0.5$ ($w_{b\rightarrow a}^{\mathrm{BiNI}}\rightarrow1$ in the first case and $w_{b\rightarrow a}^{\mathrm{BiNI}}\approx0.5$ in the latter). 
We therefore define the \emph{discontinuity activation} terms as
\begin{equation}
    \beta_{b\rightarrow a}^{(t)} = \sigma\left(q\cdot\left(\rho - w_{b\rightarrow a}^{\mathrm{BiNI}^{(t-1)}}\right)\right),
    \label{eq:beta_definition_ours}
\end{equation}
where we set $q=50.0$ and $\rho=0.25$, which guarantees that $\beta_{b\rightarrow a}^{(t)}$ tends smoothly to $1$ as $w_{b\rightarrow a}^{\mathrm{BiNI}^{(t-1)}}\rightarrow 0$ and smoothly to $0$ as $w_{b\rightarrow a}^{\mathrm{BiNI}^{(t-1)}}\rightarrow 0.5^{-}$ (\cf. \cref{fig:beta_a_from_b}). We study the impact of the
hyperparameters $q$ and $\rho$ in \Cref{sec_suppl:ablation_beta_a_from_b_discont_activation_term}.

\section{Experiments\label{sec:experiments}}
This Section provides
the
experimental evaluation of our method, describing our experimental setup (\cref{sec:experimental_settings}), comparing the accuracy of its formulation to that of existing ones (\cref{sec:comparison_accuracy_formulation}), evaluating its normal integration accuracy on a standard benchmark (\cref{sec:benchmark_experiments_pinhole}), and demonstrating its applicability to generic central 
cameras
(\cref{sec:experiments_nonpinhole_camera_models}) and real-world input normals (\cref{sec:experiments_real_world_data}).
Readers can find
ablations, additional experimental results, and a discussion of the limitations of our method in the Supplementary Material.
\subsection{Experimental settings\label{sec:experimental_settings}}
\textbf{Baselines.} We compare our method to the state-of-the-art 
BiNI~\cite{Cao2022BiNI} and
Kim~\etal~\cite{Kim2024DiscontinuityPreserving} on the DiLiGenT benchmark~\cite{Shi2016DiLiGenT}. As no source code is publicly available for~\cite{Kim2024DiscontinuityPreserving}, in the remaining evaluations we only compare our method to BiNI,
setting its hyperparameter to its default value.\\
\textbf{Hardware and timing.} We run all our evaluations on a standard CPU-only machine, on which our unoptimized implementation takes between $\num{50}$ and $\num{120}$ seconds for $\num{1200}$ iterations with an input normal map of size $512\times612$.

\subsection{Comparison of formulation accuracy\label{sec:comparison_accuracy_formulation}}
Before examining the quality of the reconstruction produced by our optimization, we assess how accurately our
formulation approximates the ground-truth relation between depth and surface normals compared to existing PDE-derived formulations. 
To this
end,
we
compute
for both
our method and
BiNI 
the
absolute residual emerging
from the respective
formulations;
as previously noted, this
has for both
the same functional
form
$|\gamma_{b\rightarrow a}(\tilde{z}_a - \tilde{z}_b) - \mathrm{RHS}|$, where $\mathrm{RHS}$ denotes the right-hand side of \eqref{eq:bini_formulation} for BiNI and of \eqref{eq:ours_formulation_log_with_bini_factor} for
our method.
We evaluate
this quantity on
the DiLiGenT 
dataset~\cite{Shi2016DiLiGenT},
assuming
for fairness
unknown discontinuity values, thereby
setting the terms $\alpha_{b\rightarrow a}$ in our
formulation
to
$0$.
As shown in \cref{tab:comparison_formulation_accuracy}, our
method
achieves
mean error lower by one or two orders of magnitude on all but one object.
We provide additional
comparisons
using
relative residuals in \cref{sec_suppl:additional_evaluations_formulation_accuracy},
where we find
similar results.

\subsection{Benchmark experiments\label{sec:benchmark_experiments_pinhole}}
\begin{figure*}[!ht]
\centering
\def\colwidth{0.1\textwidth}
\newcolumntype{M}[1]{>{\centering\arraybackslash}m{#1}}
\addtolength{\tabcolsep}{-4pt}
\begin{tabular}{m{0.7em} M{\colwidth} M{\colwidth} M{\colwidth}  M{\colwidth}  M{\colwidth} M{\colwidth} M{\colwidth} M{\colwidth} M{\colwidth}}
 & \small\texttt{bear} & \small\texttt{buddha} & \small\texttt{cat} & \small\texttt{cow} & \small\texttt{harvest} & \small\texttt{pot1} & \small\texttt{pot2} & \small\texttt{reading} & \small\texttt{goblet} \tabularnewline

\begin{turn}{90}
GT
\end{turn} & 
\includegraphics[width=\linewidth]{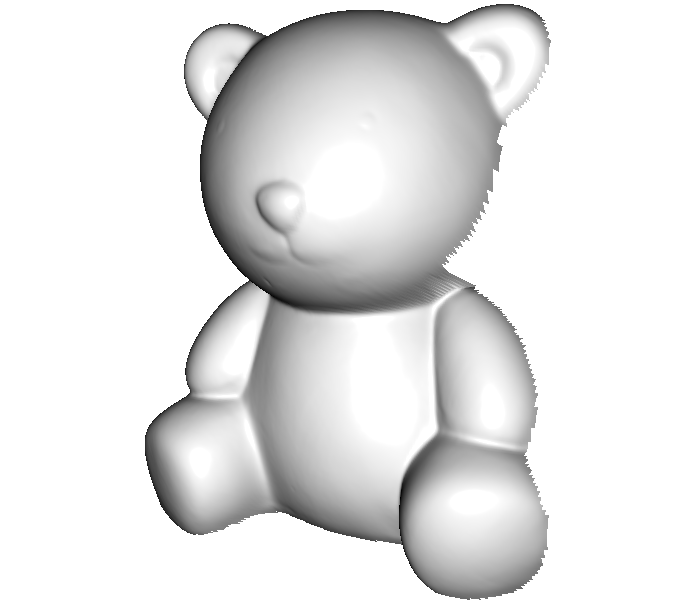} & 
\includegraphics[width=\linewidth]{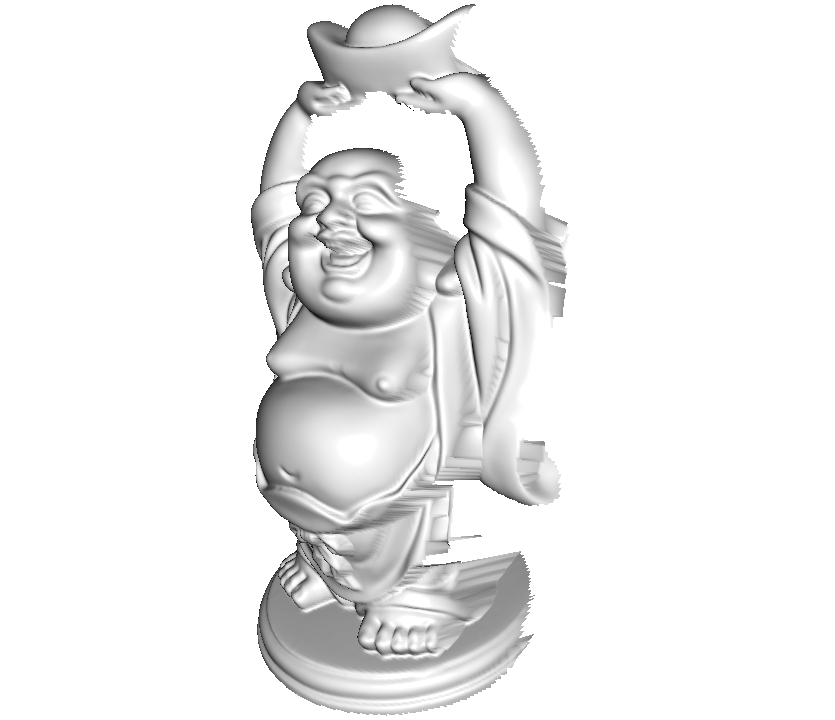} &
\includegraphics[width=\linewidth]{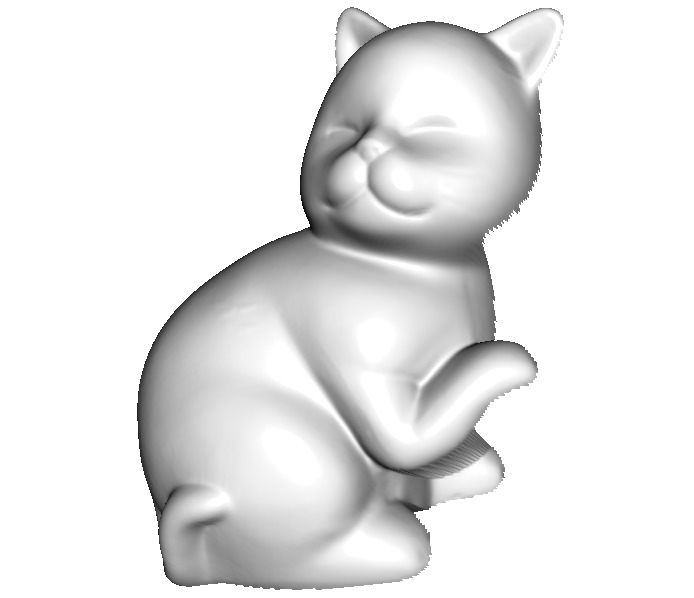} &
\includegraphics[width=\linewidth]{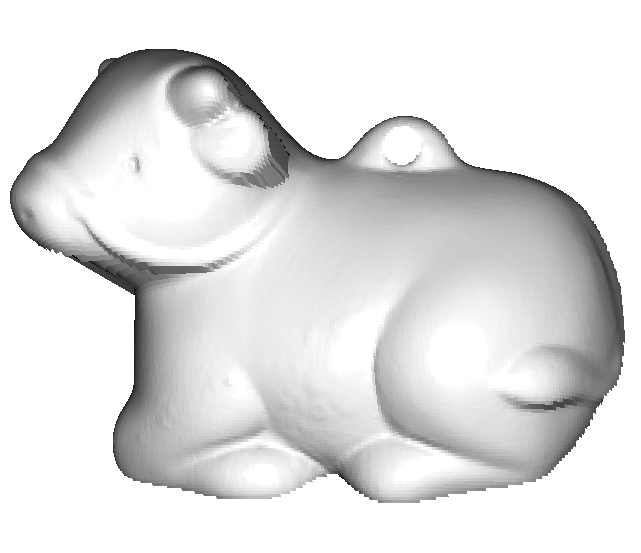} &
\includegraphics[width=\linewidth]{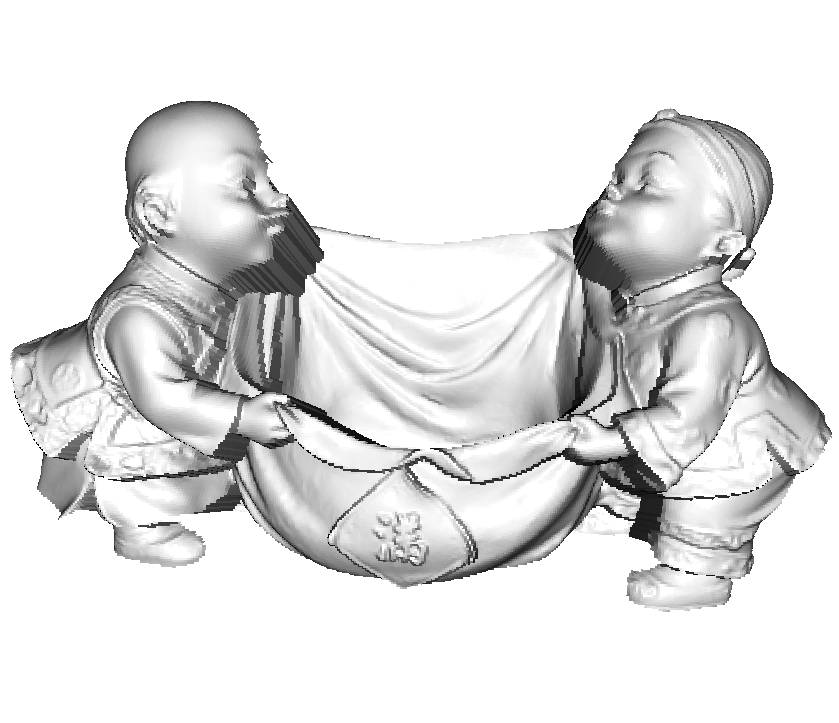} &
\includegraphics[width=\linewidth]{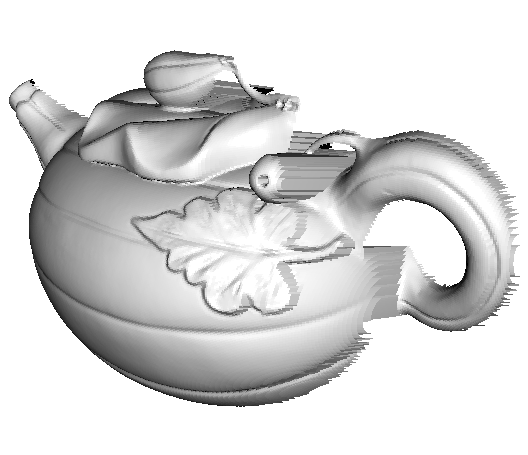} &
\includegraphics[width=\linewidth]{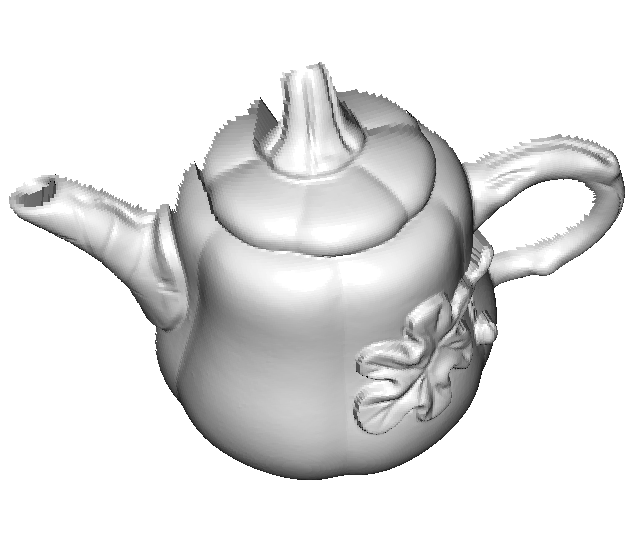} &
\includegraphics[width=\linewidth]{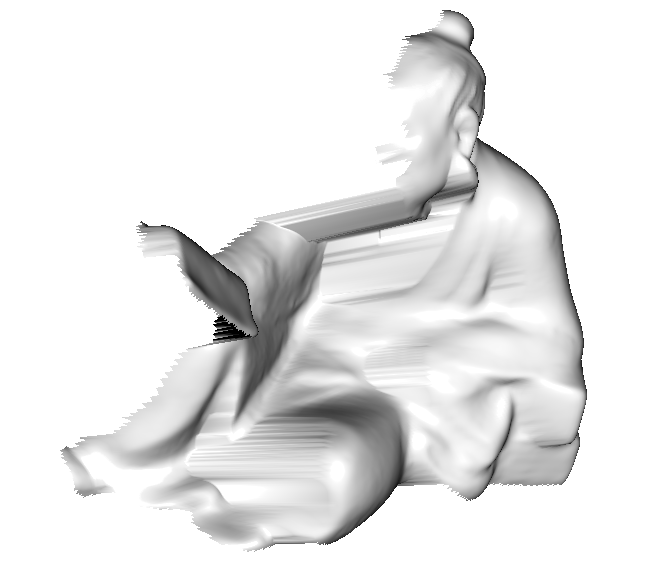} &
\includegraphics[width=\linewidth]{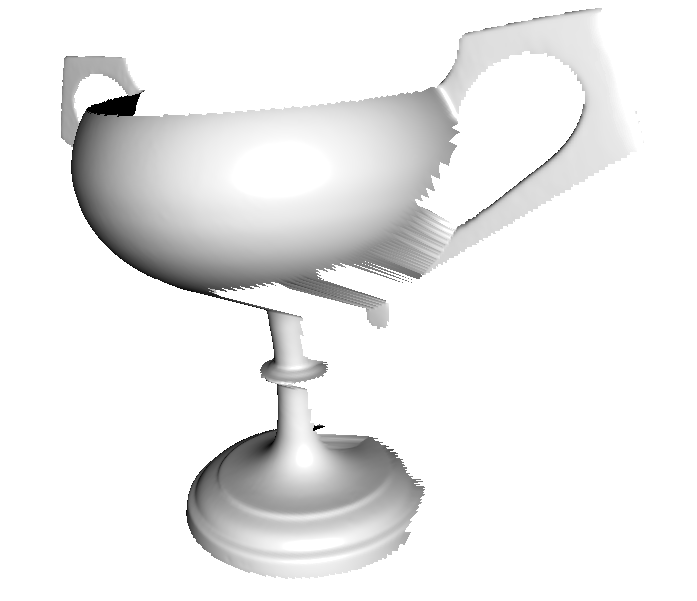}
\\
\noalign{\vskip -0.2em} 
\cline{2-10}
\noalign{\vskip 0.2em}
\multirow{3}{*}{\parbox[c][10ex][c]{\ht\strutbox}{\begin{turn}{90}
BiNI~\cite{Cao2022BiNI}
\end{turn}}} & 
\includegraphics[width=\linewidth]{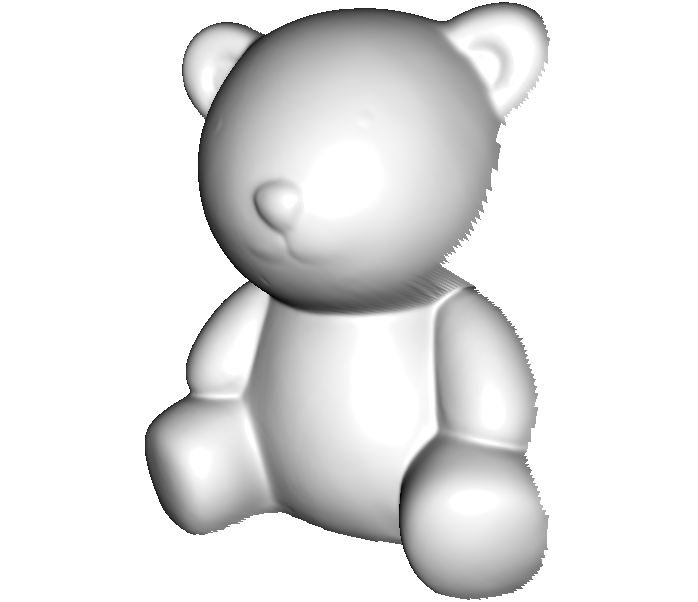} & 
\includegraphics[width=\linewidth]{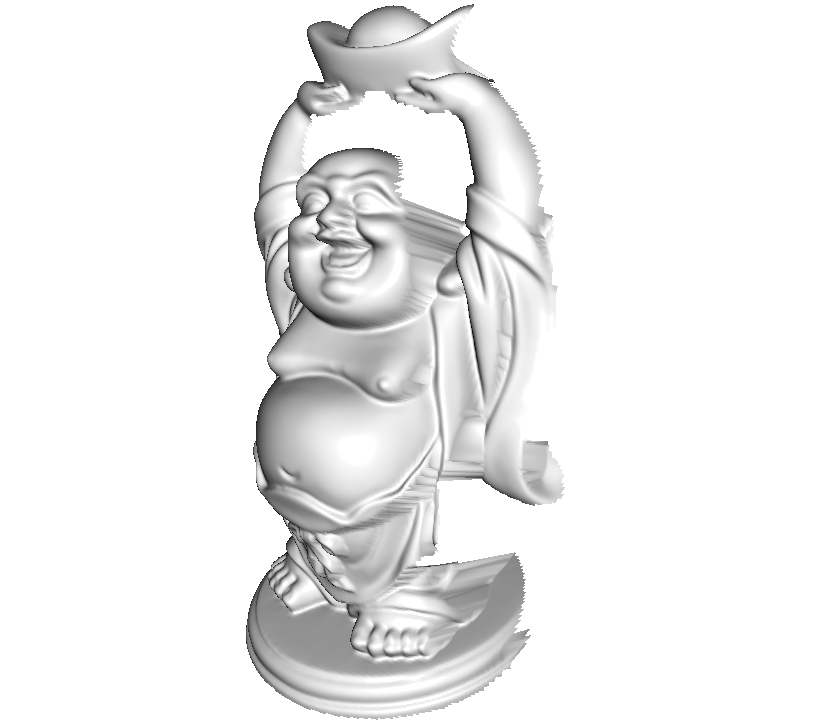} &
\includegraphics[width=\linewidth]{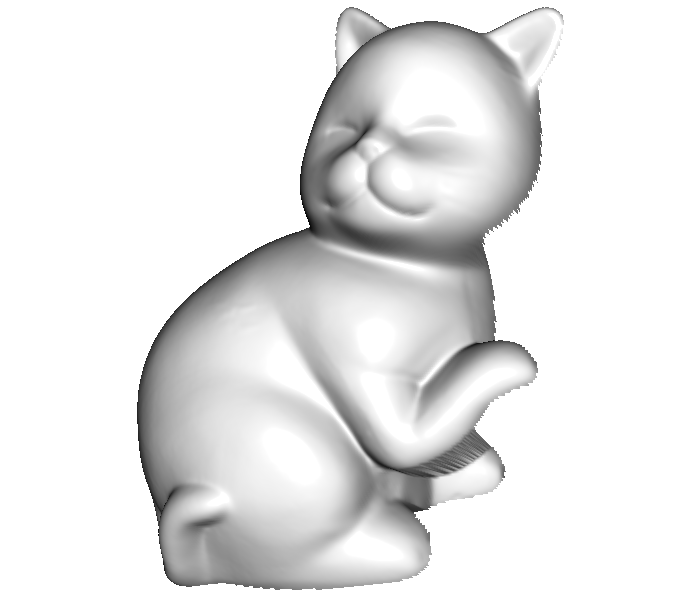} &
\includegraphics[width=\linewidth]{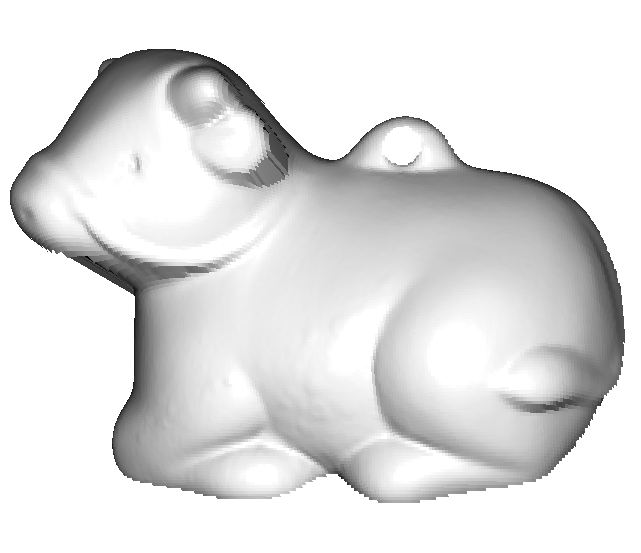} &
\includegraphics[width=\linewidth]{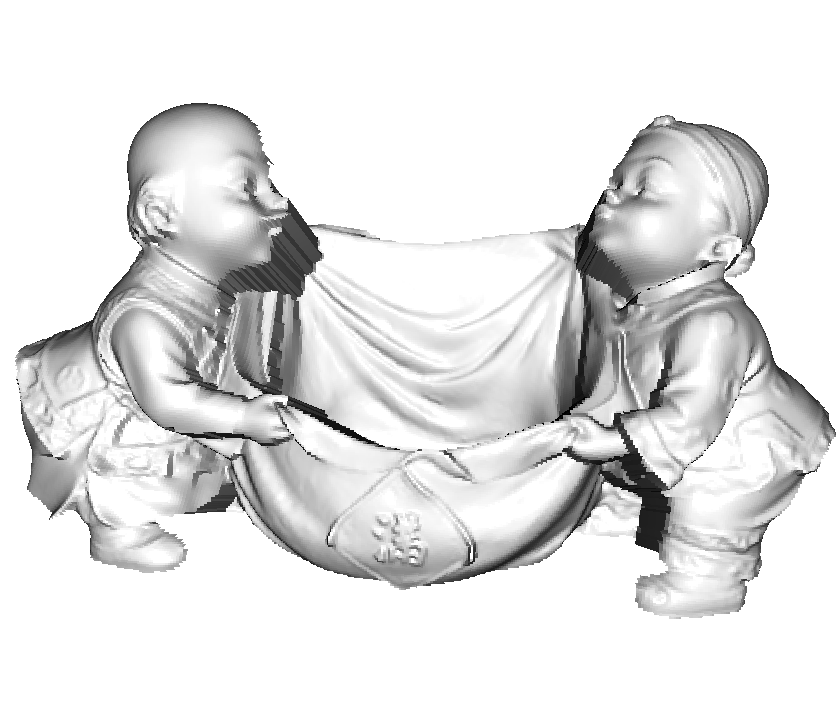} &
\includegraphics[width=\linewidth]{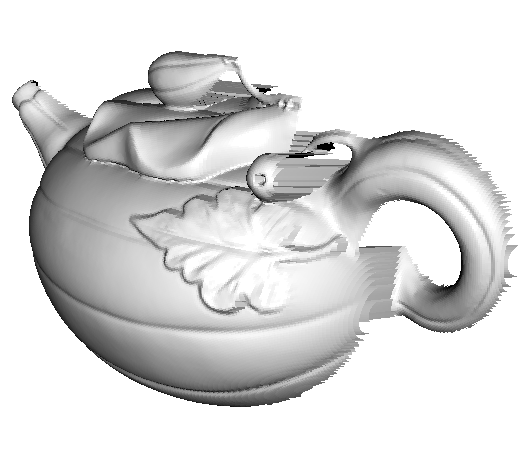} &
\includegraphics[width=\linewidth]{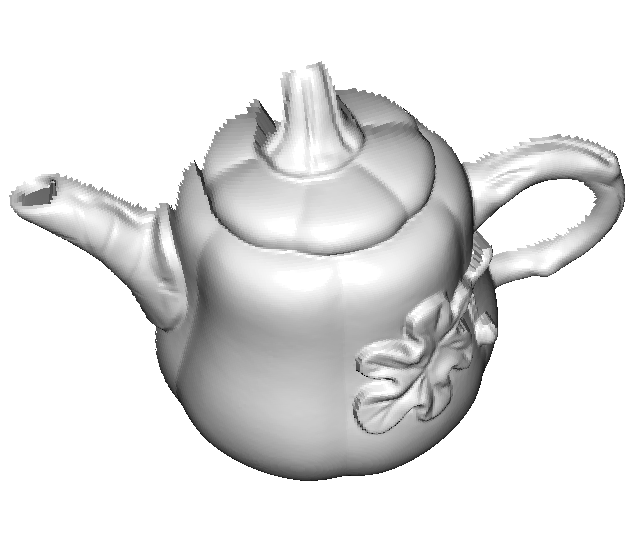} &
\includegraphics[width=\linewidth]{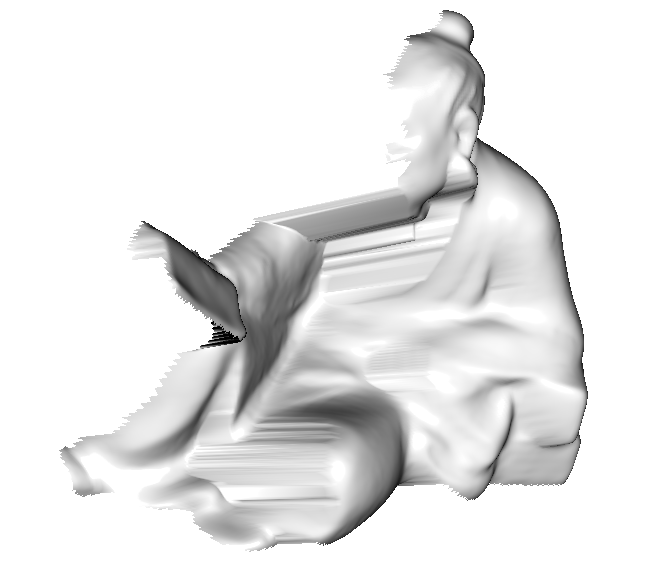} &
\includegraphics[width=\linewidth]{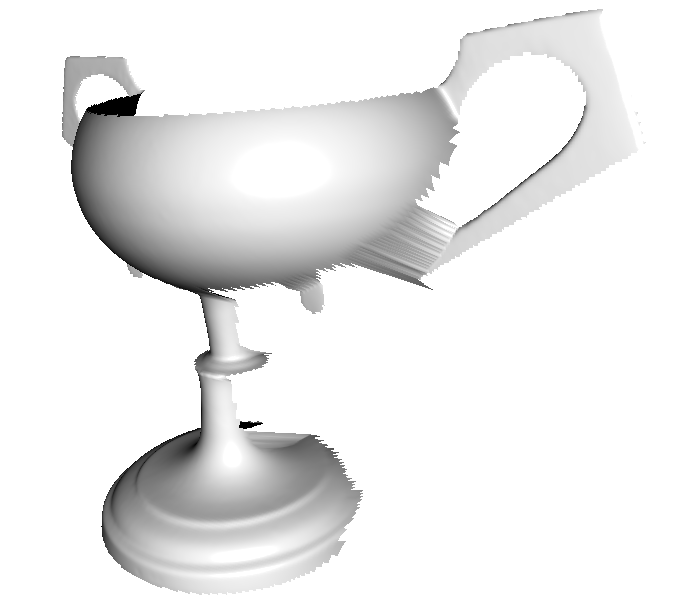}
\tabularnewline
& 
\includegraphics[width=\linewidth]{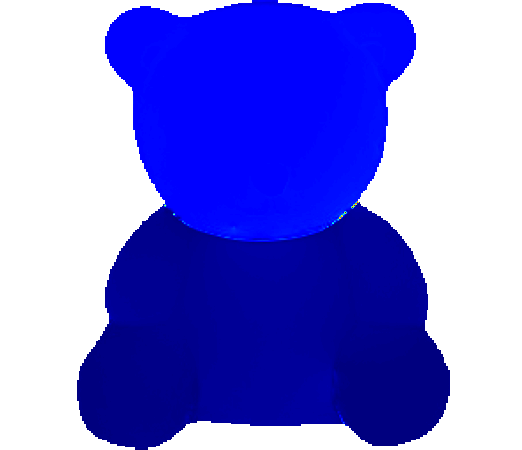} & 
\includegraphics[width=\linewidth]{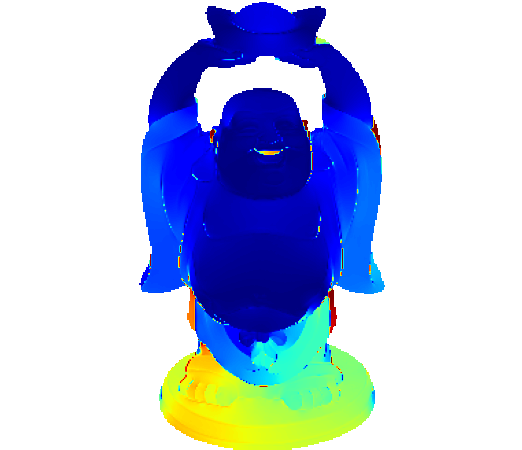} &
\includegraphics[width=\linewidth]{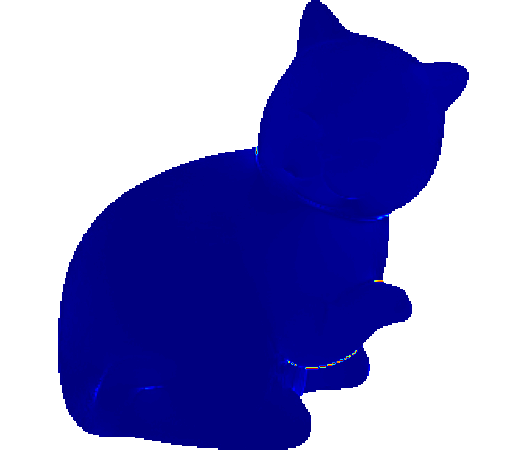} &
\includegraphics[width=\linewidth]{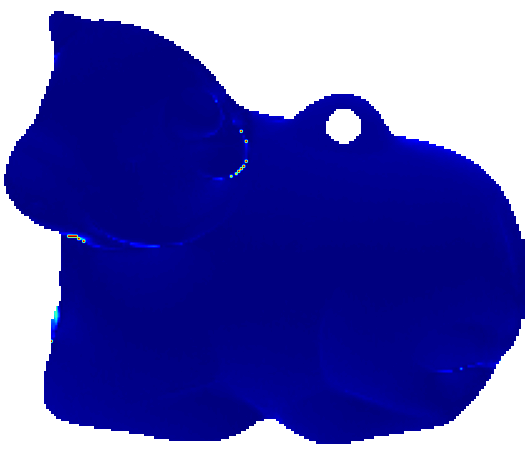} &
\includegraphics[width=\linewidth]{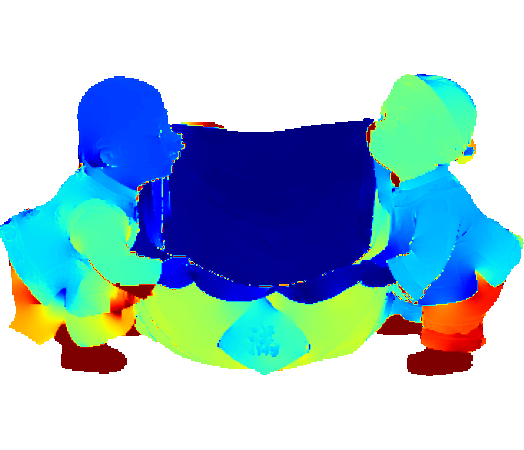} &
\includegraphics[width=\linewidth]{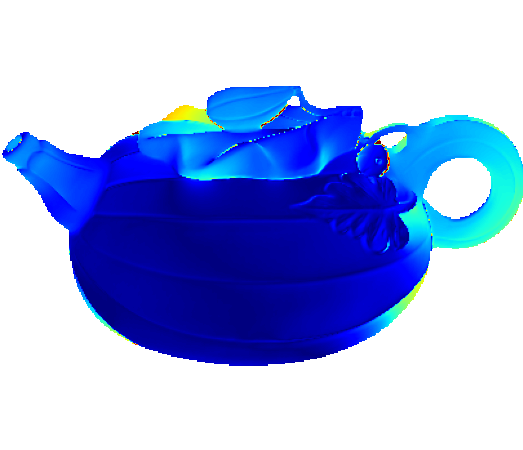} &
\includegraphics[width=\linewidth]{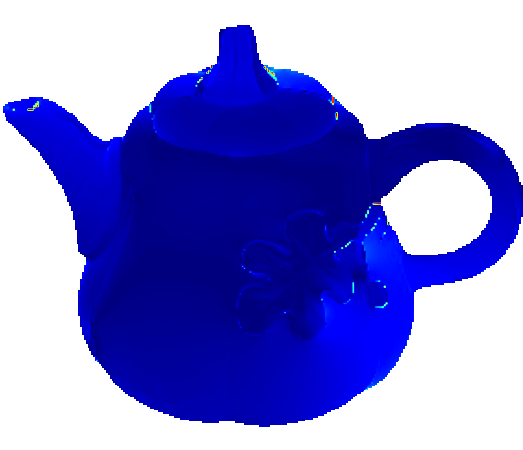} &
\includegraphics[width=\linewidth]{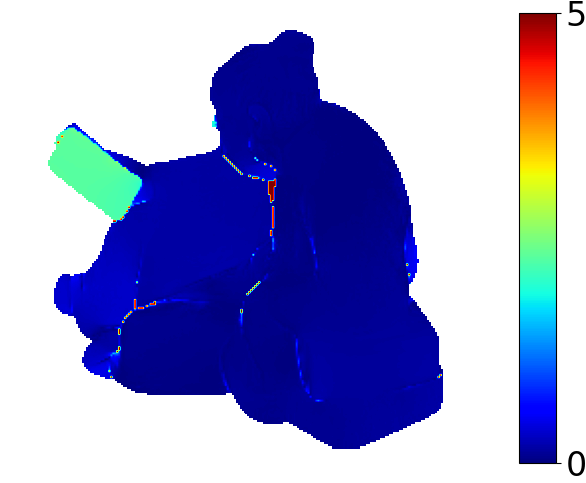} &
\includegraphics[width=\linewidth]{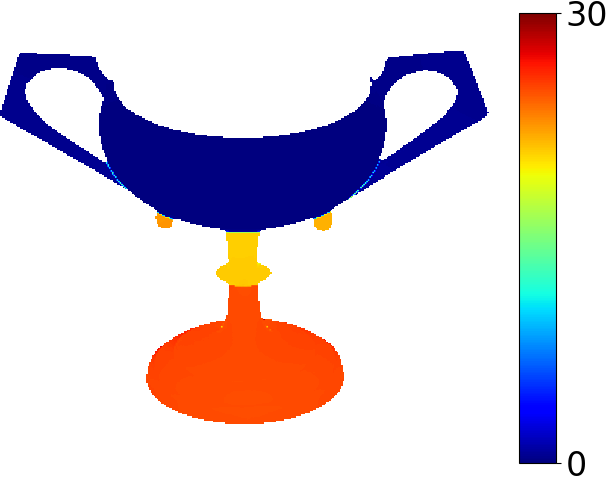}
\\[-5pt]
& \scriptsize{$0.33$} & \scriptsize{$1.06$} & \scriptsize{$0.07$} & \scriptsize{$0.06$} & \scriptsize{$1.84$} & \scriptsize{$0.64$} & \scriptsize{$0.22$} & \scriptsize{$0.26$} & \scriptsize{$9.00$}
\tabularnewline
\noalign{\vskip -0.2em} 
\cline{2-10}
\noalign{\vskip 0.2em}
\multirow{3}{*}{\parbox[c][10ex][c]{\ht\strutbox}{\begin{turn}{90}
Ours
\end{turn}}} & 
\includegraphics[width=\linewidth]{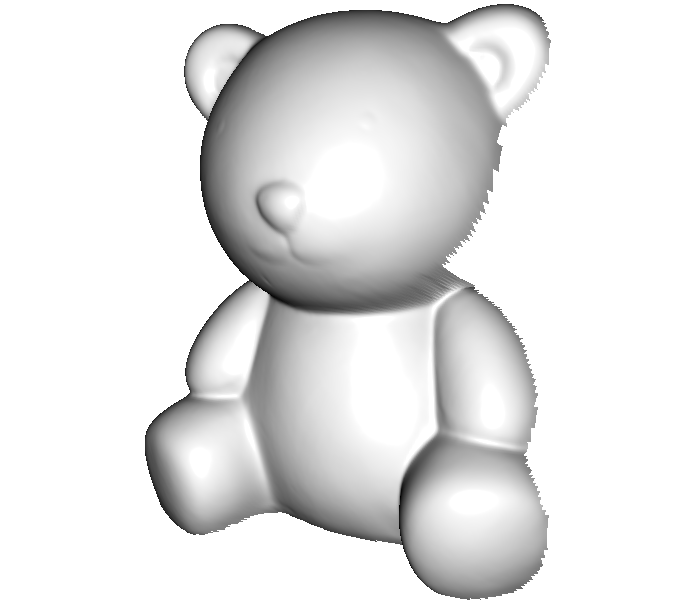} & 
\includegraphics[width=\linewidth]{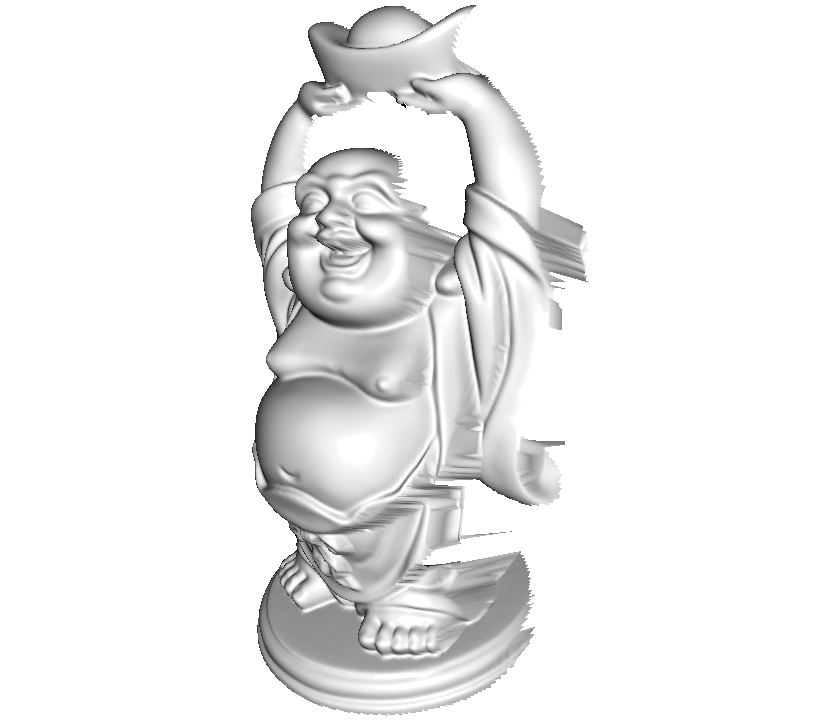} &
\includegraphics[width=\linewidth]{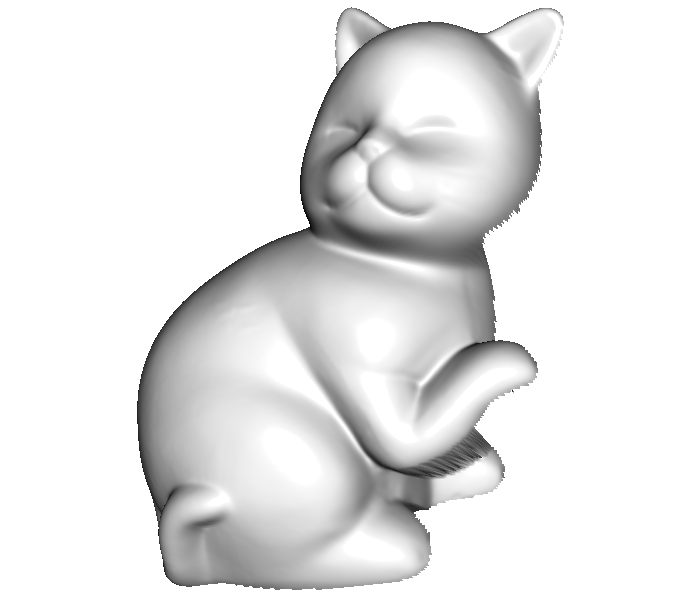} &
\includegraphics[width=\linewidth]{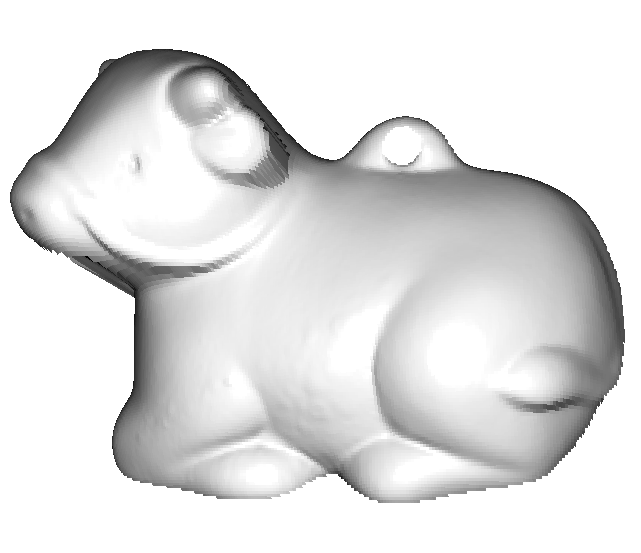} &
\includegraphics[width=\linewidth]{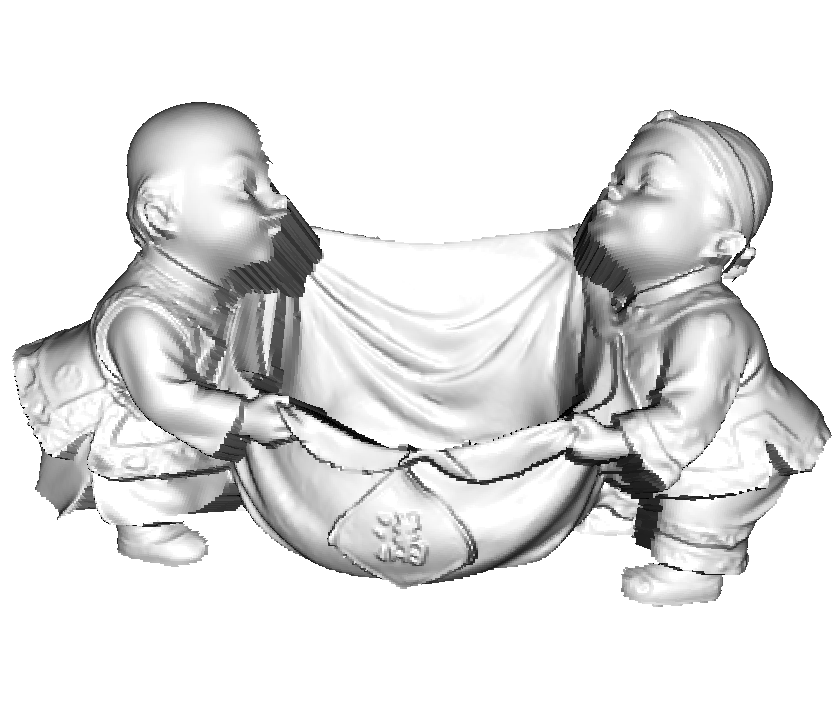} &
\includegraphics[width=\linewidth]{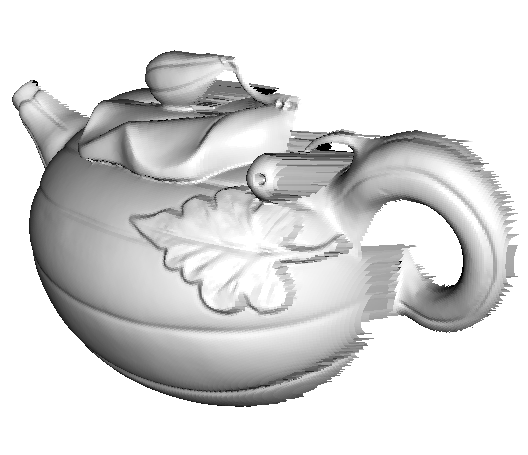} &
\includegraphics[width=\linewidth]{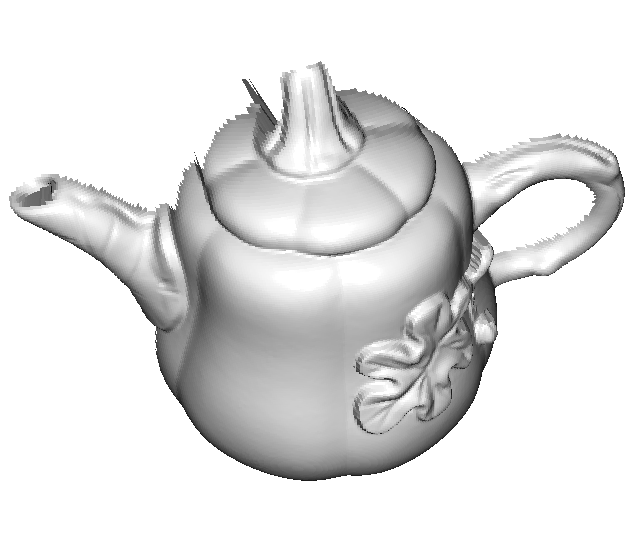} &
\includegraphics[width=\linewidth]{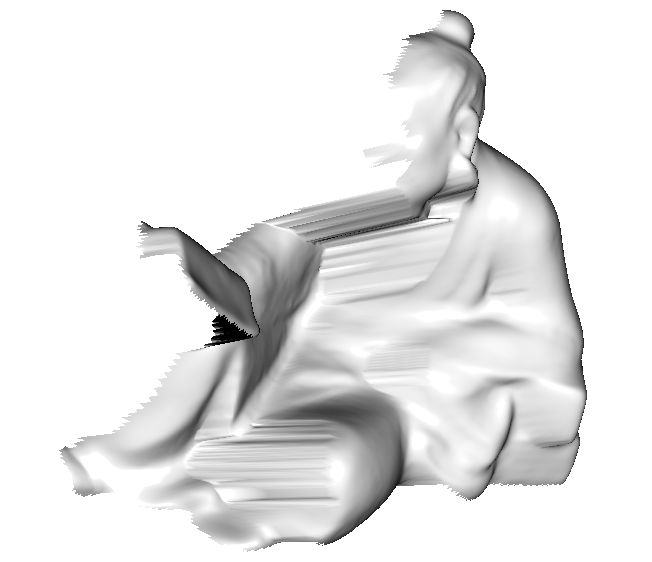} &
\includegraphics[width=\linewidth]{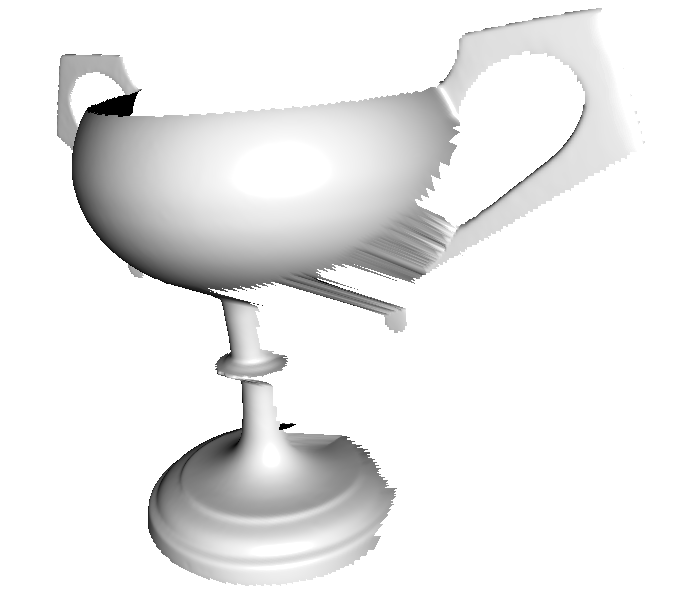}
\tabularnewline
& 
\includegraphics[width=\linewidth]{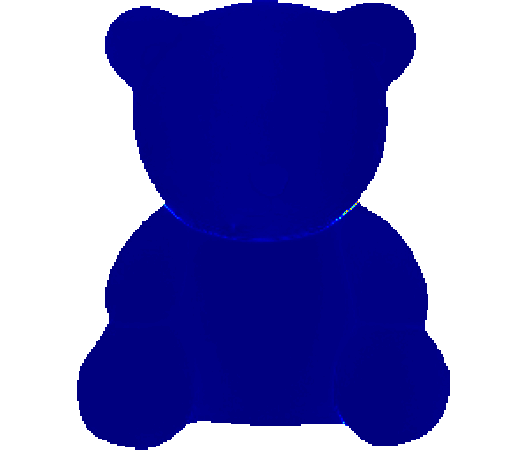} & 
\includegraphics[width=\linewidth]{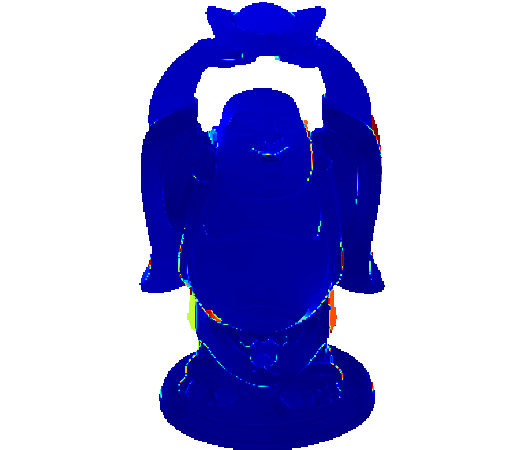} &
\includegraphics[width=\linewidth]{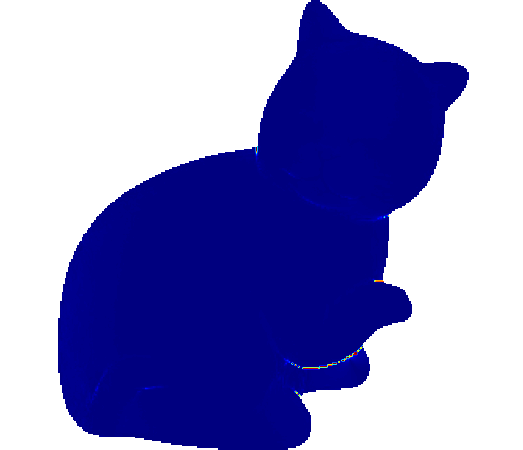} &
\includegraphics[width=\linewidth]{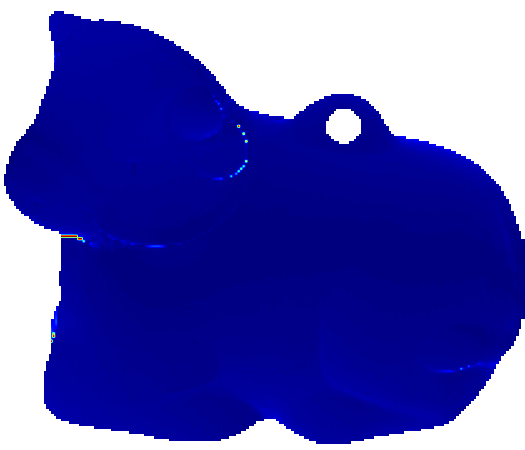} &
\includegraphics[width=\linewidth]{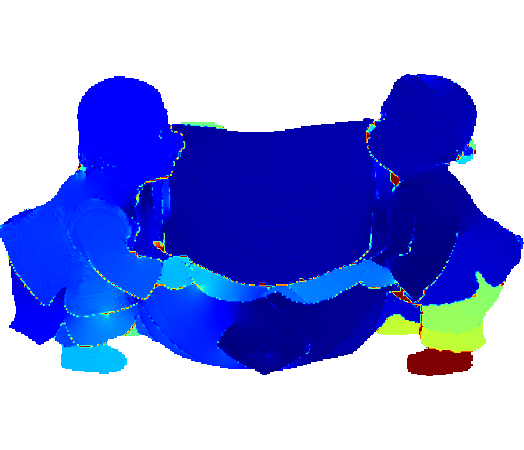} &
\includegraphics[width=\linewidth]{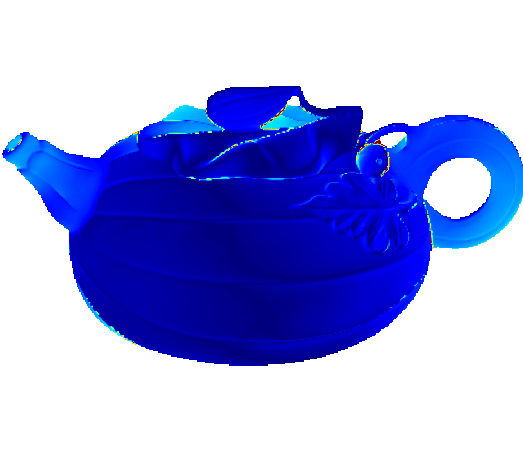} &
\includegraphics[width=\linewidth]{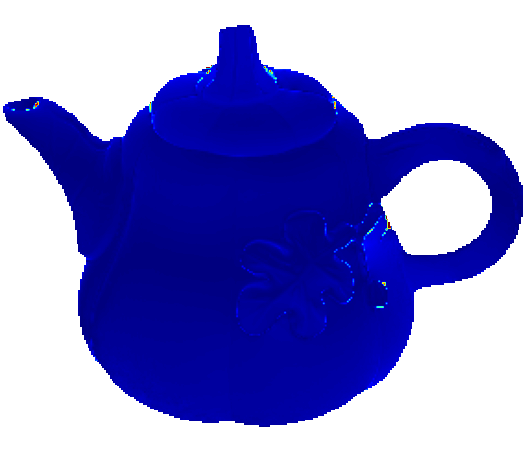} &
\includegraphics[width=\linewidth]{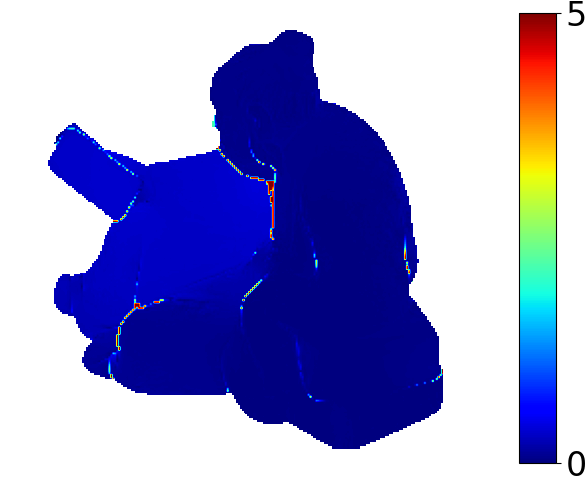} &
\includegraphics[width=\linewidth]{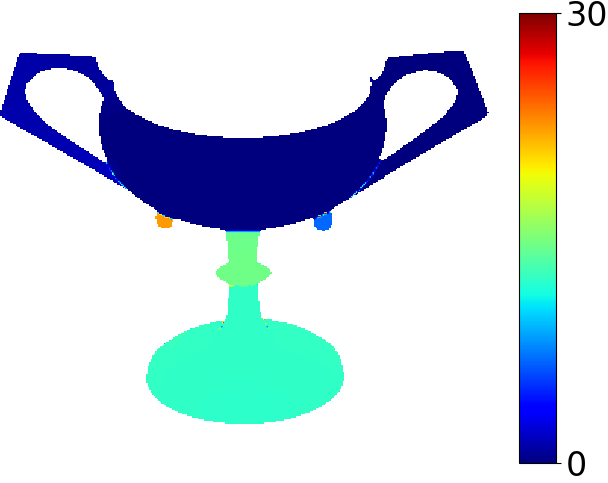}
\\[-5pt]
& \scriptsize{$0.03$} & \scriptsize{$0.24$} & \scriptsize{$0.06$} & \scriptsize{$0.08$} & \scriptsize{$0.73$} & \scriptsize{$0.49$} & \scriptsize{$0.13$} & \scriptsize{$0.17$} & \scriptsize{$4.72$}
\tabularnewline
\end{tabular}
\addtolength{\tabcolsep}{4pt}
\caption{\textbf{Comparison on the DiLiGenT benchmark~\cite{Shi2016DiLiGenT}.} First row: Ground-truth surfaces. Second and third row: Surface reconstructed by BiNI~\cite{Cao2022BiNI}; absolute depth errors maps (in $\textrm{mm}$). Fourth and fifth row: Surface reconstructed by our method with explicit discontinuity computation; absolute depth error maps. The color map is the same for the first eight columns. Below each absolute depth error map is the corresponding mean value (MADE) in $\textrm{mm}$. The absolute depth error maps are displayed from the viewpoint of the input normal map.}
\vspace{-6pt}
\label{fig:visualizations_diligent}
\end{figure*}

\begin{figure*}[!ht]
\centering
\def\colwidth{0.25\textwidth}
\newcolumntype{M}[1]{>{\centering\arraybackslash}m{#1}}
\addtolength{\tabcolsep}{10pt}
\begin{tabular}{M{\colwidth} M{\colwidth} M{\colwidth}}
\tabularnewline
\includegraphics[width=\linewidth]{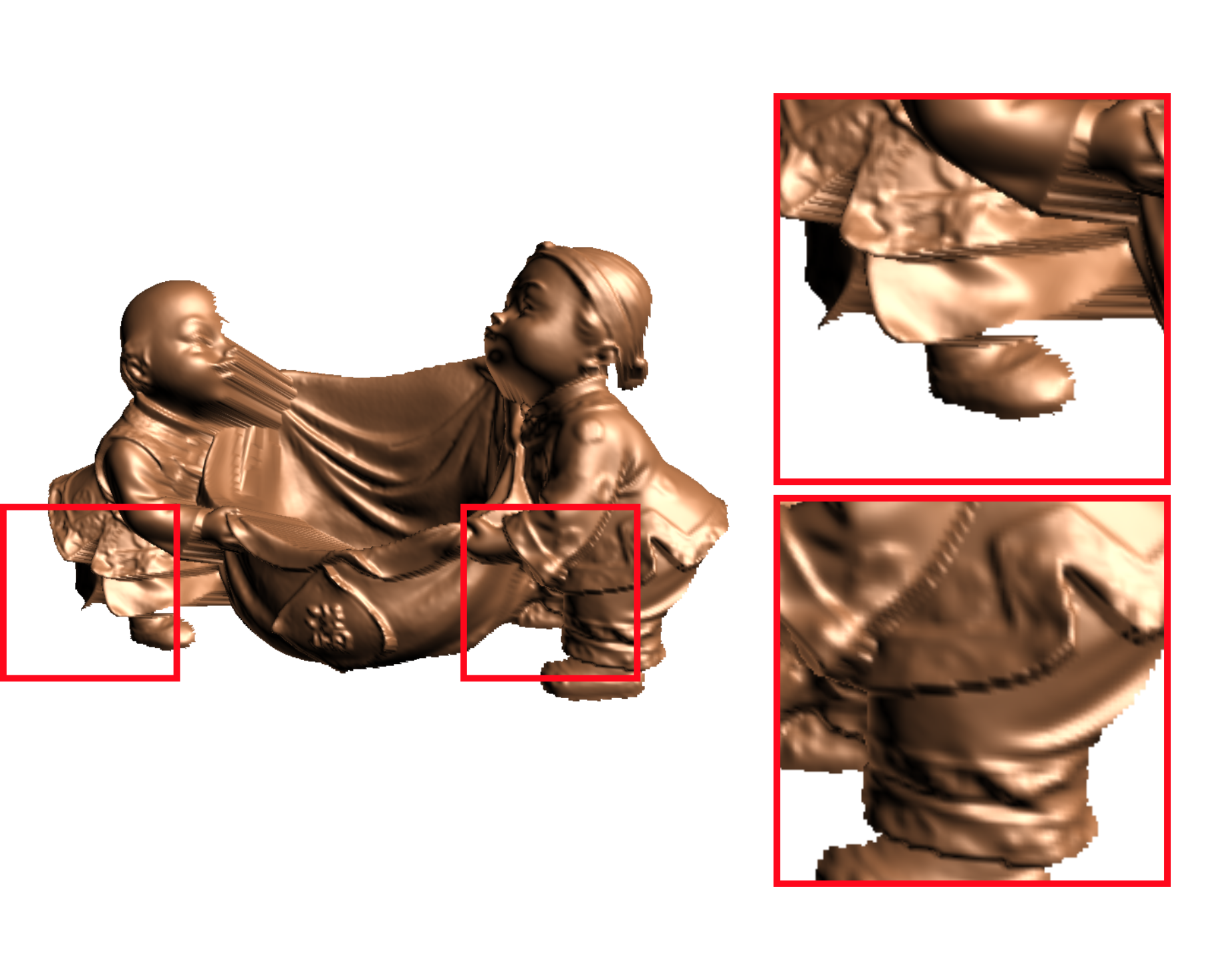} & 
\includegraphics[width=\linewidth]{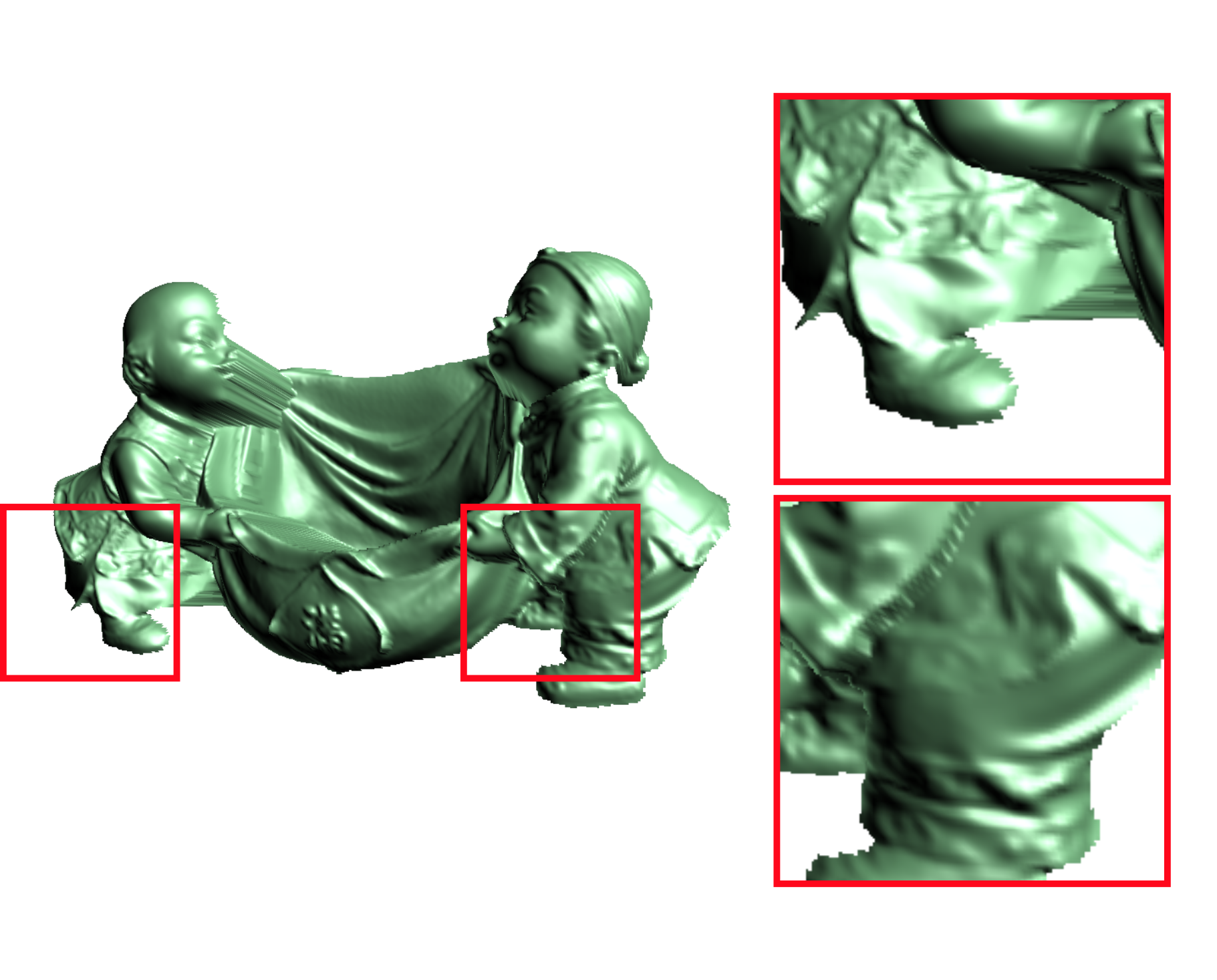} & \includegraphics[width=\linewidth]{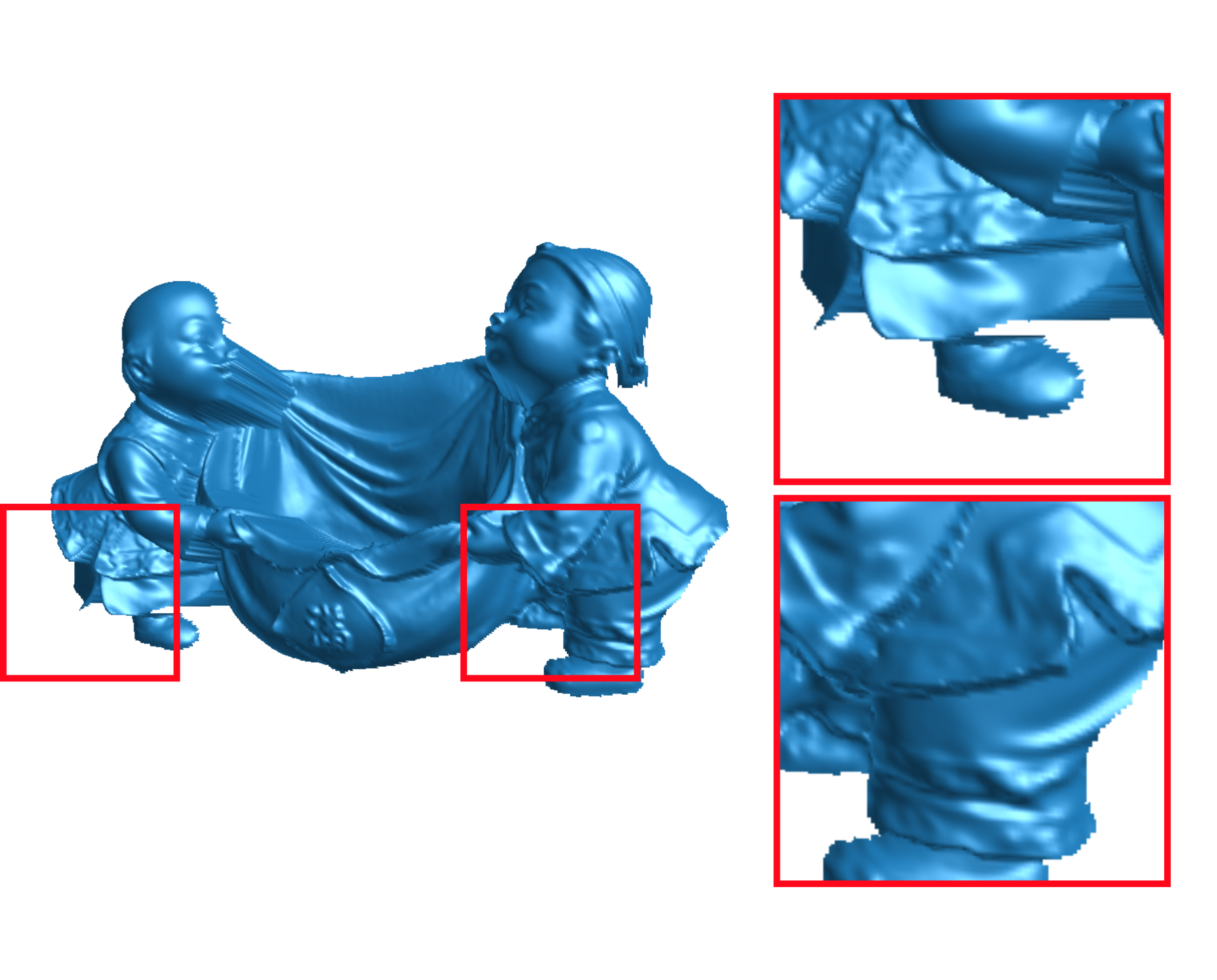}
\\[-10pt]
\includegraphics[width=\linewidth]{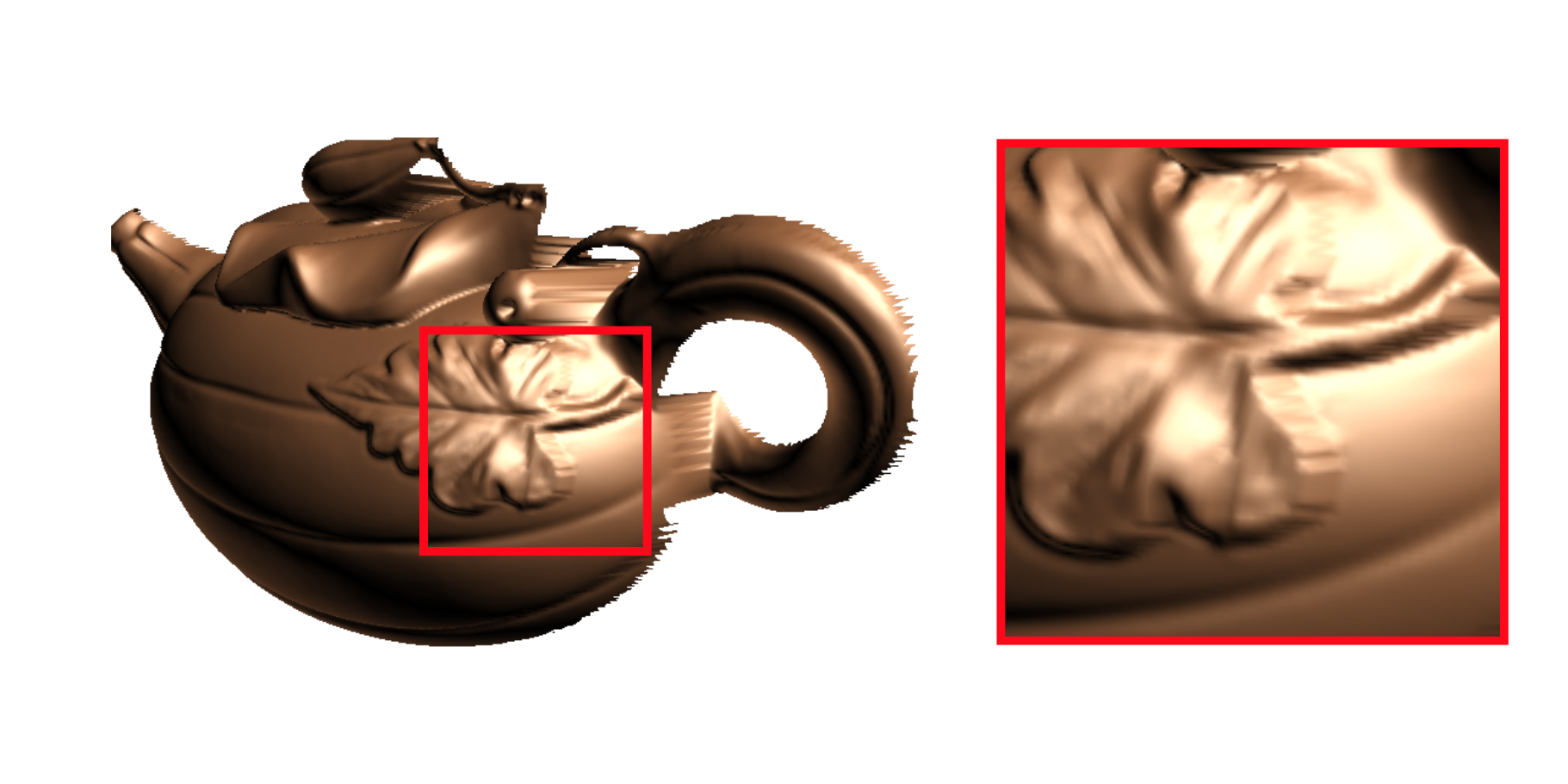} & 
\includegraphics[width=\linewidth]{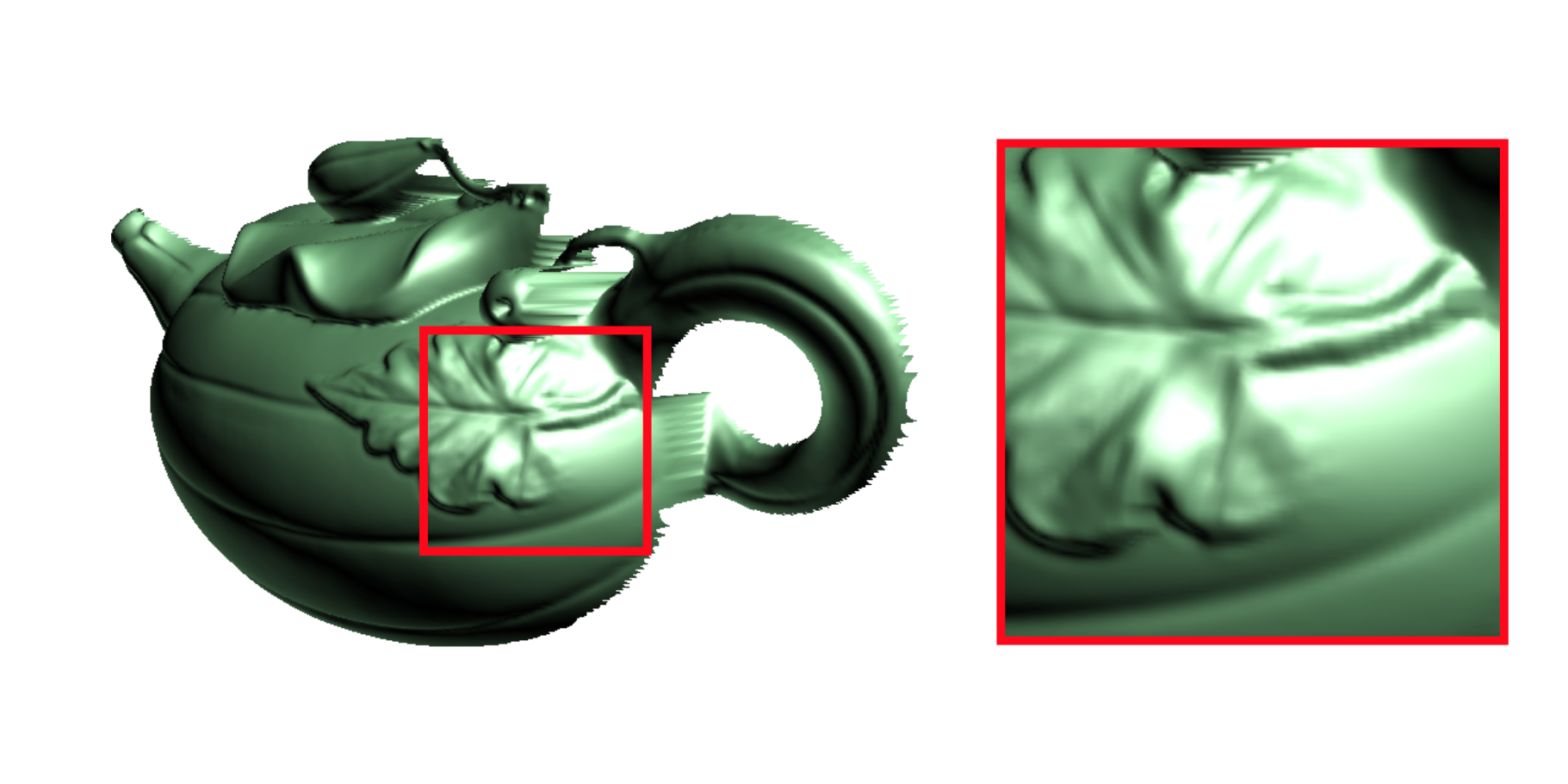} & \includegraphics[width=\linewidth]{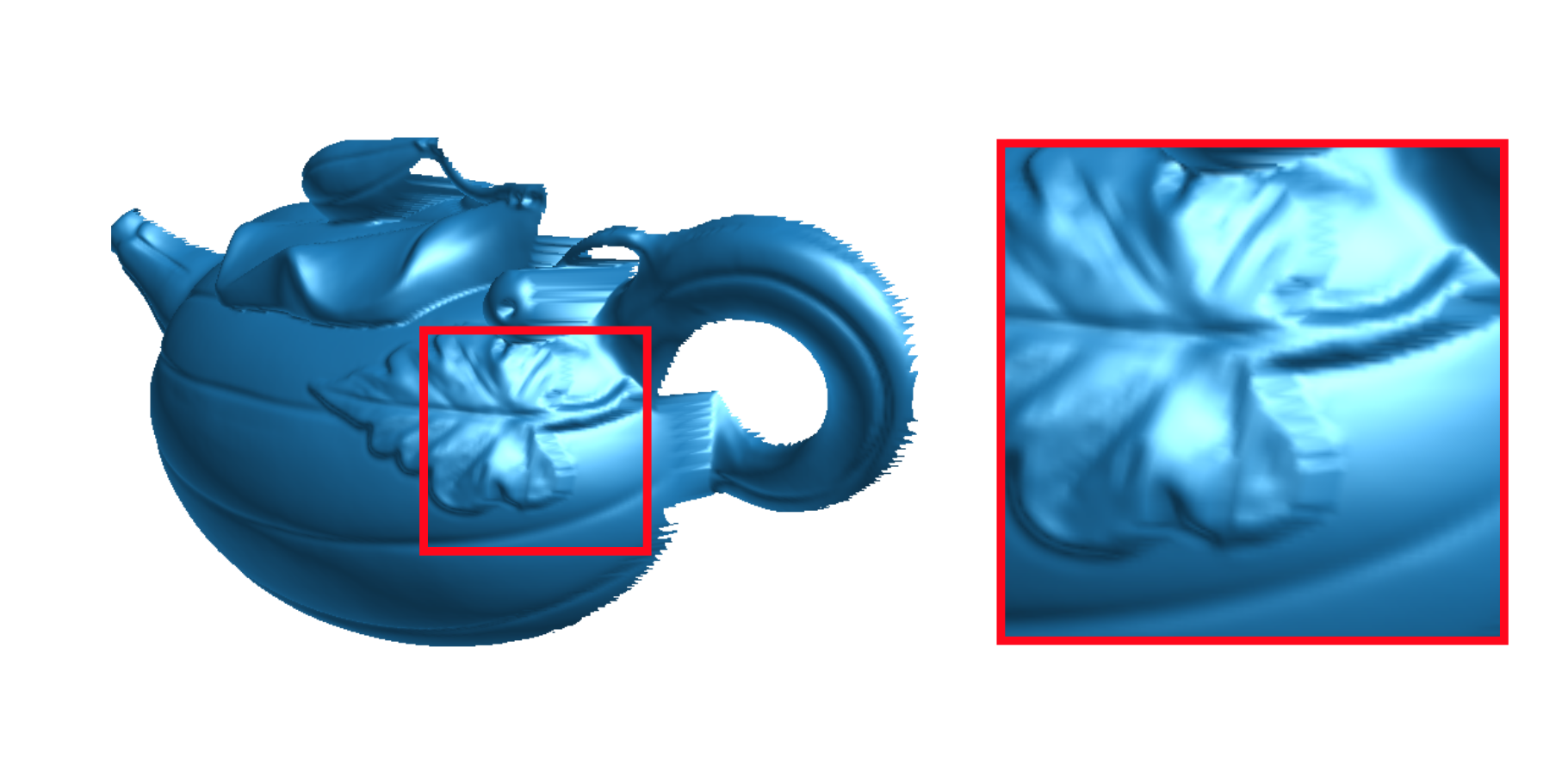}
\tabularnewline
\small{Ground truth} & \small{BiNI~\cite{Cao2022BiNI}} & \small{Ours}
\tabularnewline
\end{tabular}
\addtolength{\tabcolsep}{-10pt}
\caption{\textbf{Detail of the reconstructed surfaces.} Our
formulation
allows
capturing
discontinuities
with higher accuracy than
the previous method of BiNI~\cite{Cao2022BiNI}. Top and bottom rows show respectively objects \texttt{harvest} and \texttt{pot1} from the DiLiGenT benchmark~\cite{Shi2016DiLiGenT}.}
\label{fig:diligent_details}
\end{figure*}

We evaluate the reconstruction accuracy
of
our
normal integration
method
compared to
the state-of-the-art approaches~\cite{Cao2022BiNI} and~\cite{Kim2024DiscontinuityPreserving} on the standard DiLiGenT benchmark~\cite{Shi2016DiLiGenT}, which provides ground-truth normal maps produced by an ideal pinhole camera.
As shown by \cref{fig:visualizations_diligent} and \cref{tab:diligent_results_main}, our method without discontinuity computation (\ie, setting $\alpha_{b\rightarrow a} = 0$)
achieves
accuracy
that is state-of-the-art
for
$7$ out of $9$ objects, comparable
for $1$ object,
and worse
for a single
object.
This result shows that the higher accuracy of our formulation can effectively translate into better reconstruction quality through
the
optimization process. This is further confirmed by verifying that
using
coefficients
based on
discontinuity values $\alpha_{b\rightarrow a}$ from the ground-truth surface,
the optimization 
results in
an extremely low
error
for virtually all the objects.
We note that our method converges more slowly than BiNI, and we therefore run it for a larger number of iterations ($\num{1200}$);  however, after the same number of iterations necessary for BiNI to achieve convergence ($\num{150}$) our method already achieves better results than the other approaches on 
most
objects.
Iterative computation of
the terms $\alpha_{b\rightarrow a}$ allows more accurately capturing discontinuities (\cref{fig:diligent_details}) and further 
reduces
the reconstruction error, resulting in state of the art on virtually all objects.
Further, object-specific
improvements can be obtained
by tuning the hyperparameters of our discontinuity activation term (\cf. \cref{sec_suppl:ablation_beta_a_from_b_discont_activation_term}).

\subsection{Experiments with non-ideal pinhole cameras~\label{sec:experiments_nonpinhole_camera_models}}
\begin{figure*}[!t]
\centering
\def\colwidth{0.15\textwidth}
\def\minicolwidth{0.01\textwidth}
\newcolumntype{M}[1]{>{\centering\arraybackslash}m{#1}}
\addtolength{\tabcolsep}{-4pt}
\begin{tabular}{M{\colwidth} M{\colwidth} M{\colwidth} M{\minicolwidth} M{\colwidth} M{\colwidth}  M{\colwidth}  }
\includegraphics[width=\linewidth]{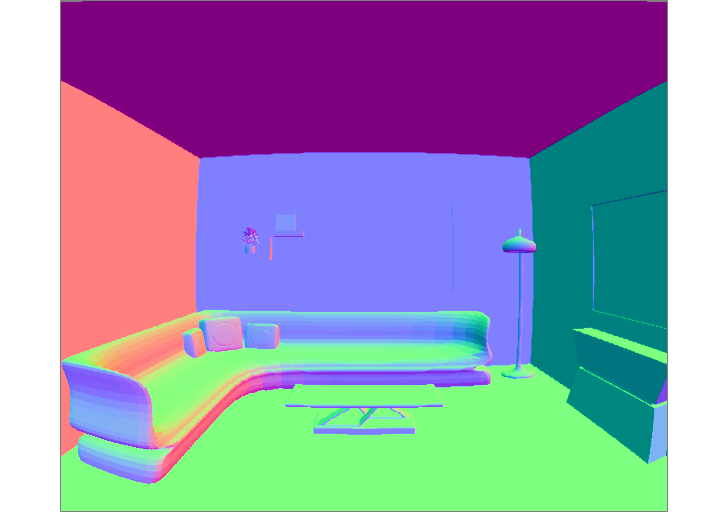} & 
\includegraphics[width=\linewidth]{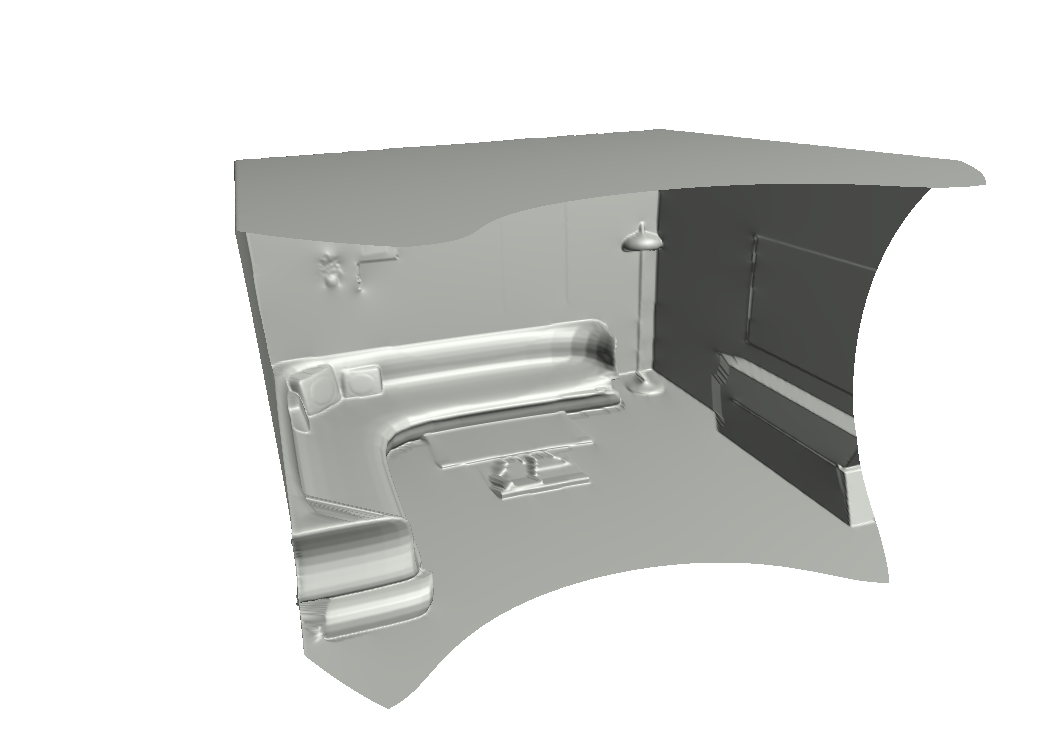} & 
\includegraphics[width=\linewidth]{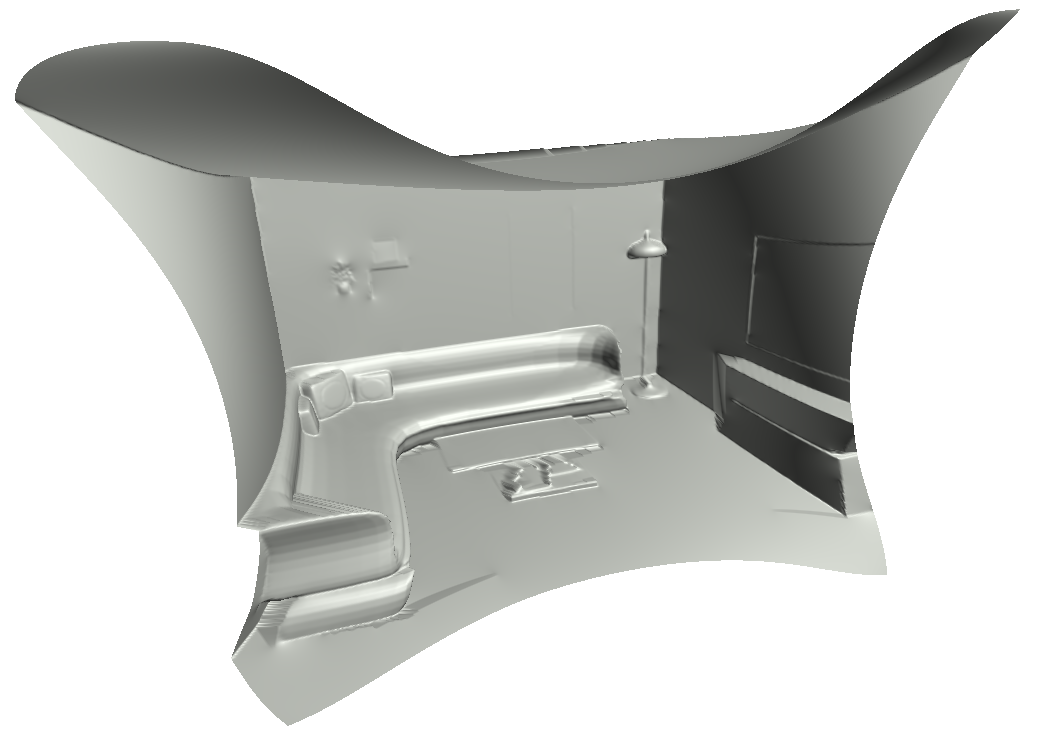} & 
& \includegraphics[width=\linewidth]{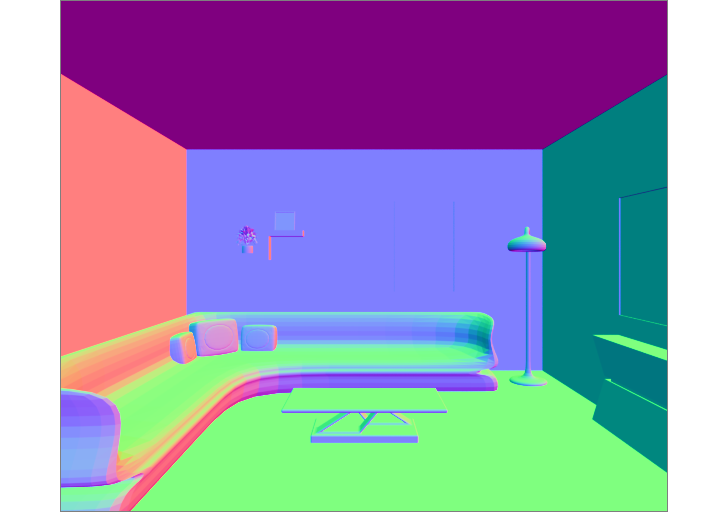} &
\includegraphics[width=\linewidth]{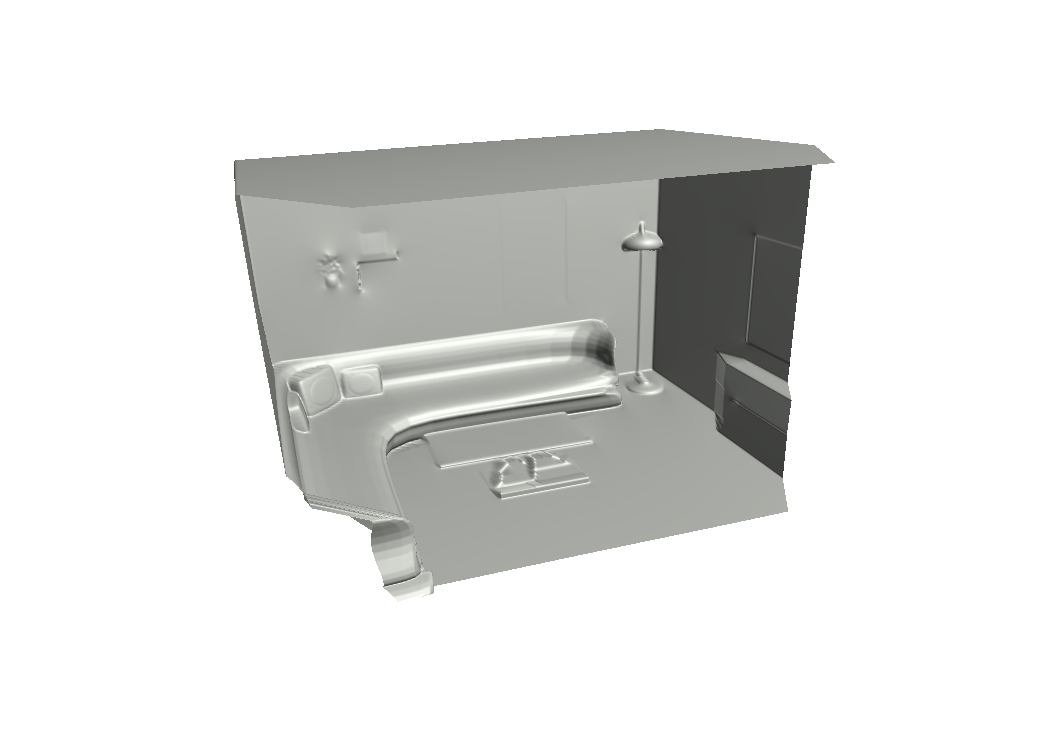} &
\includegraphics[width=\linewidth]{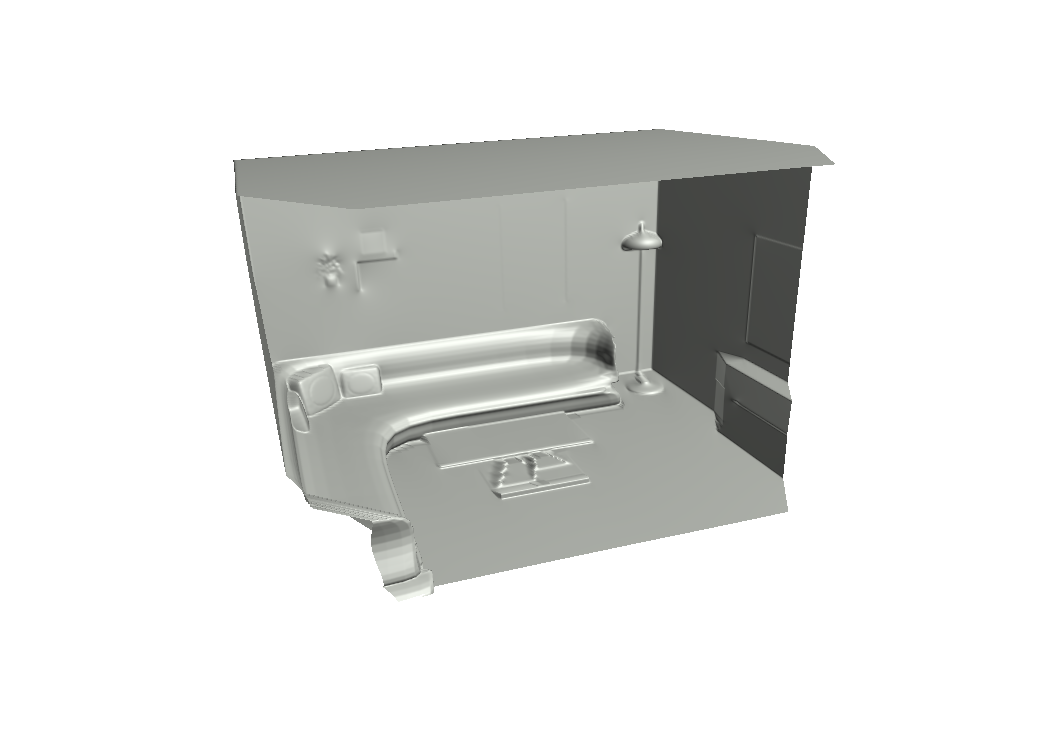} \\[-5pt]
&  \scriptsize{\textbf{RE}: $2.08$, \textbf{ERA}: $1.99$} & \scriptsize{\textbf{RE}: $8.01$, \textbf{ERA}: $6.47$} & & & \scriptsize{\textbf{RE}: $2.26$, \textbf{ERA}: $2.07$} & \scriptsize{\textbf{RE}: $2.30$, \textbf{ERA}: $2.09$}\\
\includegraphics[width=\linewidth]{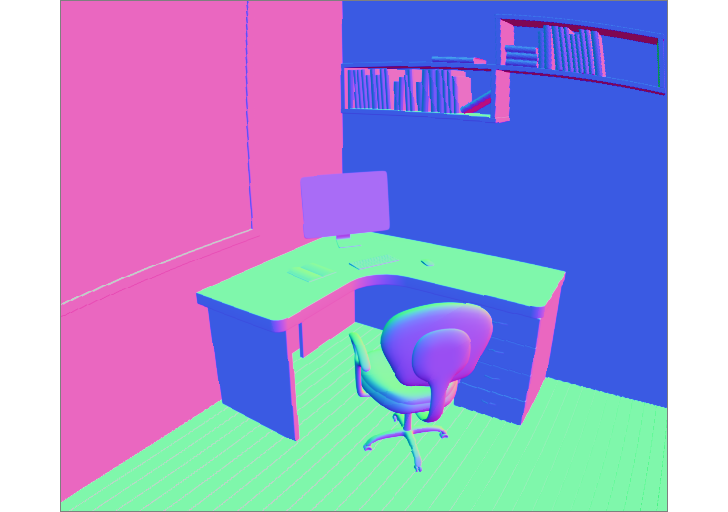} & 
\includegraphics[width=\linewidth]{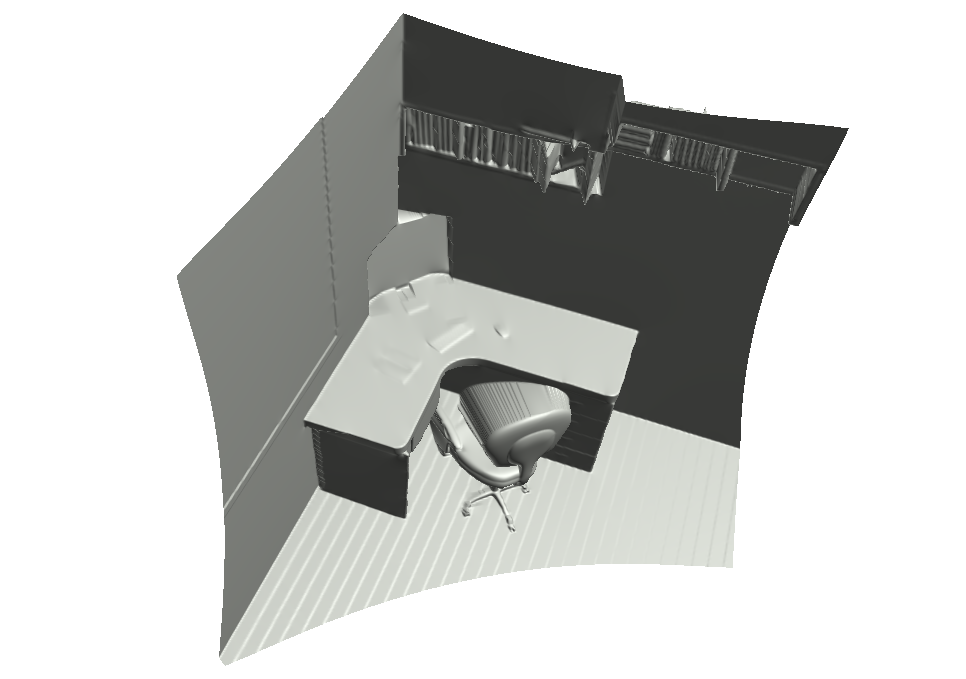} & 
\includegraphics[width=\linewidth]{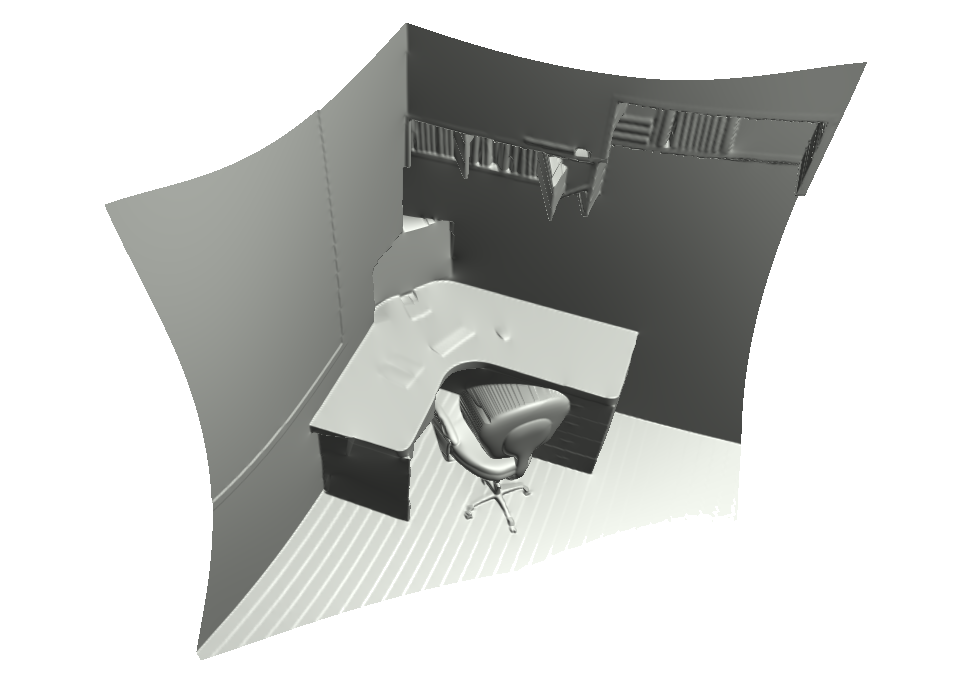} & 
& \includegraphics[width=\linewidth]{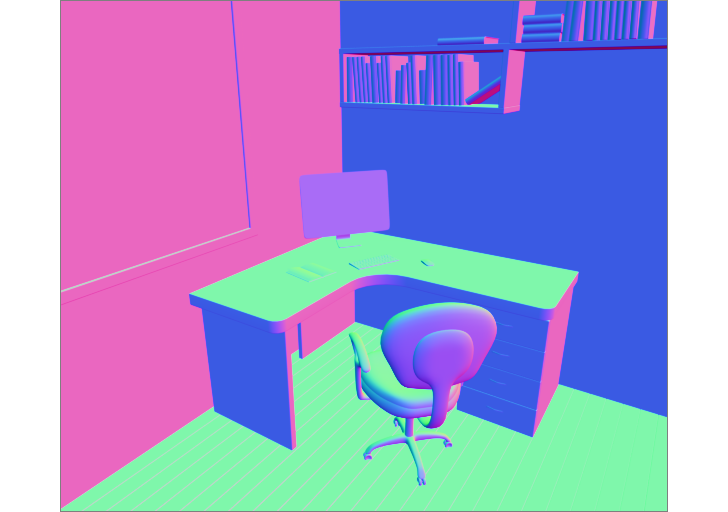} &
\includegraphics[width=\linewidth]{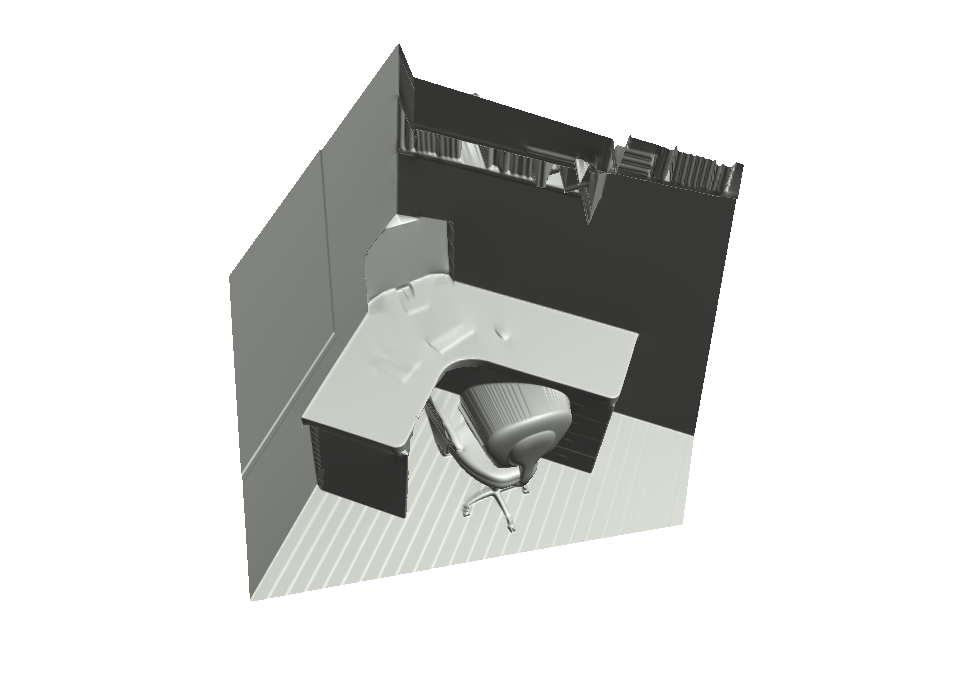} &
\includegraphics[width=\linewidth]{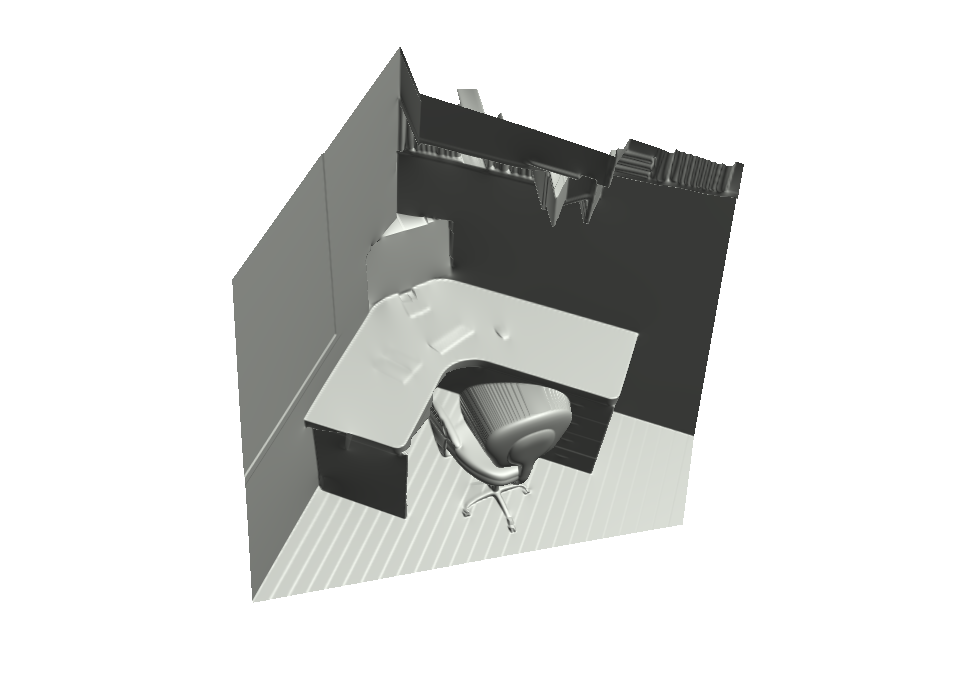} \\[-5pt]
&  \scriptsize{\textbf{RE}: $2.94$, \textbf{ERA}: $3.02$} & \scriptsize{\textbf{RE}: $4.44$, \textbf{ERA}: $4.29$} & & & \scriptsize{\textbf{RE}: $2.49$, \textbf{ERA}: $2.48$} & \scriptsize{\textbf{RE}: $2.63$, \textbf{ERA}: $2.61$}\\
\includegraphics[width=\linewidth]{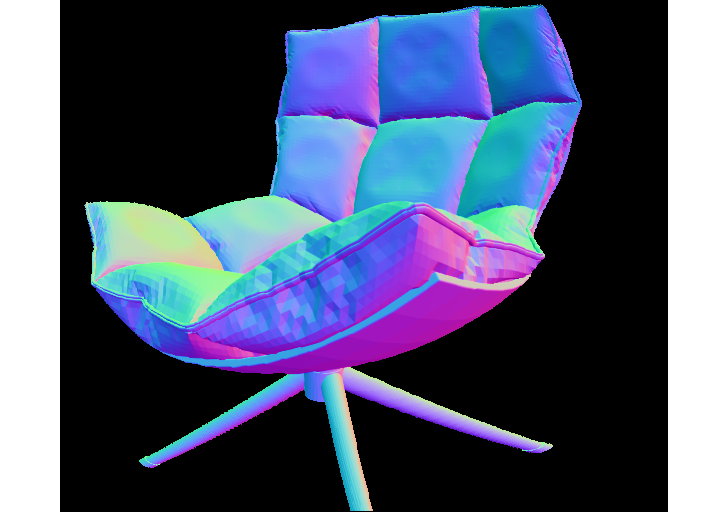} & 
\includegraphics[width=\linewidth]{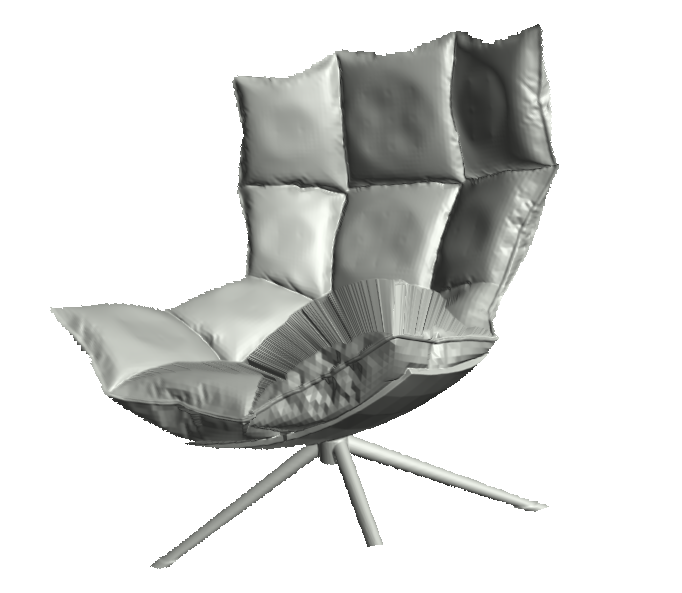} & 
\includegraphics[width=\linewidth]{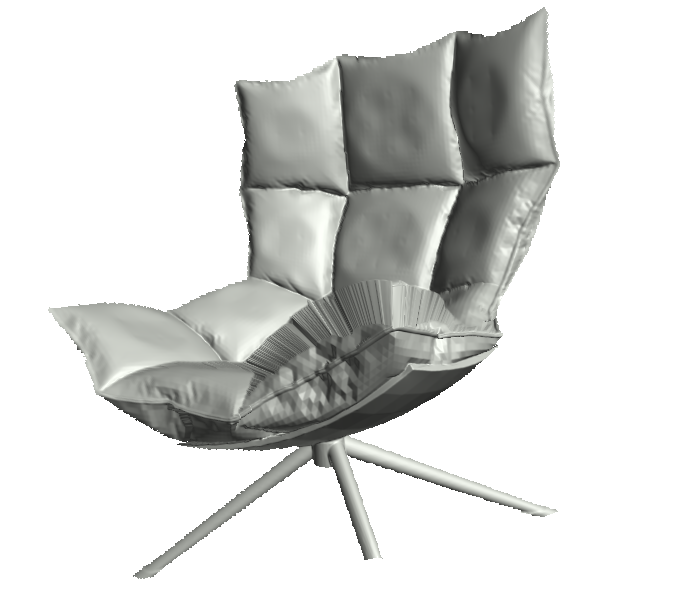} & 
& \includegraphics[width=\linewidth]{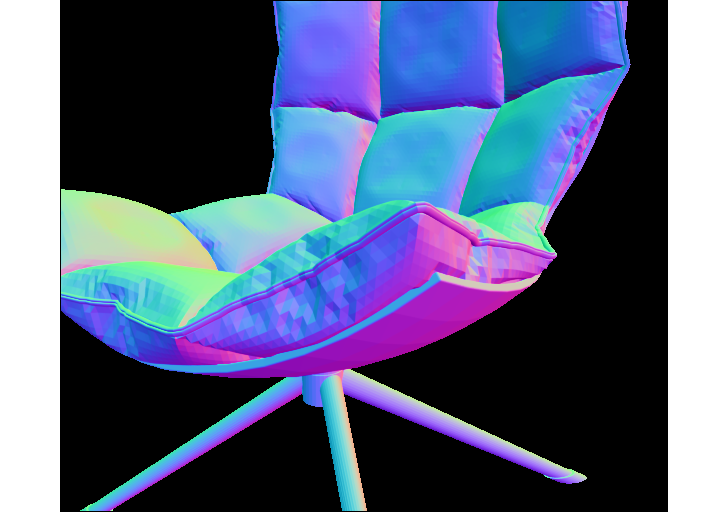} &
\includegraphics[width=\linewidth]{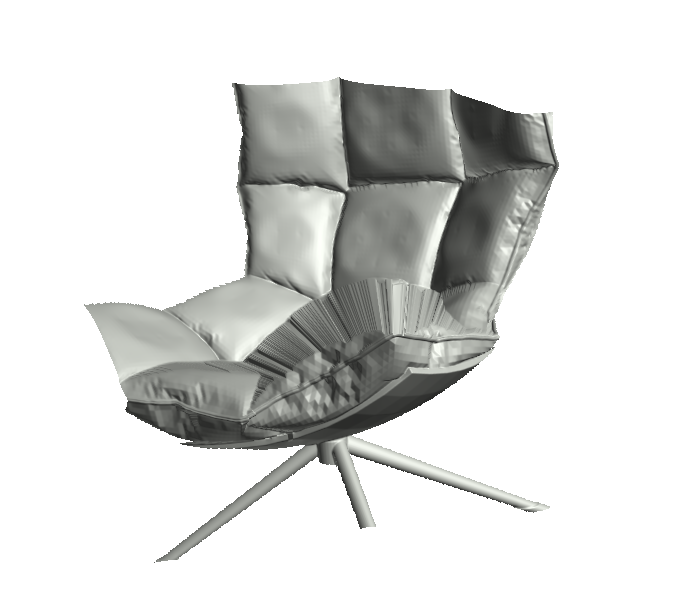} &
\includegraphics[width=\linewidth]{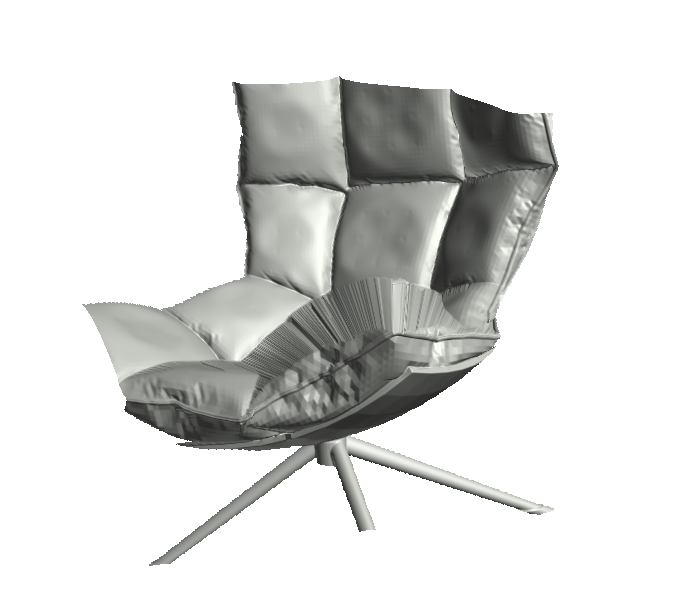} \\[-5pt]
&  \scriptsize{\textbf{RE}: $8.20$, \textbf{ERA}: $7.04$} & \scriptsize{\textbf{RE}: $13.32$, \textbf{ERA}: $10.09$} & & & \scriptsize{\textbf{RE}: $8.35$, \textbf{ERA}: $6.85$} & \scriptsize{\textbf{RE}: $8.69$, \textbf{ERA}: $6.73$}\\
\includegraphics[width=\linewidth]{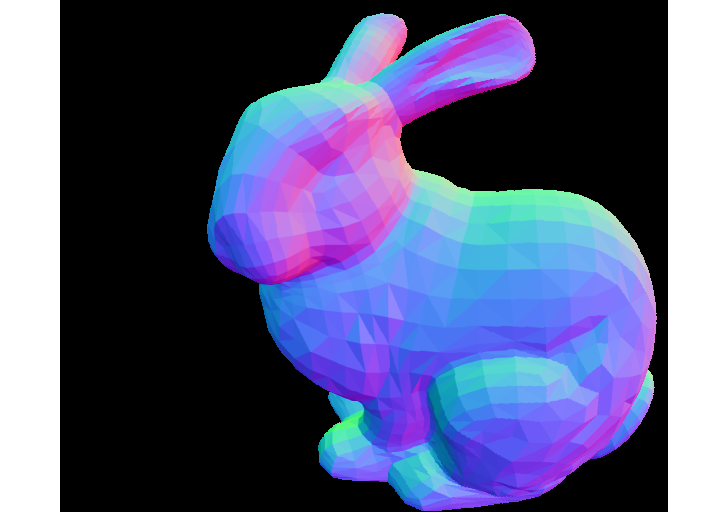} & 
\includegraphics[width=\linewidth]{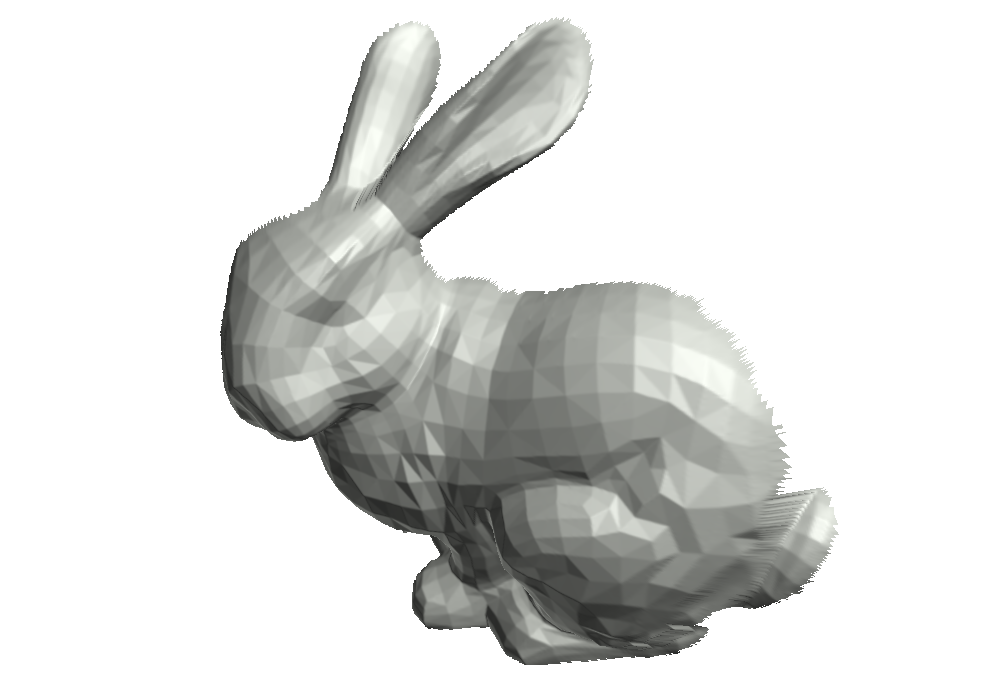} & 
\includegraphics[width=\linewidth]{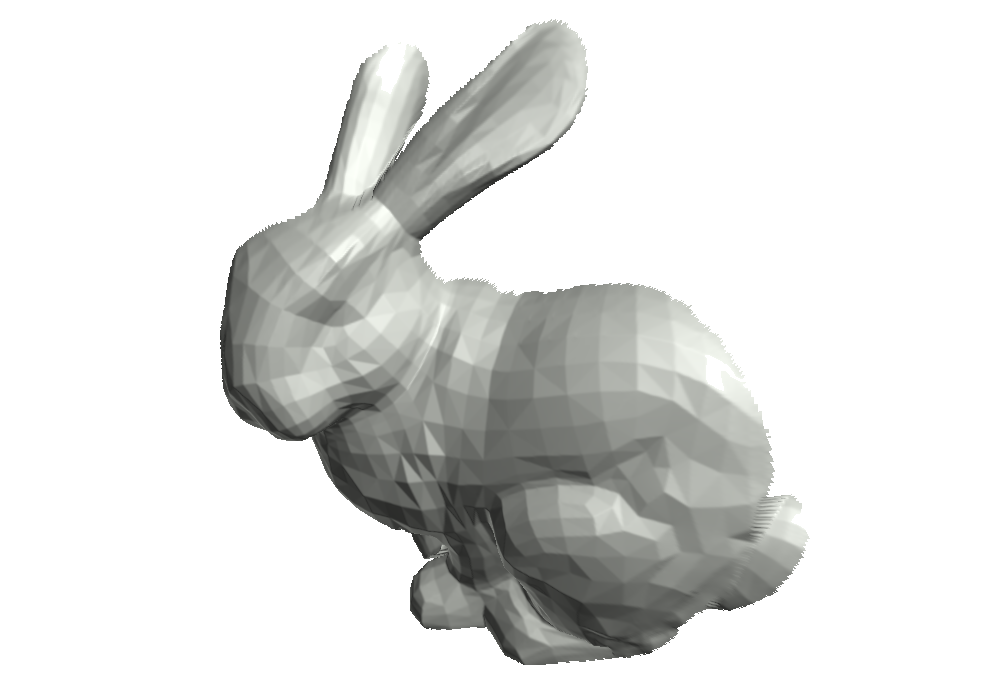} & 
& \includegraphics[width=\linewidth]{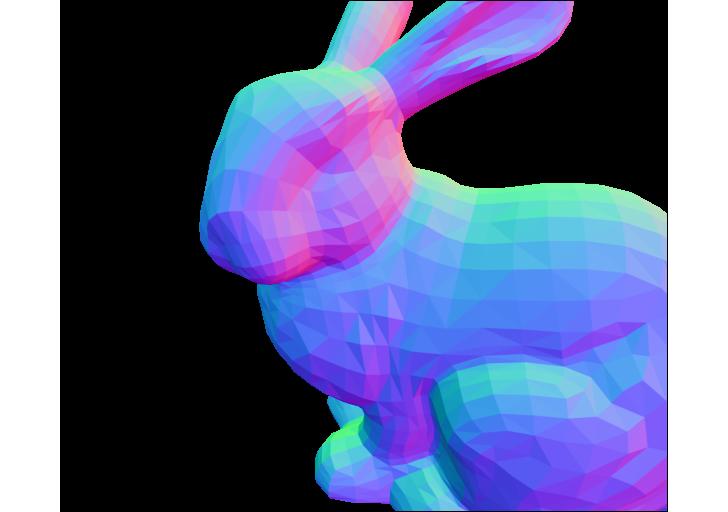} &
\includegraphics[width=\linewidth]{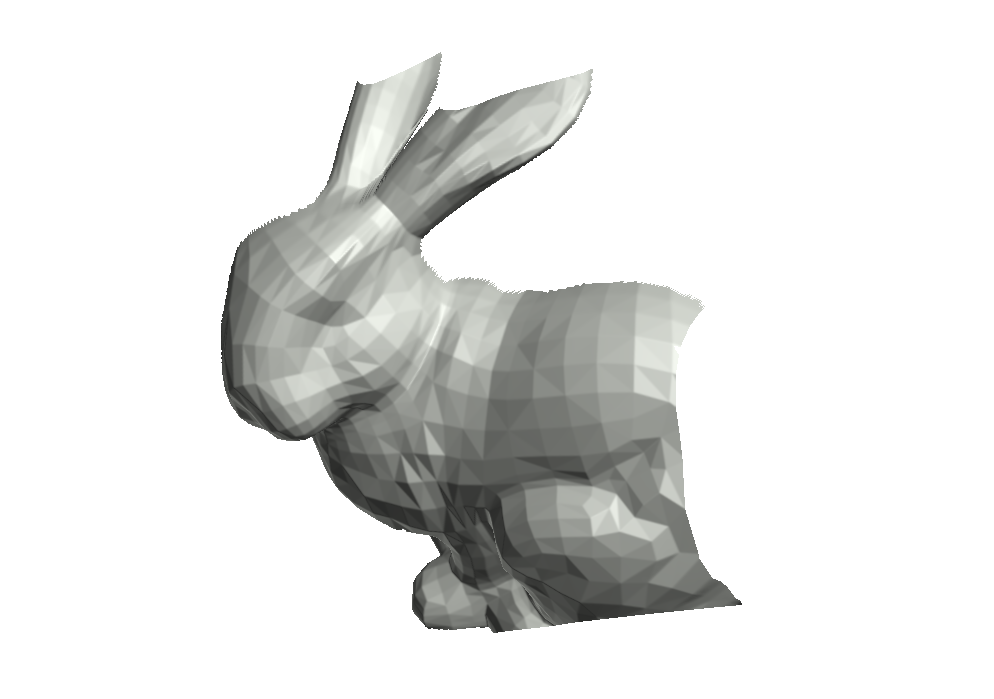} &
\includegraphics[width=\linewidth]{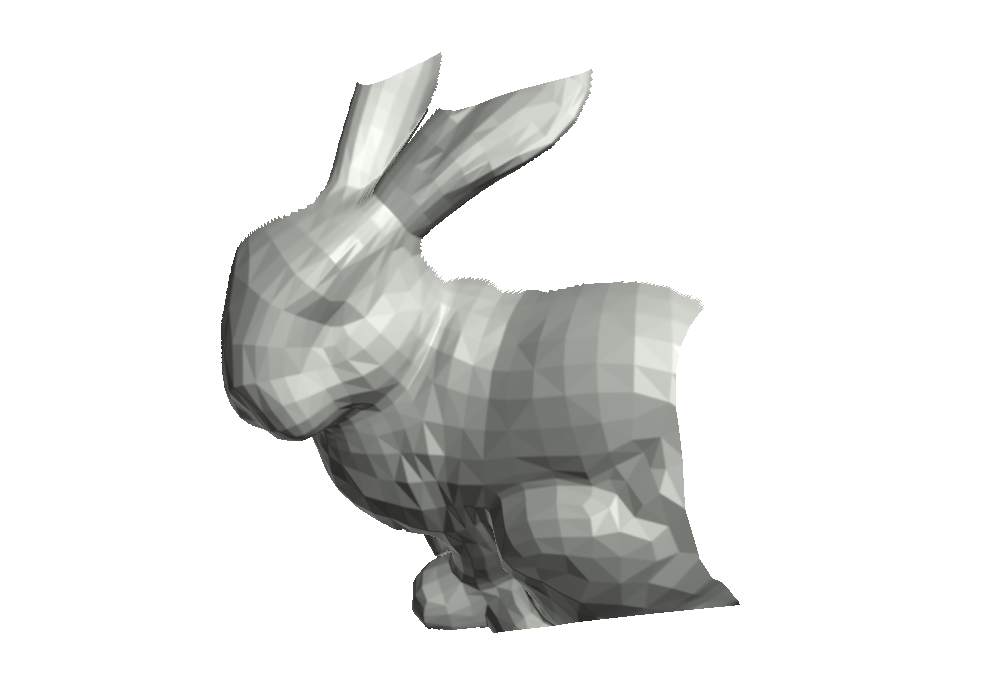} \\[-5pt]
&  \scriptsize{\textbf{RE}: $0.27$, \textbf{ERA}: $0.30$} & \scriptsize{\textbf{RE}: $1.16$, \textbf{ERA}: $1.29$} & & & \scriptsize{\textbf{RE}: $0.16$, \textbf{ERA}: $0.19$} & \scriptsize{\textbf{RE}: $0.17$, \textbf{ERA}: $0.21$}\\
\end{tabular}
\caption{\textbf{Reconstructions for non-ideal pinhole normal maps.} From the left, first three columns: input normal map with Brown-Conrady distortion, reconstruction of our method, and reconstruction of BiNI; last three columns: undistorted input normal map, reconstruction of our method, reconstruction of BiNI. Below each reconstruction are the corresponding mean relative depth error (\textbf{RE}) and mean depth error relative to the average depth of the scene (\textbf{ERA}), both expressed as percentages.
Note that undistortion causes part of the scene to be cropped. Source of the mesh models from the top to the bottom row:~\cite{MeshModernLivingRoom2023}, \cite{MeshOffice2021}, \cite{MeshOfficeChair2023} \cite{MeshStanfordBunny1994}.}
\vspace{-5pt}
\label{fig:visualizations_nonpinhole_camera_models}
\end{figure*}

To verify the applicability of our method to generic central camera models, we synthetically render normal maps (and depth for evaluation) observed by a pinhole camera with Brown-Conrady lens distortion~\cite{Brown1966DecenteringDistortionLenses}, using
BlenderProc~\cite{Denninger2023BlenderProc}.
Since to our knowledge no other methods are available that can directly handle normals from non-ideal pinhole cameras, we show, for
illustration purposes only,
the output of BiNI for such distorted maps; we remark that a quantitative evaluation is unfair, since
BiNI
assumes normals from an ideal pinhole camera. We additionally render the normal and depth maps observed by an ideal pinhole camera with intrinsics resulting from undistortion and resolution matching the original one. As shown in \cref{fig:visualizations_nonpinhole_camera_models}, our method effectively handles the case with lens distortion both for scene-level maps of medium complexity and for object-level ones, while the reconstruction from BiNI 
suffers from
noticeable distortion, as expected.
The undistorted reconstructions show comparable results between the two methods, with slightly better accuracy for
ours,
but at the cost of a reduced field of view, 
due to
barrel
distortion.

\subsection{Experiments with real-world data~\label{sec:experiments_real_world_data}}
\begin{figure}[!t]
\centering
\def\colwidth{0.23\linewidth}
\def\minicolwidth{0.01\textwidth}
\newcolumntype{M}[1]{>{\centering\arraybackslash}m{#1}}
\addtolength{\tabcolsep}{-4pt}
\begin{tabular}{M{\colwidth} M{\colwidth} M{\colwidth}  M{\colwidth}  }
\includegraphics[width=\linewidth]{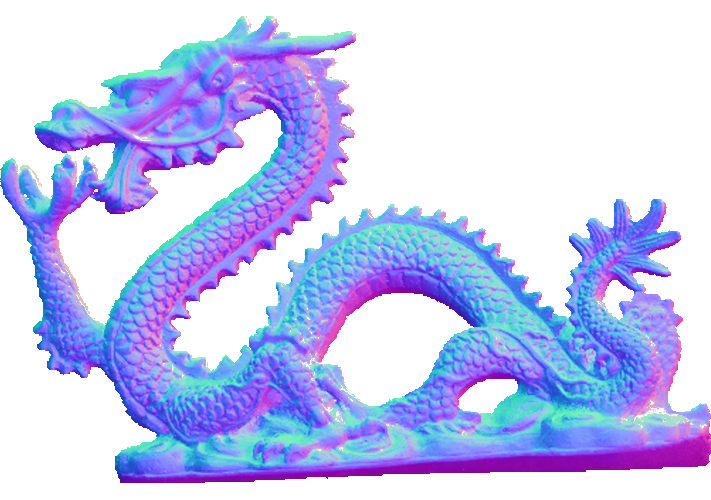} & 
\includegraphics[width=\linewidth]{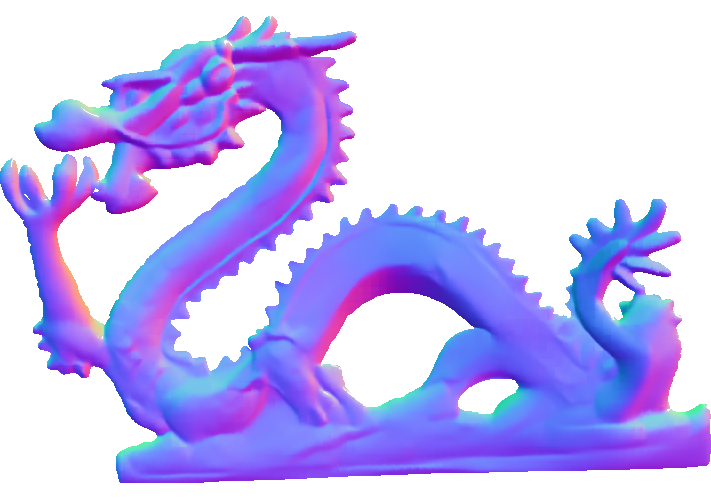} & 
\includegraphics[width=\linewidth]{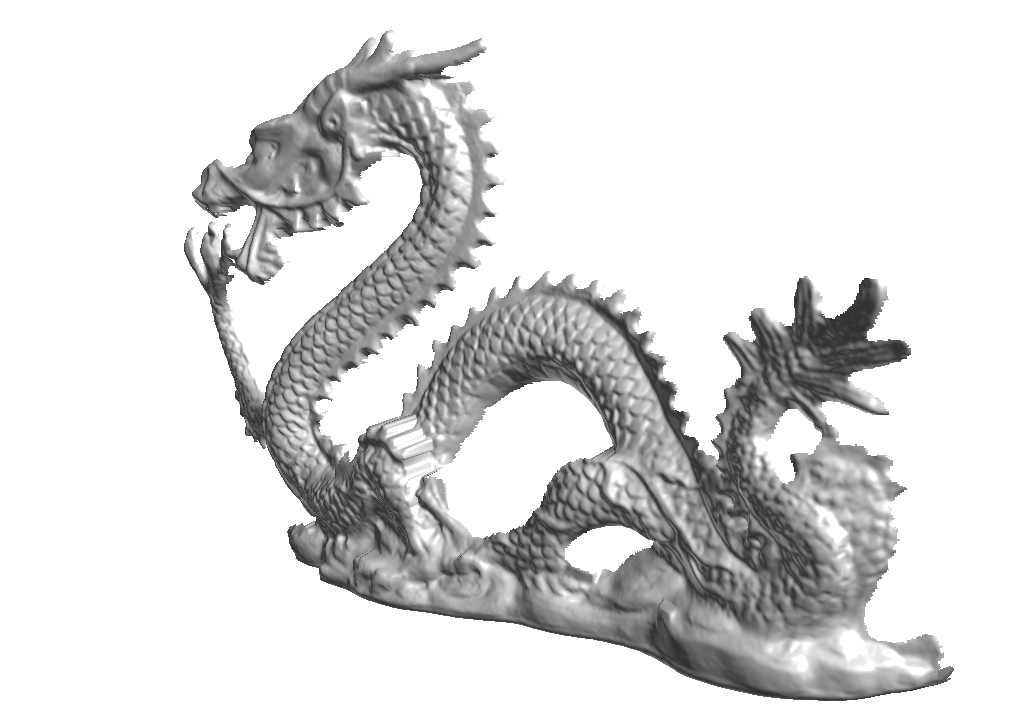} & 
\includegraphics[width=\linewidth]{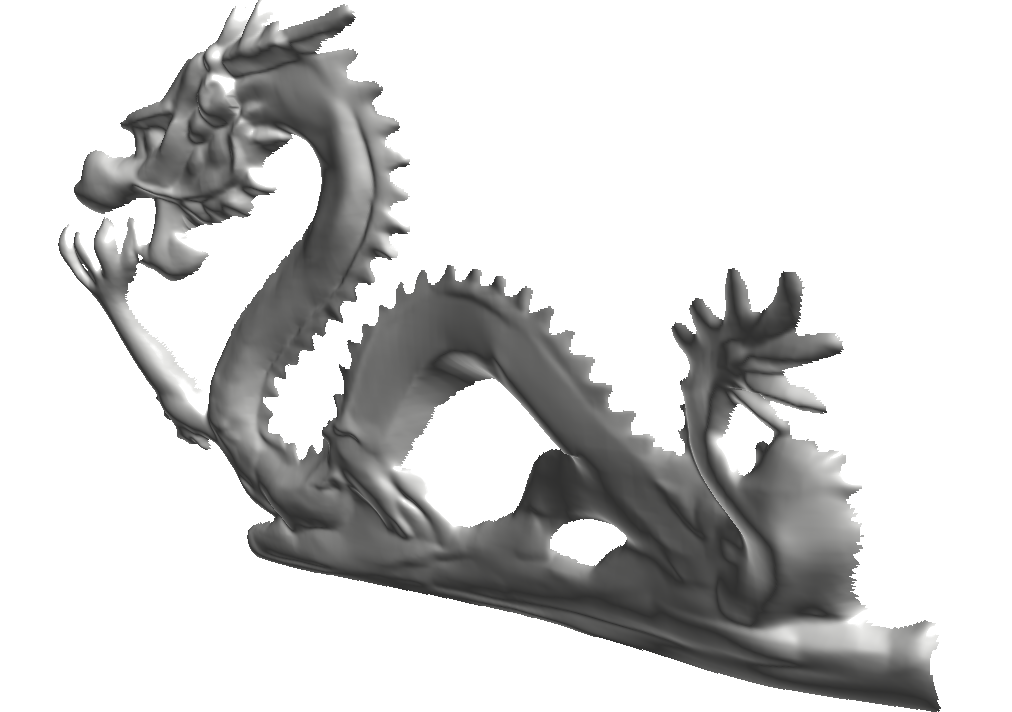} \\
\includegraphics[width=\linewidth]{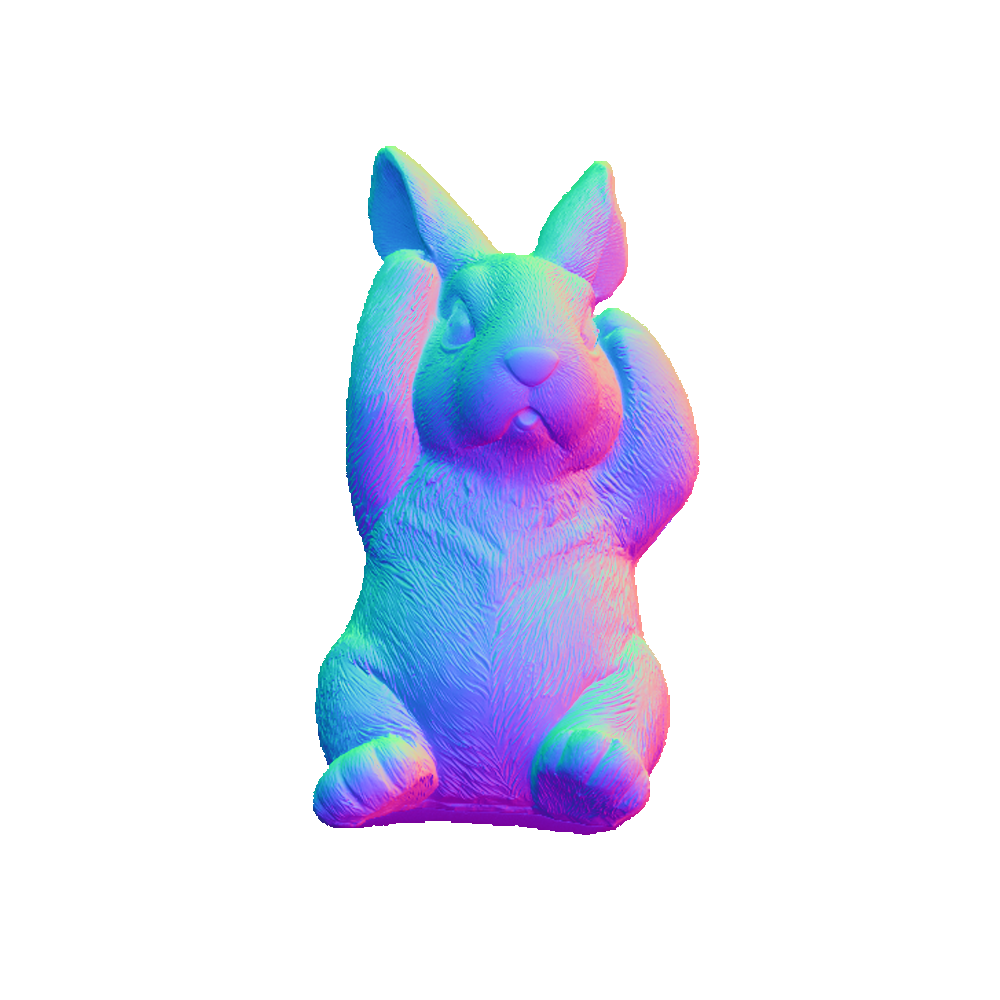} & 
\includegraphics[width=\linewidth]{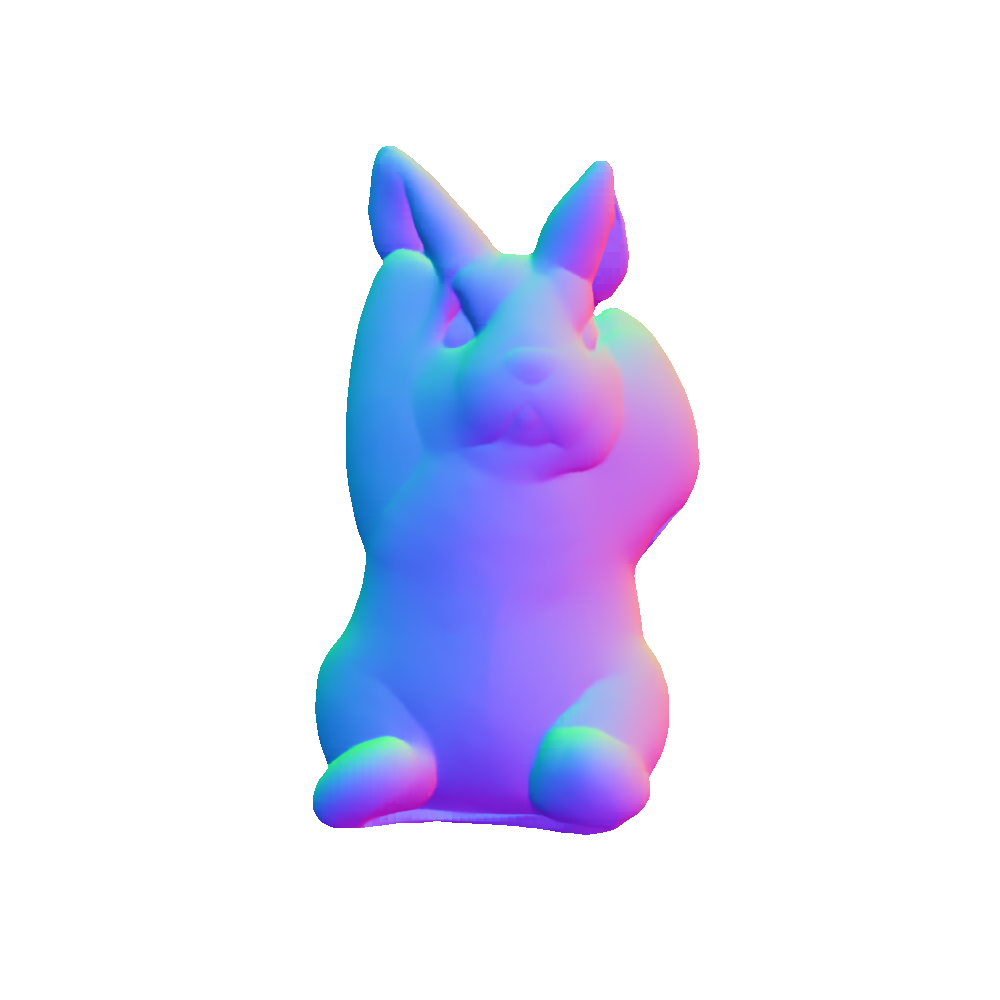} & 
\includegraphics[width=\linewidth]{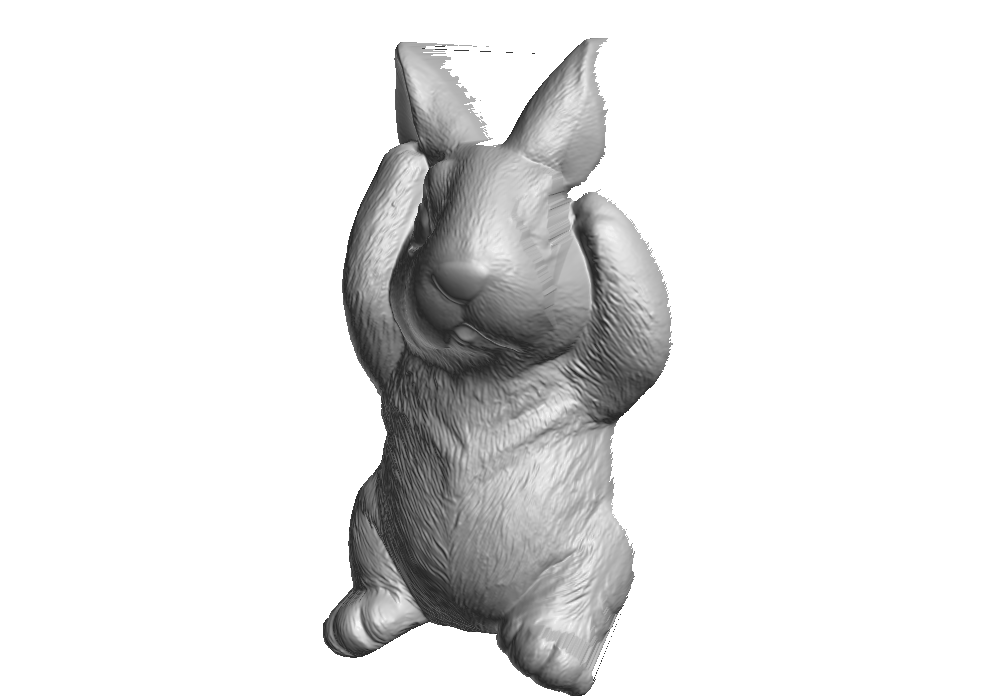} & 
\includegraphics[width=\linewidth]{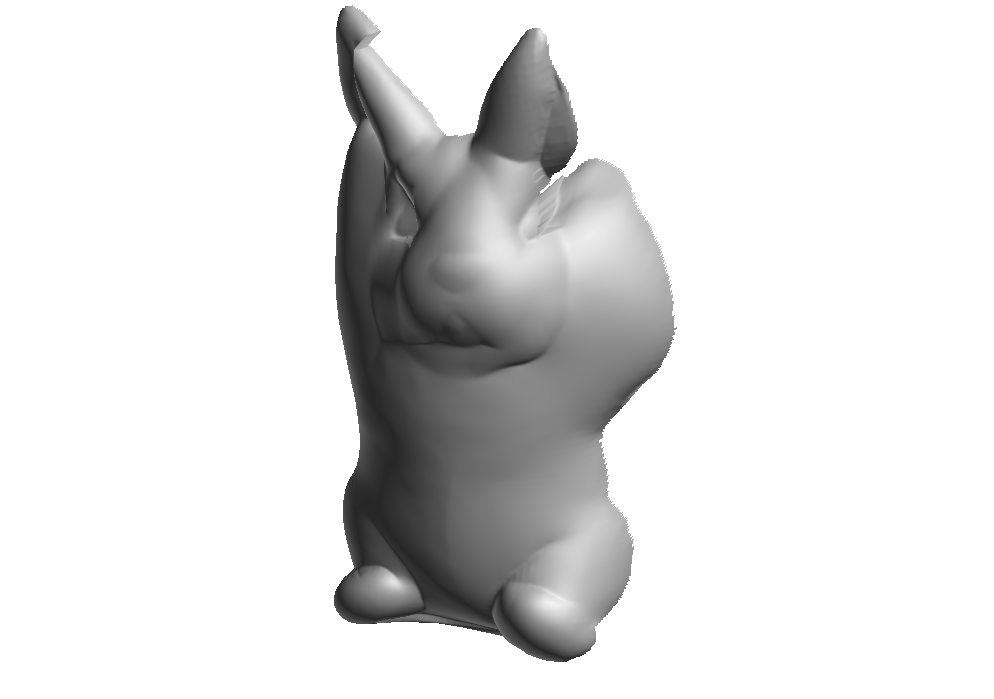} \\
\includegraphics[width=\linewidth]{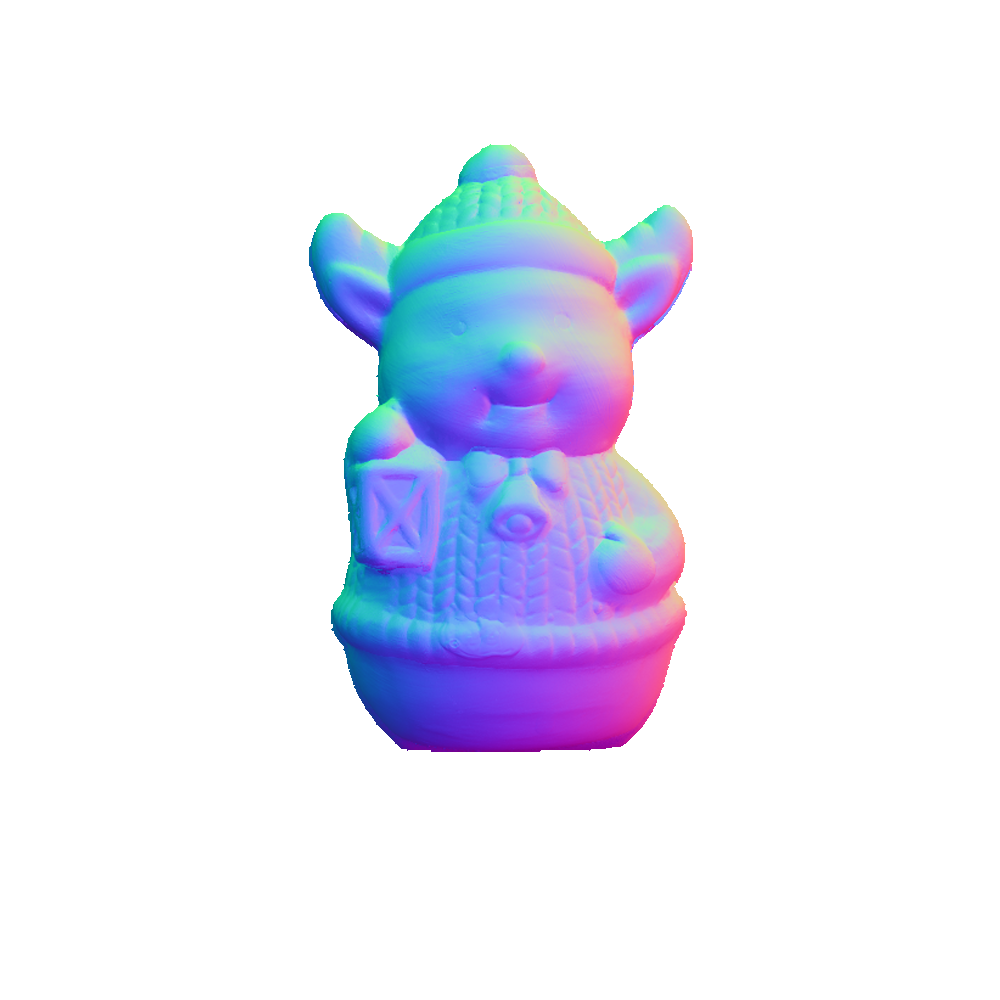} & 
\includegraphics[width=\linewidth]{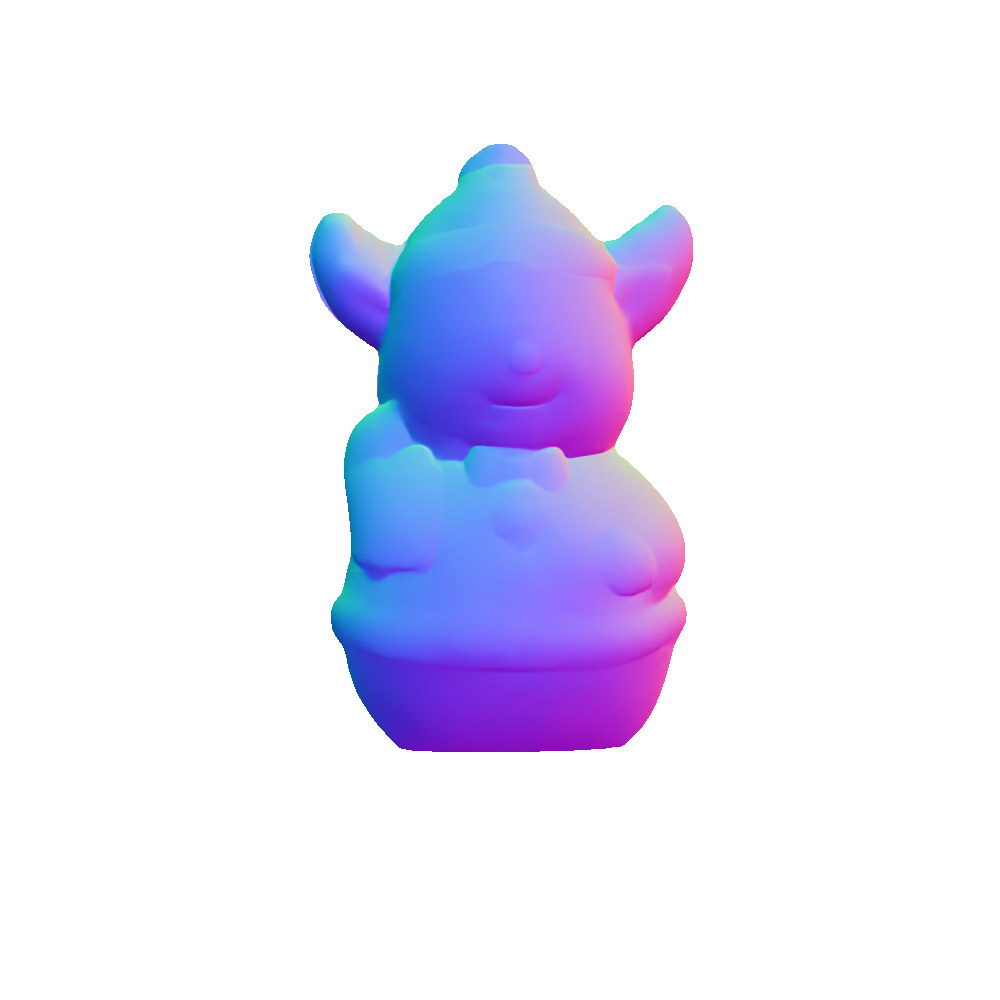} & 
\includegraphics[width=\linewidth]{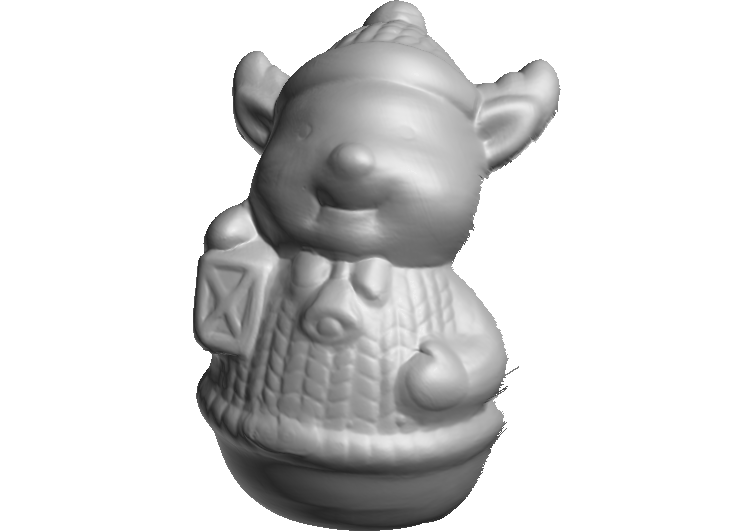} & 
\includegraphics[width=\linewidth]{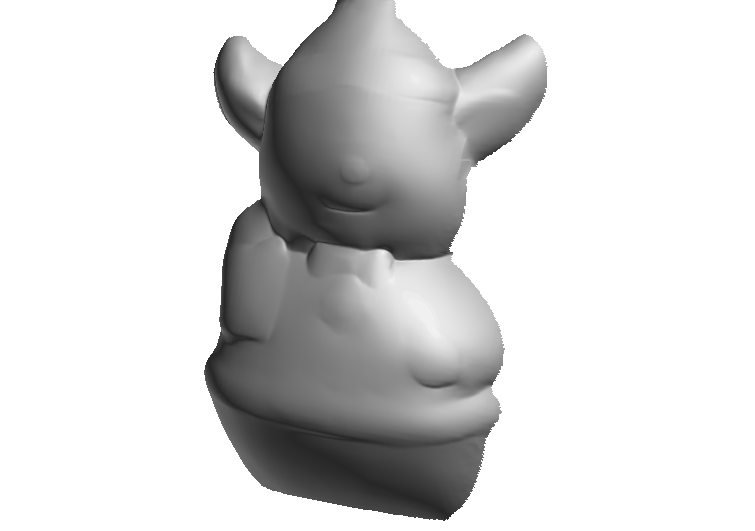} \\[-10pt]
\end{tabular}
\caption{\textbf{Reconstructions of our method from real-world data.} From the left, the third and fourth column show our reconstruction based on normals respectively from photometric stereo~\cite{Ikehata2023SDMUniPS} (first column) and from DSINE~\cite{Bae2024DSINE} (second column).}
\vspace{-10pt}
\label{fig:real_world_tests}
\end{figure}
\Cref{fig:real_world_tests} shows qualitative examples of the reconstructions produced by our method using normals obtained from real-world images~\cite{Ikehata2023SDMUniPS}, both through a recent photometric stereo approach~\cite{Ikehata2023SDMUniPS} and through prediction by a state-of-the-art learning-based normal estimation method~\cite{Bae2024DSINE}. The results
indicate
that our method
can
be applied effectively
to real-world normal maps, producing reasonably accurate reconstructions also for the overly smooth normals of~\cite{Bae2024DSINE}.

\begin{table*}[!ht]
    \centering
    \resizebox{\linewidth}{!}{
    \begin{tabular}{lccccccccc}
    \toprule
    Method & \texttt{bear} & \texttt{buddha} & \texttt{cat} & \texttt{cow} & \texttt{harvest} & \texttt{pot1} & \texttt{pot2} & \texttt{reading} & \texttt{goblet}\\
    \midrule
    BiNI~\cite{Cao2022BiNI} & $(3.72 \pm 2.71) \times 10^{-1}$ & $(4.57 \pm 9.21) \times 10^{-1}$ & $(0.54 \pm 1.09) \times 10^{0}$ & $(4.46 \pm 3.58) \times 10^{-1}$ & $(0.52 \pm 2.71) \times 10^{0}$ & $(4.18 \pm 5.63) \times 10^{-1}$ & $(3.96 \pm 4.26) \times 10^{-1}$ & $(0.50 \pm 1.24) \times 10^{0}$ & $(0.43 \pm 1.33) \times 10^{0}$\\
    Ours & $(0.82 \pm 7.39) \times 10^{-2}$ & $(0.90 \pm 9.12) \times 10^{-1}$ & $(0.03 \pm 1.27) \times 10^{0}$ & $(0.19 \pm 1.61) \times 10^{-1}$ & $(0.22 \pm 2.80) \times 10^{0}$ & $(0.09 \pm 2.72) \times 10^{0}$ & $(0.39 \pm 3.11) \times 10^{-1}$ & $(0.08 \pm 1.21) \times 10^{0}$ & $(0.06 \pm 1.20) \times 10^{0}$\\
    \bottomrule
    \end{tabular}
    }
    \caption{\textbf{Absolute formulation accuracy on the ground-truth log-depth map, DiLiGenT dataset~\cite{Shi2016DiLiGenT}}. For both methods, we report mean and standard deviation across the pixels of the absolute residual $|\gamma_{b\rightarrow a}(\tilde{z}_a - \tilde{z}_b) - \mathrm{RHS}|$ computed on the ground-truth log-depth map, where $\mathrm{RHS}$ denotes the right-hand side of \eqref{eq:bini_formulation} for BiNI and \eqref{eq:ours_formulation_log_with_bini_factor} for Ours. We use $\boldsymbol{\tau_m} = (\boldsymbol{\tau_a} + \boldsymbol{\tau_b}) / 2$ and $\alpha_{b\rightarrow a}=0$ for Ours.}
    \label{tab:comparison_formulation_accuracy}
\end{table*}
\begin{table*}[!ht]
    \centering
    \resizebox{0.95\linewidth}{!}{
    \begin{tabular}{lccccccccc}
    \toprule
    Method & \texttt{bear} & \texttt{buddha} & \texttt{cat} & \texttt{cow} & \texttt{harvest} & \texttt{pot1} & \texttt{pot2} & \texttt{reading} & \texttt{goblet}\textsuperscript{*}\\
    \midrule
    BiNI~\cite{Cao2022BiNI} -- From paper & $0.49$ & $0.86$ & $0.11$ & $\underline{0.07}$ & $2.73$ & $0.62$ & $0.22$ & $0.34$ & $8.53$\\
    BiNI~\cite{Cao2022BiNI} -- From code~\cite{Cao2022BiNI_code}, $\num{1200}$ iterations~${}^\dagger$ & $0.33$ & $1.06$ & $\underline{0.07}$ & $\mathbf{0.06}$ & $1.84$ & $0.64$ & $0.22$ & $0.26$ & $9.00$\\
    Kim \etal~\cite{Kim2024DiscontinuityPreserving} & $0.45$ & $0.67$ & $0.24$ & $\mathbf{0.06}$ & $2.44$ & $0.57$ & $0.19$ & $0.15$ & $9.02$\\
    \arrayrulecolor{gray!70}\specialrule{0.2pt}{0.2pt}{0.2pt}
    \arrayrulecolor{black}
    Ours w/o $\alpha_{b\rightarrow a}$ computation, $\num{150}$ iterations & $0.08$ & $0.30$ & $\textbf{0.06}$ & $0.09$ &  $4.98$ & $0.52$ & $\textbf{0.13}$ & $0.21$ &  $6.46$\\
    Ours w/o $\alpha_{b\rightarrow a}$ computation, $\num{1200}$ iterations & $\underline{0.07}$ & $\underline{0.26}$ & $\textbf{0.06}$ & $0.08$ & $4.83$ & $\underline{0.50}$ & $\textbf{0.13}$ & $\textbf{0.12}$ & $6.56$ \\
    Ours, $\num{150}$ iterations & $\textbf{0.03}$ & $0.37$ & $\textbf{0.06}$ & $0.08$ & $\underline{1.35}$ & $\underline{0.50}$ & $\underline{0.14}$ & $\underline{0.15}$ & $\underline{5.98}$ \\
    Ours, $\num{1200}$ iterations & $\textbf{0.03}$ & $\textbf{0.24}$ & $\textbf{0.06}$ & $0.08$ & $\textbf{0.73}$ & $\textbf{0.49}$ & $\textbf{0.13}$ & $0.17$ & $\textbf{4.72}$ \\
    \arrayrulecolor{gray!70}\specialrule{0.2pt}{0.2pt}{0.2pt}
    \arrayrulecolor{black}
    Ours with known
    discontinuity values
    & $0.01$ & $0.10$ & $0.03$ & $<10^{-2}$ & $0.34$ & $0.04$ & $0.03$ & $0.08$ & $0.11$\\
    \bottomrule
    \end{tabular}
    }
    \caption{\textbf{Mean absolute depth error (MADE) [$\boldsymbol{\si{mm}}$] on the DiLiGenT benchmark~\cite{Shi2016DiLiGenT}.} For each object, \textbf{bold} and \underline{underlined} denote respectively the best and the second-best result across the methods. The results of Kim~\etal are taken from~\cite{Kim2024DiscontinuityPreserving}. \textsuperscript{*}This object contains a full depth discontinuity. {${}^{\dagger}$}~BiNI achieves full convergence already after $\num{150}$ iterations.}
    \label{tab:diligent_results_main}
\end{table*}
%
\section{Conclusions\label{sec:conclusions}}
We presented a novel formulation for normal integration based on a local planarity assumption modeled through ray directions and explicit discontinuity terms. Compared to existing methods, our approach more accurately approximates the relation between depth and surface normals and achieves state-of-the-art results on the standard benchmark
for normal integration. 
Furthermore,
thanks to
its formulation based on ray
directions,
our method
allows for the first time handling
normals from
generic central cameras.

\noindent\textbf{Acknowledgements.}
The authors
thank
Lionel Ott and Yujie Wei for their feedback on the manuscript draft and the anonymous reviewers for their constructive comments.

{
    \small
    \bibliographystyle{ieeenat_fullname}
    \bibliography{main}
}

\cleardoublepage
\appendix
\section*{Supplementary Material}
The Supplementary Material is organized as follows. In \cref{sec_suppl:derivation_our_formulation}, we derive the mathematical formulation at the core of our method. In \cref{sec_suppl:analysis_gamma_factor}, we provide a novel analysis of the multiplicative factor $\gamma_{b\rightarrow a}$ used by BiNI~\cite{Cao2022BiNI} and extended in our method, and provide important insights on its effect on convergence. 
\Cref{sec_suppl:analysis_positivity_log_term} provides additional insights on the positivity of the $\log$ term in our formulation (\eqref{eq:ours_relationship_equation_t} in the main
paper), including
a mathematical proof that this property is preserved throughout the
optimization,
and discusses corner cases.
In \cref{sec_suppl:ablation_lambda_m}, we study the impact of the choice of the ray direction vector $\boldsymbol{\tau_m}$, that controls our local planarity assumption. 
In \cref{sec_suppl:ablation_beta_a_from_b_discont_activation_term}, we study the effect of the discontinuity activation term $\beta_{b\rightarrow a}^{(t)}$ in our formulation.
\Cref{sec_suppl:ablation_connectivity} presents an ablation on different pixel connectivity. 
\Cref{sec_suppl:additional_evaluations_formulation_accuracy} presents an evaluation of the formulation accuracy with metrics in addition to the one introduced in \cref{sec:comparison_accuracy_formulation}.
\Cref{sec_suppl:results_for_noisy_inputs} provides results of our method under noisy input normals.
\Cref{sec_suppl:additional_evaluations} provides an evaluation on the DiLiGenT-MV dataset~\cite{Li2020DiLiGenT_MV}, which extends the DiLiGenT dataset.
Finally,
\Cref{sec_suppl:limitations} discusses the limitations of our method.

\section{Derivation of our formulation~\label{sec_suppl:derivation_our_formulation}}
In the following Section, we provide a derivation of the coefficients~\eqref{eq:ours_coefficient_definition} of our formulation~\eqref{eq:ours_formulation}. Rearranging the equations in the system~\eqref{eq:system_equations_ours} emerging from our local planarity assumption and using $x_b = \tau_{x_b}z_b$, $y_b = \tau_{y_b}z_b$ (by definition of $\tau_{x_b}$, $\tau_{y_b}$) yields the following linear system in the variables $dx_{ma}$, $dy_{ma}$, $dz_{ma}$, $dx_{mb}$, $dy_{mb}$, $dz_{mb}$:
\begin{equation}
    \mathbf{C}\cdot
    \begin{bmatrix}
        dx_{ma} \\
        dy_{ma} \\
        dz_{ma} \\
        dx_{mb} \\
        dy_{mb} \\
        dz_{mb} \\
    \end{bmatrix}=
    \mathbf{d},
    \label{eq:sys_equations_deltas}
\end{equation}

where

\begin{align}
\begin{split}
    \mathbf{C} &=\begin{bmatrix}
        0 & 0 & 0 & 1 & 0 & -\tau_{x_m}\\
        0 & 0 & 0 & 0 & 1 & -\tau_{y_m}\\
        -1 & 0 & \tau_{x_a} & 1 & 0 & -\tau_{x_a}\\
        0 & -1 & \tau_{y_a} & 0 & 1 & -\tau_{y_a}\\
        0 & 0 & 0 & n_{bx} & n_{by} & n_{bz}\\
        n_{ax} & n_{ay} & n_{az} & 0 & 0 & 0\\
    \end{bmatrix},\ \mathrm{and}\\
    \mathbf{d} &= \begin{bmatrix}
        (\tau_{x_m} - \tau_{x_b})z_b\\
        (\tau_{y_m} - \tau_{y_b})z_b\\
        (\tau_{x_a} - \tau_{x_b})z_b\\
        (\tau_{y_a} - \tau_{y_b})z_b\\
        0\\
        -n_{az}\varepsilon_{b\rightarrow a}
    \end{bmatrix}.\\
\end{split}
\end{align}

Solving~\eqref{eq:sys_equations_deltas} yields the following expressions for $dz_{ma}$ and $dz_{mb}$:
\begin{align}
    \begin{split}
    \small{dz_{ma}} = &\small{\frac{-n_{az}}{\boldsymbol{n_a}^\mathsf{T} \boldsymbol{\tau_a}}\cdot\varepsilon_{b\rightarrow a}}\ +\\
        & \small{\frac{(n_{ax}\tau_{x_a} + n_{ay}\tau_{y_a} -n_{ax}\tau_{x_m} -n_{ay}\tau_{y_m})\cdot\boldsymbol{n_b}^\mathsf{T} \boldsymbol{\tau_b}}{\boldsymbol{n_a}^\mathsf{T} \boldsymbol{\tau_a}\cdot\boldsymbol{n_b}^\mathsf{T} \boldsymbol{\tau_m}}\cdot z_b}\\
        =&\small{\frac{-n_{az}}{\boldsymbol{n_a}^\mathsf{T} \boldsymbol{\tau_a}}\cdot\varepsilon_{b\rightarrow a}+\frac{(\boldsymbol{n_a}^\mathsf{T} \boldsymbol{\tau_a} - \boldsymbol{n_a}^\mathsf{T} \boldsymbol{\tau_m})\cdot\boldsymbol{n_b}^\mathsf{T} \boldsymbol{\tau_b}}{\boldsymbol{n_a}^\mathsf{T} \boldsymbol{\tau_a}\cdot\boldsymbol{n_b}^\mathsf{T} \boldsymbol{\tau_m}}\cdot z_b},\\
        \small{dz_{mb}} =&\small{\frac{n_{bx}\tau_{x_b} + n_{by}\tau_{y_b} - n_{bx}\tau_{x_m} - n_{by}\tau_{y_m}}{\boldsymbol{n_b}^\mathsf{T} \boldsymbol{\tau_m}}\cdot z_b}\\
        =&\small{\frac{\boldsymbol{n_a}^\mathsf{T} \boldsymbol{\tau_a}\cdot(\boldsymbol{n_b}^\mathsf{T} \boldsymbol{\tau_b} - \boldsymbol{n_b}^\mathsf{T} \boldsymbol{\tau_m})}{\boldsymbol{n_a}^\mathsf{T} \boldsymbol{\tau_a}\cdot \boldsymbol{n_b}^\mathsf{T} \boldsymbol{\tau_m}}\cdot z_b}.
    \end{split}
\end{align}

The final step to obtain our formulation~\eqref{eq:ours_formulation},~\eqref{eq:ours_coefficient_definition} follows from writing:
\begin{align}
\begin{split}
    \small{z_a=}&\small{z_b + dz_{mb} - dz_{ma}}\\
    \small{=}&\small{\frac{\boldsymbol{n_a}^\mathsf{T} \boldsymbol{\tau_a}\cdot \boldsymbol{n_b}^\mathsf{T}\boldsymbol{\tau_m}}{\boldsymbol{n_a}^\mathsf{T} \boldsymbol{\tau_a}\cdot \boldsymbol{n_b}^\mathsf{T} \boldsymbol{\tau_m}}\cdot z_b\ +} \\
    &\small{\frac{\boldsymbol{n_a}^\mathsf{T} \boldsymbol{\tau_a}\cdot(\boldsymbol{n_b}^\mathsf{T} \boldsymbol{\tau_b} - \boldsymbol{n_b}^\mathsf{T} \boldsymbol{\tau_m})}{\boldsymbol{n_a}^\mathsf{T} \boldsymbol{\tau_a}\cdot \boldsymbol{n_b}^\mathsf{T} \boldsymbol{\tau_m}}\cdot z_b\ +}\\
    &\small{\frac{-(\boldsymbol{n_a}^\mathsf{T} \boldsymbol{\tau_a} - \boldsymbol{n_a}^\mathsf{T} \boldsymbol{\tau_m})\cdot\boldsymbol{n_b}^\mathsf{T} \boldsymbol{\tau_b}}{\boldsymbol{n_a}^\mathsf{T} \boldsymbol{\tau_a}\cdot \boldsymbol{n_b}^\mathsf{T} \boldsymbol{\tau_m}}\cdot z_b\ +}\\
    &\small{\frac{n_{az}}{\boldsymbol{n_a}^\mathsf{T} \boldsymbol{\tau_a}}\cdot\varepsilon_{b\rightarrow a}}\\
    \small{=}&\small{\frac{n_{az}}{\boldsymbol{n_a}^\mathsf{T} \boldsymbol{\tau_a}}\cdot\varepsilon_{b\rightarrow a}+\frac{\boldsymbol{n_a}^\mathsf{T} \boldsymbol{\tau_m}\cdot\boldsymbol{n_b}^\mathsf{T} \boldsymbol{\tau_b}}{\boldsymbol{n_a}^\mathsf{T} \boldsymbol{\tau_a}\cdot\boldsymbol{n_b}^\mathsf{T} \boldsymbol{\tau_m}}\cdot z_b.}
\end{split}
\end{align}

\paragraph{Alternative derivation.} An alternative, more concise derivation\footnote{We thank the anonymous reviewer \texttt{NayZ} for suggesting this alternative derivation.} can be obtained by noting that the perpendicularity constraints encoded by the last two equations in \eqref{eq:system_equations_ours}
can be more compactly expressed as
\begin{align}
        &\boldsymbol{n_a}^\mathsf{T}(\boldsymbol{p_m} + \boldsymbol{\varepsilon_z} - \boldsymbol{p_a}) = 0 \label{eq:alt_derivation_formulation_eq_1}\\
        &\boldsymbol{n_b}^\mathsf{T}(\boldsymbol{p_m} - \boldsymbol{p_b}) = 0, \label{eq:alt_derivation_formulation_eq_2}
\end{align}
where $\boldsymbol{\varepsilon_z}\coloneq (0, 0, \varepsilon_{b\rightarrow a})^\mathsf{T}$. From \eqref{eq:alt_derivation_formulation_eq_2} it follows that 
\begin{equation}
    \frac{\boldsymbol{n_b}^\mathsf{T}\boldsymbol{p_b}}{\boldsymbol{n_b}^\mathsf{T}\boldsymbol{p_m}} = 1.
    \label{eq:alt_derivation_formulation_consequence_eq_2}
\end{equation}
Expanding \eqref{eq:alt_derivation_formulation_eq_1} and multiplying its first term by $1$ using the equivalence \eqref{eq:alt_derivation_formulation_consequence_eq_2} yields
\begin{equation}
    \frac{\boldsymbol{n_a}^\mathsf{T}\boldsymbol{p_m}\cdot\boldsymbol{n_b}^\mathsf{T}\boldsymbol{p_b}}{\boldsymbol{n_b}^\mathsf{T}\boldsymbol{p_m}} + \boldsymbol{n_a}^\mathsf{T}\boldsymbol{\varepsilon_z} - \boldsymbol{n_a}^\mathsf{T}\boldsymbol{p_a} = 0.
    \label{eq:alt_derivation_formulation_eq_1_rewritten_ver_1}
\end{equation}
Using $\boldsymbol{p_i}=z_i\boldsymbol{\tau_i}, i\in\{a, b, m\}$ (by definition) and the fact that $\boldsymbol{n_a}^\mathsf{T}\boldsymbol{\varepsilon_z} = n_{az}\cdot\varepsilon_{b\rightarrow a}$, \eqref{eq:alt_derivation_formulation_eq_1_rewritten_ver_1} can be rewritten as
\begin{equation}
    \frac{\boldsymbol{n_a}^\mathsf{T}\boldsymbol{\tau_m}\cdot\boldsymbol{n_b}^\mathsf{T}\boldsymbol{\tau_b}}{\boldsymbol{n_b}^\mathsf{T}\boldsymbol{\tau_m}}z_b + n_{az}\cdot\varepsilon_{b\rightarrow a} - (\boldsymbol{n_a}^\mathsf{T}\boldsymbol{\tau_a})z_a = 0.
    \label{eq:alt_derivation_formulation_eq_1_rewritten_ver_2}
\end{equation}
Dividing all terms in \eqref{eq:alt_derivation_formulation_eq_1_rewritten_ver_2} by $\boldsymbol{n_a}^\mathsf{T}\boldsymbol{\tau_a}$ and rearranging yields our formulation \eqref{eq:ours_formulation},~\eqref{eq:ours_coefficient_definition}.

\section{Influence of the multiplicative factor $\gamma_{b\rightarrow a}$~\label{sec_suppl:analysis_gamma_factor}}
As noted in Sec.~2 of the Supplementary Material of BiNI~\cite{Cao2022BiNI_supplementary}, the coefficient $\gamma_{b\rightarrow a}$\footnote{Denoted as $\tilde{n}_z$ in~\cite{Cao2022BiNI_supplementary}.}, which we extend
in
our 
formulation, 
is crucial to achieving optimal convergence during
optimization. In particular, their formulation based on the functional $\gamma_{b\rightarrow a}(\tilde{z}_a - \tilde{z}_b) = \delta_{b\rightarrow a}$ (\eqref{eq:bini_formulation} in the main paper) performs significantly better than the one derived from the equivalent equation $\tilde{z}_a - \tilde{z}_b = \delta_{b\rightarrow a} / \gamma_{b\rightarrow a}$. Similarly, we find that our formulation $\gamma_{b\rightarrow a}(\tilde{z}_a - \tilde{z}_b) = \gamma_{b\rightarrow a}\log\left(\omega_{b\rightarrow a}+\omega_{\varepsilon_a}\cdot\alpha_{b\rightarrow a}\right)$ (\eqref{eq:ours_formulation_log_with_bini_factor} in the main paper) achieves significantly better convergence than the equivalent $\tilde{z}_a - \tilde{z}_b = \log\left(\omega_{b\rightarrow a}+\omega_{\varepsilon_a}\cdot\alpha_{b\rightarrow a}\right)$.

In the following, we provide below a novel analysis of this phenomenon in light of our generic formulation based on ray direction vectors, which allows rewriting $\gamma_{b\rightarrow a}$
as
\begin{equation}
    \gamma_{b\rightarrow a} = f\cdot\boldsymbol{n_a}^\mathsf{T} \boldsymbol{\tau_a},
    \label{eq:gamma_a_from_b_bini_formulation_2_suppl}
\end{equation}
where $f$ is the (fixed) focal length, which we generalize to the (pixel-pair specific) factor $\left\|\boldsymbol{u_b}-\boldsymbol{u_a}\right\|/ \left\|\boldsymbol{\tau_b}-\boldsymbol{\tau_a}\right\|$\footnote{Note that for an ideal pinhole camera with $f=f_x=f_y$ one has $\|\boldsymbol{u_b}-\boldsymbol{u_a}\| = \left\|\left(u_b - u_a, v_b -v_a\right)\right\|$ and $\|\boldsymbol{\tau_b}-\boldsymbol{\tau_a}\| = \left\|\left((u_b - u_a) / f, (v_b - v_a) / f, 0\right)\right\| = \|\boldsymbol{u_b}-\boldsymbol{u_a}\| / f$, from which one recovers $\left\|\boldsymbol{u_b}-\boldsymbol{u_a}\right\|/ \left\|\boldsymbol{\tau_b}-\boldsymbol{\tau_a}\right\| = f$.}.
All the supporting experiments in this Section are run on the DiLiGenT benchmark, for $\num{1200}$ iterations and for simplicity using our version without $\alpha_{b\rightarrow a}$ computation.

We start by noting that, for each pixel pair $(a, b$),
the coefficient
$\gamma_{b\rightarrow a}$
has two effects on the optimization:
\begin{itemize}
    \item \textbf{Effect 1 (weighting):}~On one side, it introduces a quadratic factor $\gamma_{b\rightarrow a}^2$ in the corresponding term of the optimization cost function $(\mathbf{\tilde{A}\tilde{z}}-\mathbf{\tilde{b}})^\mathsf{T}\mathbf{\tilde{W}}(\mathbf{\tilde{A}\tilde{z}}-\mathbf{\tilde{b}})$ (\cf.~\eqref{eq:general_optimization_problem} in the main paper), or equivalently in its
    associated
    normal equation 
    $\mathbf{\tilde{A}}^\mathsf{T}\mathbf{\tilde{W}\tilde{A}\tilde{z}} = \mathbf{\tilde{A}}^\mathsf{T}\mathbf{\tilde{W}\tilde{b}}$,
    since 
    both the rows of $\mathbf{\tilde{A}}$ and the corresponding elements of $\mathbf{\tilde{b}}$ are scaled by a factor $\gamma_{b\rightarrow a}$ (\cf.~\eqref{eq:bini_formulation} and~\eqref{eq:ours_formulation_log_with_bini_factor} in the main paper). In other words,
    the optimization cost function reads as 
    \begin{equation}
    \small
            (\mathbf{\tilde{A}\tilde{z}}-\mathbf{\tilde{b}})^\mathsf{T}\mathbf{\tilde{W}}(\mathbf{\tilde{A}\tilde{z}}-\mathbf{\tilde{b}}) = \sum_{(a, b)}w_{b\rightarrow a}^{\mathrm{BiNI}}\cdot\gamma^2_{b\rightarrow a}\cdot(\tilde{z}_a - \tilde{z}_b - \mathrm{RHS})^2,
            \label{eq:optimization_cost_function_with_gamma_a_from_b}
    \end{equation}
    where $\mathrm{RHS}$ is $\delta_{b\rightarrow a} / \gamma_{b\rightarrow a}$ for BiNI and $\log\left(\omega_{b\rightarrow a}+\omega_{\varepsilon_a}\cdot\alpha_{b\rightarrow a}\right)$ for Ours. Therefore, each residual is effectively scaled by $w_{b\rightarrow a}^{\mathrm{BiNI}}\cdot\gamma^2_{b\rightarrow a}$ rather than only by $w_{b\rightarrow a}^{\mathrm{BiNI}}$.
    \item \textbf{Effect 2 (sharpness of the bilateral weights):} On the other side, it impacts the
    magnitude
    of the bilateral weights
    $w_{b\rightarrow a}^{\mathrm{BiNI}}=\sigma_k(\mathrm{res}_{-b\rightarrow a}^2 - \mathrm{res}_{b\rightarrow a}^2)$,
    where $\mathrm{res}_{b\rightarrow a}\coloneq\gamma_{b\rightarrow a}\left(\tilde{z}_a - \tilde{z}_b\right)$    
    (see also~\eqref{eq:bini_weight_definition} in the main
    paper).
    Since from~\eqref{eq:gamma_a_from_b_bini_formulation_2_suppl} $\gamma_{b\rightarrow a}\approx\gamma_{-b\rightarrow a}$, with exact equality when $f$ is constant, it follows that 
    \begin{align}
        \begin{split}
            w_{b\rightarrow a}^{\mathrm{BiNI}}&=\sigma_k(\gamma_{b\rightarrow a}^2\cdot((\tilde{z}_a - \tilde{z}_b)^2 - (\tilde{z}_a - \tilde{z}_{-b})^2))\\
            &=\sigma_{k\cdot\gamma_{b\rightarrow a}^2}((\tilde{z}_a - \tilde{z}_b)^2 - (\tilde{z}_a - \tilde{z}_{-b})^2),
            \label{eq:bini_weight_with_gamma_a_from_b}
        \end{split}
    \end{align}
    \ie, $\gamma_{b\rightarrow a}^2$ can be subsumed into the parameter $k$ of the sigmoid $\sigma_k$. As a consequence,
    $\gamma_{b\rightarrow a}$
    controls the
    convergence of the bilateral weights, so that for fixed $\tilde{z}_a$ and $\tilde{z}_b$, a larger $\gamma^2_{b\rightarrow a}$
    causes
    smaller depth differences between the two sides to be detected as a 
    one-sided
    discontinuity, and smaller values result in a less sharp convergence.
    
\end{itemize}

Crucially, we observe that the effects of the two terms $f$ and $\boldsymbol{n_a}^\mathsf{T} \boldsymbol{\tau_a}$  in~\eqref{eq:gamma_a_from_b_bini_formulation_2_suppl} can be decoupled and summarized in the following two Propositions:

\begin{propositionbox}{1}{Effect of the term $f$}
    The term $f$ acts as a constant (or near constant, in the case of $f=\left\|\boldsymbol{u_b}-\boldsymbol{u_a}\right\|/ \left\|\boldsymbol{\tau_b}-\boldsymbol{\tau_a}\right\|$) that controls the
    sharpness of the bilateral weights $w_{b\rightarrow a}^{\mathrm{BiNI}}$.
\end{propositionbox}

\begin{propositionbox}{2}{Effect of the term $\boldsymbol{n_a}^\mathsf{T} \boldsymbol{\tau_a}$}
    The term $\boldsymbol{n_a}^\mathsf{T} \boldsymbol{\tau_a}$ 
    introduces
    an active weighting mechanism (in addition to $w_{b\rightarrow a}^{\mathrm{BiNI}}$)
    based on the collinearity between surface normals and ray directions, reducing the influence of pixel pairs close to a discontinuity.
\end{propositionbox}
We provide below arguments and empirical verifications supporting the above Propositions.

\begin{argument}[Argument for Proposition 1]
Since $f$ is constant (or approximately constant), it can be factored out of each term $\gamma_{b\rightarrow a}^2$ in the optimization cost function~\eqref{eq:optimization_cost_function_with_gamma_a_from_b}. Since multiplying the cost function by a constant factor does not affect
its
minimizing solution, it follows that the term $f$
is not an influencing factor for Effect~1 (weighting).
We verify this by running our method using $\gamma_{b\rightarrow a} = \boldsymbol{n_a}^\mathsf{T} \boldsymbol{\tau_a}$ in our cost function~\eqref{eq:optimization_cost_function_with_gamma_a_from_b} and $\gamma_{b\rightarrow a} = f\cdot\boldsymbol{n_a}^\mathsf{T} \boldsymbol{\tau_a}$ in the bilateral weights~\eqref{eq:bini_weight_with_gamma_a_from_b}. As expected, up to minimal differences that we attribute to machine precision, the results match those obtained when using the full factor $\gamma_{b\rightarrow a} = f\cdot\boldsymbol{n_a}^\mathsf{T} \boldsymbol{\tau_a}$ in the cost function (\cf. first and second row in \cref{tab:analysis_gamma_a_from_b}).

We verify that instead the term $f$ does indeed contribute to Effect~2 (sharpness of the bilateral weights) by varying its value in the $\gamma_{b\rightarrow a}$ factor of the bilateral weights, while maintaining a fixed $\gamma_{b\rightarrow a} = \boldsymbol{n_a}^\mathsf{T} \boldsymbol{\tau_a}$ in our cost function. Comparing rows $2$ to $5$ in \cref{tab:analysis_gamma_a_from_b} shows that indeed different values of $f$ result in different convergence; while the change is object-specific, the
main emerging trend appears to indicate that
worse convergence
is obtained
for lower values of $f$, which correspond to a less sharp sigmoid.
\end{argument}

\begin{figure*}[!ht]
\centering
\def\colwidth{0.1\textwidth}
\newcolumntype{M}[1]{>{\centering\arraybackslash}m{#1}}
\addtolength{\tabcolsep}{-4pt}
\begin{tabular}{M{\colwidth} M{\colwidth} M{\colwidth}  M{\colwidth}  M{\colwidth} M{\colwidth} M{\colwidth} M{\colwidth} M{\colwidth}}
 \small\texttt{bear} & \small\texttt{buddha} & \small\texttt{cat} & \small\texttt{cow} & \small\texttt{harvest} & \small\texttt{pot1} & \small\texttt{pot2} & \small\texttt{reading} & \small\texttt{goblet} \tabularnewline
\includegraphics[width=\linewidth]{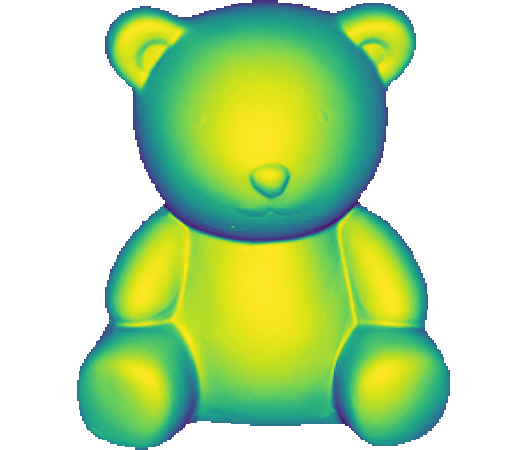} & 
\includegraphics[width=\linewidth]{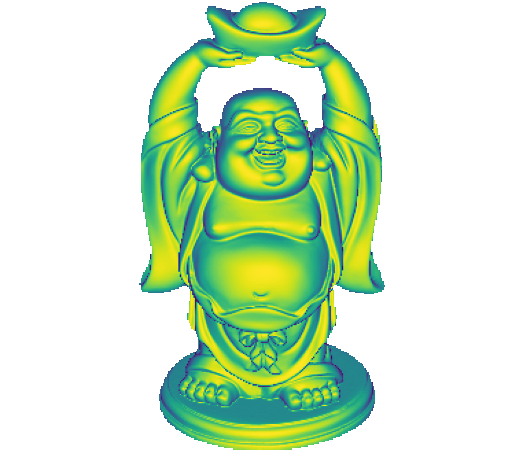} &
\includegraphics[width=\linewidth]{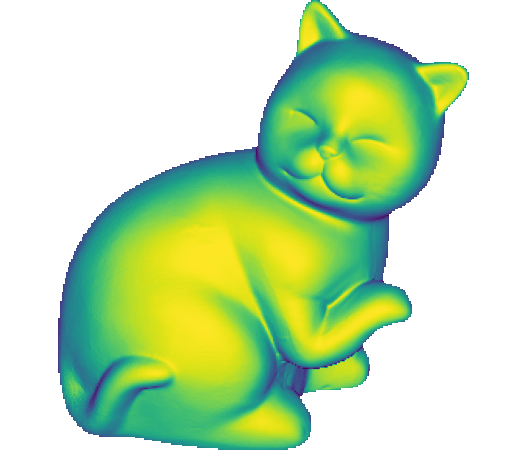} &
\includegraphics[width=\linewidth]{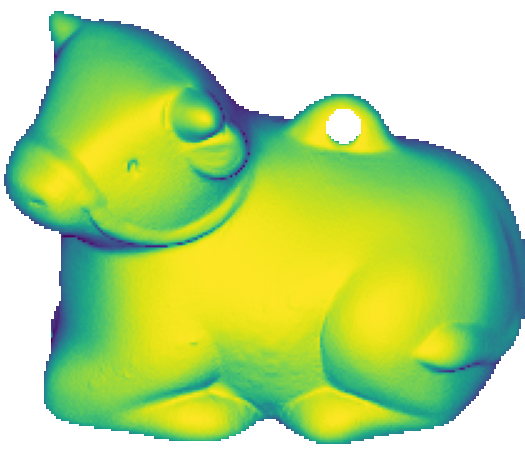} &
\includegraphics[width=\linewidth]{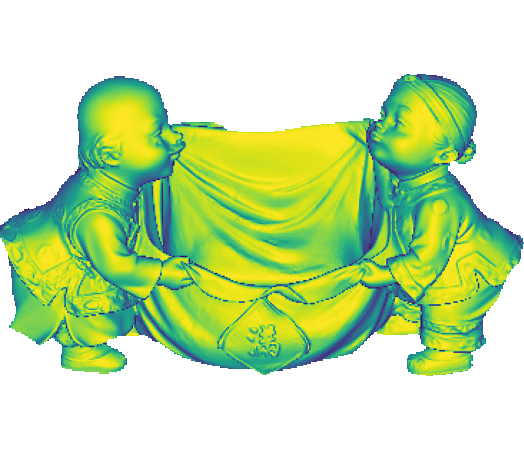} &
\includegraphics[width=\linewidth]{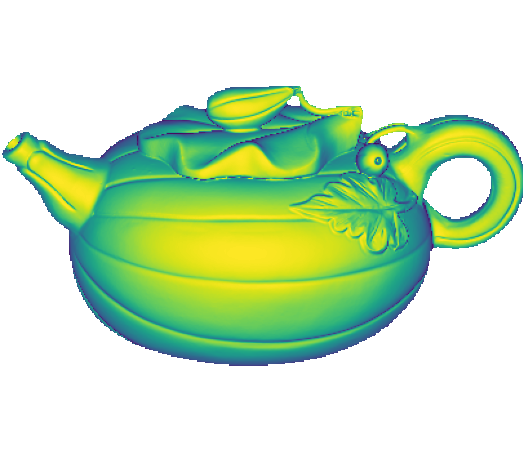} &
\includegraphics[width=\linewidth]{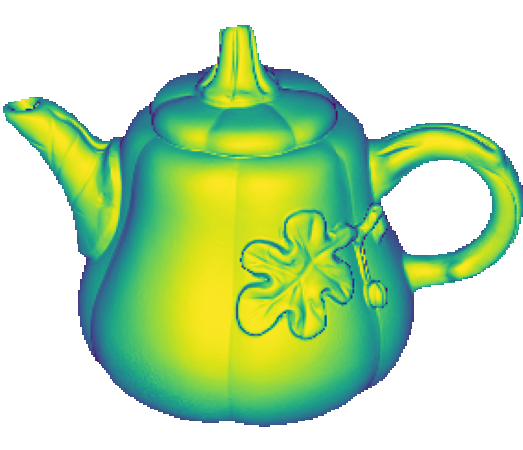} &
\includegraphics[width=\linewidth]{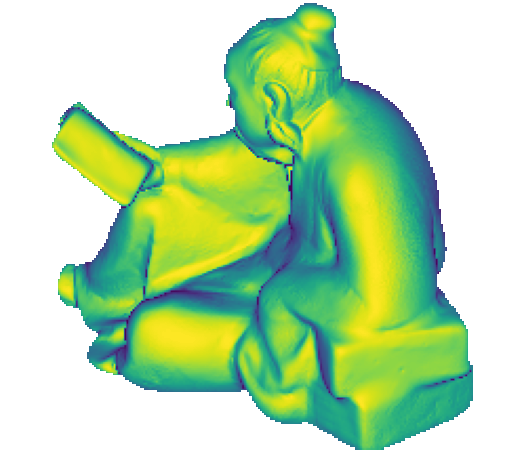} &
\includegraphics[width=\linewidth]{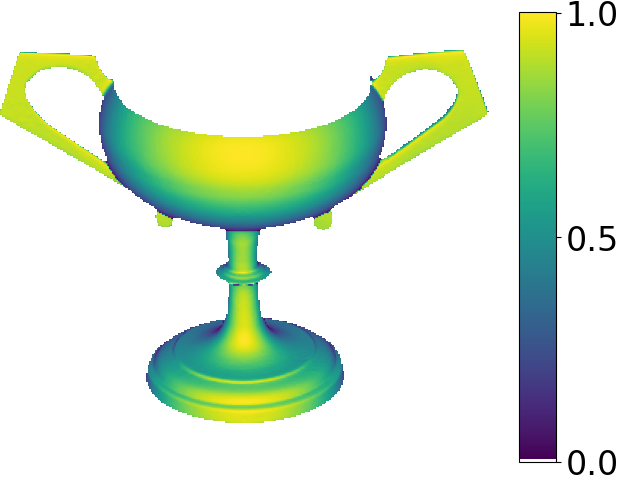}
\tabularnewline
\end{tabular}
\addtolength{\tabcolsep}{4pt}
\caption{\textbf{Visualization of the terms $|\boldsymbol{n_a}^\mathsf{T} \boldsymbol{\tau_a}|$, DiLiGenT dataset~\cite{Shi2016DiLiGenT}.} The terms encode the degree of collinearity between the surface normals and the ray direction vectors. Low values are attained at pixels where the ray direction vector is perpendicular to the surface normal, a necessary condition for the corresponding point to lie on the object boundary.}
\label{fig:visualization_n_a_dot_tau_a}
\end{figure*}
\begin{table*}[!ht]
    \centering
    \resizebox{0.9\linewidth}{!}{
    \begin{tabular}{ccccccccccc}
    \toprule
    \multicolumn{2}{c}{Value of $\gamma_{b\rightarrow a}$} & \multirow{2}{*}{\texttt{bear}} & \multirow{2}{*}{\texttt{buddha}} & \multirow{2}{*}{\texttt{cat}} & \multirow{2}{*}{\texttt{cow}} & \multirow{2}{*}{\texttt{harvest}} & \multirow{2}{*}{\texttt{pot1}} & \multirow{2}{*}{\texttt{pot2}} & \multirow{2}{*}{\texttt{reading}} & \multirow{2}{*}{\texttt{goblet}}\\
    \cline{1-2}
    Cost function~\eqref{eq:optimization_cost_function_with_gamma_a_from_b} & \rule{0pt}{10pt} $w_{b\rightarrow a}^{\mathrm{BiNI}}$~\eqref{eq:bini_weight_with_gamma_a_from_b}\\
    \midrule
     $f\cdot\boldsymbol{n_a}^\mathsf{T} \boldsymbol{\tau_a}$ & $f\cdot\boldsymbol{n_a}^\mathsf{T} \boldsymbol{\tau_a}$ & $0.07$ & $0.26$ & $0.06$ & $0.08$ & $5.54$ & $0.49$ & $0.13$ & $0.11$ & $6.33$ \\
     \arrayrulecolor{gray!70}\specialrule{0.2pt}{0.2pt}{0.2pt}
    \arrayrulecolor{black}
     $\boldsymbol{n_a}^\mathsf{T} \boldsymbol{\tau_a}$ & $f\cdot\boldsymbol{n_a}^\mathsf{T} \boldsymbol{\tau_a}$ & $0.07$ & $0.25$ & $0.06$ & $0.08$ & $5.33$ & $0.49$ & $0.13$ & $0.12$ & $6.60$ \\
     $\boldsymbol{n_a}^\mathsf{T} \boldsymbol{\tau_a}$ & $\num{3000}\cdot\boldsymbol{n_a}^\mathsf{T} \boldsymbol{\tau_a}$ & $0.09$ & $0.27$ & $0.11$ & $0.09$ & $3.89$ & $0.47$ & $0.15$ & $0.12$ & $7.96$ \\
     $\boldsymbol{n_a}^\mathsf{T} \boldsymbol{\tau_a}$ & $\num{2000}\cdot\boldsymbol{n_a}^\mathsf{T} \boldsymbol{\tau_a}$ & $0.06$ & $0.98$ & $0.17$ & $0.18$ & $1.71$ & $0.48$ & $0.25$ & $0.27$ & $8.63$ \\
     $\boldsymbol{n_a}^\mathsf{T} \boldsymbol{\tau_a}$ & $\num{1000}\cdot\boldsymbol{n_a}^\mathsf{T} \boldsymbol{\tau_a}$ & $0.04$ & $1.41$ & $0.08$ & $0.30$ & $2.51$ & $0.72$ & $0.28$ & $1.19$ & $9.46$ \\
     \arrayrulecolor{gray!70}\specialrule{0.2pt}{0.2pt}{0.2pt}
    \arrayrulecolor{black}
    $f$ & $f\cdot\boldsymbol{n_a}^\mathsf{T} \boldsymbol{\tau_a}$ & $0.48$ & $2.53$ & $0.69$ & $0.39$ & $4.84$ & $14.40$ & $0.42$ & $3.16$ & $10.28$ \\
    \bottomrule
    \end{tabular}
    }
    \caption{\textbf{Ablation on the terms in $\gamma_{b\rightarrow a}$, DiLiGenT dataset~\cite{Shi2016DiLiGenT}.}
    For each experiment, we report the mean absolute depth error (MADE) [$\boldsymbol{\si{mm}}$].
    All experiments are without $\alpha_{b\rightarrow a}$ computation, $k=2$ for $w_{b\rightarrow a}^{(t)}$ (as default), and are run for $\num{1200}$ iterations. Where used, $f$ denotes $\left\|\boldsymbol{u_b}-\boldsymbol{u_a}\right\|/ \left\|\boldsymbol{\tau_b}-\boldsymbol{\tau_a}\right\|$. For reference, the values of $f_x$ and $f_y$ in the dataset are $f_x\approx \SI{3772.1}{[px]}$ and $f_y\approx \SI{3759.0}{[px]}$.}
    \label{tab:analysis_gamma_a_from_b}
\end{table*}
\begin{argument}[Argument for Proposition 2]
Since unlike $f$ the term $\boldsymbol{n_a}^\mathsf{T} \boldsymbol{\tau_a}$ is highly pixel specific, it is not possible to find a single constant that can be absorbed into the parameter $k$ of the sigmoid. It is therefore not straightforward to draw conclusions about its contribution to Effect 2 (sharpness of the bilateral weights). We can however verify that the term $\boldsymbol{n_a}^\mathsf{T} \boldsymbol{\tau_a}$
has a strong influence on
Effect 1 (weighting), by removing it from the $\gamma_{b\rightarrow a}$ 
factor of
the cost function (which is therefore set to $f$), while maintaining it in $\gamma_{b\rightarrow a}$ in the bilateral weights. Comparing the last and the first row of \cref{tab:analysis_gamma_a_from_b} confirms that the accuracy of the reconstruction dramatically decreases when the term does not contribute to the cost function, which indicates that it 
plays an active role in determining the convergence of the optimization, by introducing equation-specific weights.
Interestingly, as
we
previously observed in \cref{sec:method_general_formulation},
the term 
$\boldsymbol{n_a}^\mathsf{T} \boldsymbol{\tau_a}$ strongly correlates with surface discontinuities, with pixels close to object boundaries or local discontinuities attaining a small value for this term.
More generally,
as evident from its dot-product definition, the term $\boldsymbol{n_a}^\mathsf{T} \boldsymbol{\tau_a}$ encodes the degree of collinearity between surface normal and the ray direction vector at each pixel (\cf. \cref{fig:visualization_n_a_dot_tau_a} for a visualization). As a consequence, its
effect over the optimization
is
to
balance the
influence of the residuals,
decreasing
the weight
of errors close to discontinuities, 
while
increasing the
influence
of residuals at points
where the camera rays hit the surface at a close-to-right angle.

\end{argument}

\section{Analysis of the positivity of the $\log$ term~\label{sec_suppl:analysis_positivity_log_term}}

In this Section we provide further insights on the positivity of the $\log$ term in our formulation (\eqref{eq:ours_relationship_equation_t} in the main paper).

We start by empirically verifying that, for our choice $\boldsymbol{\tau_m} = (\boldsymbol{\tau_a} + \boldsymbol{\tau_b}) / 2$, the terms $\boldsymbol{n_a}^\mathsf{T}\boldsymbol{\tau_m}$ and $\boldsymbol{n_b}^\mathsf{T}\boldsymbol{\tau_m}$ are both strictly positive for all but a single pixel (object \texttt{pot1}) across all the objects in the DiLiGenT dataset, used for our main experiments. Furthermore, also for this outlier pixel, the effects of the two pixels cancel out and the corresponding term
$\omega_{b\rightarrow a} = (\boldsymbol{n_a}^\mathsf{T} \boldsymbol{\tau_m}\cdot\boldsymbol{n_b}^\mathsf{T} \boldsymbol{\tau_b}) / (\boldsymbol{n_a}^\mathsf{T} \boldsymbol{\tau_a}\cdot\boldsymbol{n_b}^\mathsf{T} \boldsymbol{\tau_m})$
is strictly positive,
leading to a positive $\log$ term at all pixels in the first iteration of our optimization.

\begin{figure}[!t]
    \centering
    \includegraphics[height=0.55\linewidth]{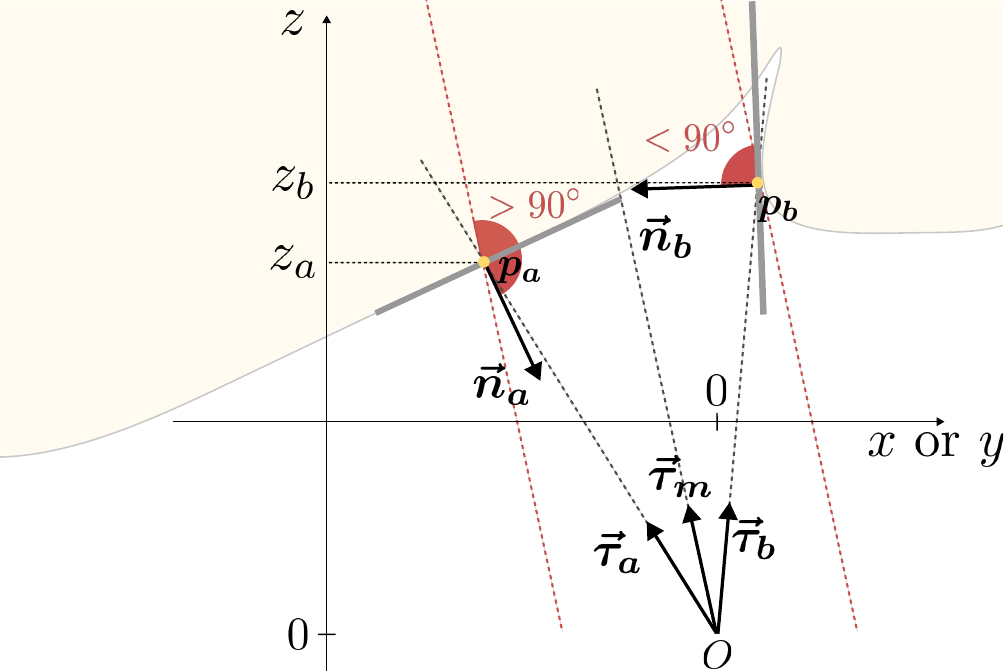}
    \caption{\textbf{Visualization of a corner case in our local planarity assumption in 3D.} For the chosen configuration, the ray direction vector $\boldsymbol{\tau_m}$
    forms an angle smaller than $90^\circ$ with $\boldsymbol{n_b}$ and larger than $90^\circ$ with $\boldsymbol{n_a}$, resulting in $\boldsymbol{n_a}^\mathsf{T}\boldsymbol{\tau_m} < 0$ and $\boldsymbol{n_b}^\mathsf{T}\boldsymbol{\tau_m}>0$.}
    \label{fig:surface_normal_integration_scheme_3d_corner_case}
    \vspace{-10pt}
\end{figure}

We now briefly analyze under which conditions we can expect an outlier, negative $\omega_{b\rightarrow a}$ term. Since, as noted in \cref{sec:method_general_formulation}, for physically meaningful normals (\ie, corresponding to observable surface points) the positivity of $\omega_{b\rightarrow a}$ reduces to the positivity of $\boldsymbol{n_a}^\mathsf{T} \boldsymbol{\tau_m}$ and $\boldsymbol{n_b}^\mathsf{T} \boldsymbol{\tau_m}$, we can focus on the case where the latter two terms have opposite signs.
\Cref{fig:surface_normal_integration_scheme_3d_corner_case} provides an illustration of
an instance
in which
such
a
corner case may arise. In the depicted setting, the surface has low inclination relative to the camera on the side of 
point $\boldsymbol{p_a}$, but
large inclination
on the side of point $\boldsymbol{p_b}$.
As consequence, on
the
side
of $\boldsymbol{p_a}$
both the angles 
between $\boldsymbol{n_a}$ and $\boldsymbol{\tau_a}$ and between $\boldsymbol{n_a}$ and $\boldsymbol{\tau_m}$ are significantly
larger
than $90^\circ$, \ie
$\boldsymbol{n_a}^\mathsf{T}\boldsymbol{\tau_a}<0$ and
$\boldsymbol{n_a}^\mathsf{T}\boldsymbol{\tau_m}<0$.
On the
opposite side,
however,
the
angle between $\boldsymbol{n_b}$ and $\boldsymbol{\tau_b}$ 
is only
slightly
larger
than $90^\circ$ (hence $\boldsymbol{n_b}^\mathsf{T}\boldsymbol{\tau_b}\approx0$, but
still negative),
while the angle between $\boldsymbol{n_b}$ and $\boldsymbol{\tau_m}$ is
smaller
than $90^\circ$, causing $\boldsymbol{n_b}^\mathsf{T}\boldsymbol{\tau_m}$ to be positive and therefore $\omega_{b\rightarrow a}$ to be negative.
While such outlier cases might indeed arise,
it is possible to detect and handle them, for instance by excluding the corresponding equation from the optimization or by choosing a different value of $\boldsymbol{\tau_m}$ (\cf \cref{sec_suppl:ablation_lambda_m}).
Furthermore, their occurrence is
unlikely
in practice, since the sign flipping between $\boldsymbol{n_b}^\mathsf{T}\boldsymbol{\tau_b}$ and $\boldsymbol{n_b}^\mathsf{T}\boldsymbol{\tau_m}$
would need to occur
within
a
very limited
angular
space:
as a reference, using $\boldsymbol{\tau_m}=(\boldsymbol{\tau_a}+\boldsymbol{\tau_b})/2$, the angle between $\boldsymbol{\tau_m}$ and $\boldsymbol{\tau_b}$ is approximately $\frac{1}{2}\arctan\left(\frac{\SI{1}{px}}{\SI{3700}{px}}\right)\approx0.008^\circ$ in the DiLiGenT dataset, for which $f_x\approx \SI{3772.1}{px}$ and $f_y\approx \SI{3759.0}{px}$.

Assuming
$\omega_{b\rightarrow a} > 0$, hence
that
the argument of the $\log$ term in \eqref{eq:ours_relationship_equation_t} is positive in the first optimization iteration, it is straightforward to show that the argument also stays positive throughout the optimization, as we prove below.\\
From \eqref{eq:update_scheme_alpha} in the main paper, $\omega_{b\rightarrow a} + \omega_{\varepsilon_a}\cdot\alpha_{b\rightarrow a}^{(t+1)} = \exp\left({\tilde{z}_a^{(t)}} - {\tilde{z}_b^{(t)}}\right)$. Since the exponential function is bijective and defined anywhere in $\mathbb{R}$, it follows that for any value of $\tilde{z}_a^{(t)}$ and $\tilde{z}_b^{(t)}$ a corresponding value for the term $\omega_{b\rightarrow a} + \omega_{\varepsilon_a}\cdot\alpha_{b\rightarrow a}^{(t+1)}$ can be found and thereby of $\alpha_{b\rightarrow a}^{(t+1)}$ (provided that $\omega_{\varepsilon_a} \neq 0$, \ie, from~\eqref{eq:ours_coefficient_definition} $n_{a_z}\neq0$, which is always the case because $n_{a_z}=0$ corresponds to a surface perpendicular to the image plane). Since the exponential function has strictly positive codomain, it also follows that for all $t$'s:
    \begin{equation}
     \omega_{b\rightarrow a} + \omega_{\varepsilon_a}\cdot\alpha_{b\rightarrow a}^{(t+1)} > 0.
     \label{eq:positivity_log_with_alpha}
    \end{equation}
From $\omega_{b\rightarrow a} > 0$ and
\eqref{eq:positivity_log_with_alpha}
and since $\beta_{b\rightarrow a}^{(t)}\in[0, 1]$ by design, it follows that 
    \hbox{$\omega_{b\rightarrow a}+\omega_{\varepsilon_a}\cdot\alpha_{b\rightarrow a}^{(t)}\cdot\beta_{b\rightarrow a}^{(t)} > 0$}, which proves the hypothesis. Indeed:\\
\begin{itemize}
    \item If $\medmath{\omega_{\varepsilon_a}\cdot\alpha_{b\rightarrow a}^{(t)}\ge0}$, one has
\begin{flalign*}
    & \medmath{\omega_{\varepsilon_a}\cdot\alpha_{b\rightarrow a}^{(t)}\cdot\beta_{b\rightarrow a}^{(t)}\ge0} && \medmath{\left(\beta_{b\rightarrow a}^{(t)}\ge0\right)} \\
    \medmath{\Rightarrow}\ &\medmath{\omega_{b\rightarrow a} + \omega_{\varepsilon_a}\cdot\alpha_{b\rightarrow a}^{(t)}\cdot\beta_{b\rightarrow a}^{(t)} \ge \omega_{b\rightarrow a}} && \medmath{\left(\omega_{b\rightarrow a}\in\mathbb{R}\right)}\\
    \medmath{\Rightarrow}\ &\medmath{\omega_{b\rightarrow a} + \omega_{\varepsilon_a}\cdot\alpha_{b\rightarrow a}^{(t)}\cdot\beta_{b\rightarrow a}^{(t)} > 0;} && \medmath{\left(\omega_{b\rightarrow a}>0\right)}
\end{flalign*}
\item If $\medmath{\omega_{\varepsilon_a}\cdot\alpha_{b\rightarrow a}^{(t)}<0}$, it follows that
\begin{flalign*}
    \ & \medmath{\omega_{\varepsilon_a}\cdot\alpha_{b\rightarrow a}^{(t)}\cdot\beta_{b\rightarrow a}^{(t)}\ge\omega_{\varepsilon_a}\cdot\alpha_{b\rightarrow a}^{(t)}} && \medmath{\left(\beta_{b\rightarrow a}^{(t)}\in[0, 1]\right)}\\
    \medmath{\Rightarrow}\ & \medmath{\omega_{b\rightarrow a} + \omega_{\varepsilon_a}\cdot\alpha_{b\rightarrow a}^{(t)}\cdot\beta_{b\rightarrow a}^{(t)} \ge}\\
    &\ \ \ \ \ \ \ \ \ \ \ \ \ \ \ \ \ \ \ \ \ \ \ \ \ \ \ \ \ \ \ \medmath{\omega_{b\rightarrow a} + \omega_{\varepsilon_a}\cdot\alpha_{b\rightarrow a}^{(t)}} && \medmath{\left(\omega_{b\rightarrow a}>0\right)}\\
    \medmath{\Rightarrow}\ & \medmath{\omega_{b\rightarrow a} + \omega_{\varepsilon_a}\cdot\alpha_{b\rightarrow a}^{(t)}\cdot\beta_{b\rightarrow a}^{(t)} > 0.} &&\medmath{\left(\mathrm{from}\ \eqref{eq:positivity_log_with_alpha}\right)}
\end{flalign*}
\end{itemize}

\section{Impact of the choice of $\boldsymbol{\tau_m}$~\label{sec_suppl:ablation_lambda_m}}
In the following Section, we provide an ablation on the choice of $\boldsymbol{\tau_m}$, which controls the planar assumption of our method (\cf \cref{fig:surface_normal_integration_scheme_3d_corner_case} and \cref{fig:surface_normal_integration_scheme_3d} in the main paper).

\begin{figure}[!t]
    \centering
    \includegraphics[height=0.55\linewidth]{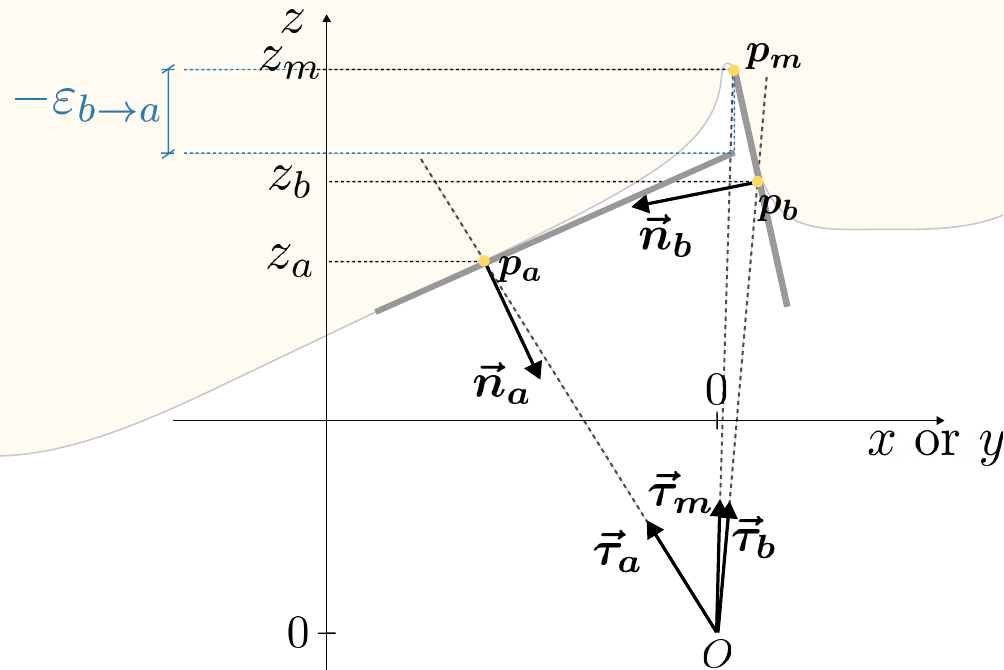}
    \caption{\textbf{Visualization of an adaptive strategy for $\boldsymbol{\tau_m}$.}
    If 
    the surface
    has a large inclination relative to the camera
    on
    one of the two 
    sides
    (here
    the side of $\boldsymbol{p_b}$, hence
    $|\boldsymbol{n_b}^\mathsf{T}\boldsymbol{\tau_b}|\ll|\boldsymbol{n_a}^\mathsf{T}\boldsymbol{\tau_a}|$), orienting $\boldsymbol{\tau_m}$ closer to the
    latter
    side 
    yields a smaller
    $|\varepsilon_{b\rightarrow a}|$.
    }
    \label{fig:surface_normal_integration_scheme_3d_ablation_tau_m}
    \vspace{-10pt}
\end{figure}

As mentioned in \cref{sec:method_general_formulation} in the main paper,
$\boldsymbol{\tau_m}$
can be parametrized
as interpolating
between $\boldsymbol{\tau_a}$ and $\boldsymbol{\tau_b}$, \ie, $\boldsymbol{\tau_m} = \boldsymbol{\tau_a} + \lambda_m(\boldsymbol{\tau_b} - \boldsymbol{\tau_a})$, with $\lambda_m\in[0, 1]$. A natural choice, which we adopt in our main experiments, is to orient $\boldsymbol{\tau_m}$ at an equal angular distance from $\boldsymbol{\tau_a}$ and $\boldsymbol{\tau_b}$, \ie setting $\lambda_m=0.5$ uniformly for all pixels.
However, we note that in certain settings a pixel-pair-specific choice $\lambda_{m, b\rightarrow a}: \boldsymbol{\tau_m} = \boldsymbol{\tau_a} + \lambda_{m, b\rightarrow a}(\boldsymbol{\tau_b} - \boldsymbol{\tau_a})$ might be desirable.
An argument in favor of this point is for instance shown 
through
a
corner case
similar to that
considered in \cref{sec_suppl:analysis_positivity_log_term} (\cref{fig:surface_normal_integration_scheme_3d_ablation_tau_m}), in which on one of the two sides (the side of $\boldsymbol{p_b}$ in 
\cref{fig:surface_normal_integration_scheme_3d_ablation_tau_m})
the surface has a significantly larger inclination relative to the camera.
As a consequence, as exemplified by \cref{fig:surface_normal_integration_scheme_3d_ablation_tau_m}, our planar assumption
holds
more accurately if $\boldsymbol{\tau_m}$
is
oriented closer to the side with the larger inclination, in which case a smaller discontinuity term $|\varepsilon_{b\rightarrow a}|$
is
obtained. Since, as mentioned in \cref{sec_suppl:analysis_gamma_factor}, the quantity $\boldsymbol{n_a}^\mathsf{T}\boldsymbol{\tau_a}$ naturally encodes surface orientation with respect to the camera, the condition of unbalanced inclination
between the two sides
can also be expressed as 
$|\boldsymbol{n_b}^\mathsf{T}\boldsymbol{\tau_b}|\ll|\boldsymbol{n_a}^\mathsf{T}\boldsymbol{\tau_a}|$. In this ablation, we additionally consider the quantity $n_{az}$, which similarly to $\boldsymbol{n_a}^\mathsf{T}\boldsymbol{\tau_a}$ attains a low value in proximity to discontinuities.

We note that
the interpolating function $\lambda_{m,b\rightarrow a}$ needs to be
such
that
$\boldsymbol{\tau_m}$ intersects the same surface
point $\boldsymbol{p_m}$
both in the direction $b\rightarrow a$ (\ie, when considering $b$ a neighbor of $a$) and
in
the direction $a\rightarrow b$ (\ie, when considering $a$ a neighbor of $b$).
This can be expressed mathematically by the condition
$\lambda_{m, b\rightarrow a} = 1 - \lambda_{m, a\rightarrow b}$. We note that the sigmoid function naturally fulfills this condition when composed with an even function, and we therefore set in this ablation $\lambda_{m, b\rightarrow a} = \sigma_{k_m}\left(f(a, b)\right)$, with different values for $k_m$, and with $f(a, b)$ either $(\boldsymbol{n_a}^\mathsf{T}\boldsymbol{\tau_a})^2 - (\boldsymbol{n_b}^\mathsf{T}\boldsymbol{\tau_b})^2$, $n_{az}^2-n_{bz}^2$, or $(n_{az}\cdot\boldsymbol{n_a}^\mathsf{T}\boldsymbol{\tau_a})^2 - (n_{bz}\cdot\boldsymbol{n_b}^\mathsf{T}\boldsymbol{\tau_b})^2$.

\begin{table*}[!ht]
    \centering
    \resizebox{0.9\linewidth}{!}{
    \begin{tabular}{llccccccccc}
    \toprule
    $\lambda_m$ & $k_m$ & \texttt{bear} & \texttt{buddha} & \texttt{cat} & \texttt{cow} & \texttt{harvest} & \texttt{pot1} & \texttt{pot2} & \texttt{reading} & \texttt{goblet}\\
    \midrule
    $0.5$ & $\mathrm{N/A}$ & $0.07$ & $0.26$ & $0.06$ & $0.08$ & $4.83$ & $0.50$ & $0.13$ & $0.12$ & $6.56$\\
    \arrayrulecolor{gray!70}\specialrule{0.2pt}{0.2pt}{0.2pt}
    \arrayrulecolor{black}
    \multirow{3}{*}{$\sigma_{k_m}\left((\boldsymbol{n_a}^\mathsf{T}\boldsymbol{\tau_a})^2-(\boldsymbol{n_b}^\mathsf{T}\boldsymbol{\tau_b})^2\right)$} & $1$ & $0.15$ & $0.33$ & $0.09$ & $0.12$ & $5.12$ & $0.52$ & $0.17$ & $0.19$ & $5.73$\\
    & $2$ & $0.22$ & $0.72$ & $0.13$ & $0.16$ & $2.45$ & $0.53$ & $0.22$ & $0.29$ & $6.23$ \\
    & $3$ & $0.29$ & $1.40$ & $0.16$ & $0.19$ & $3.66$ & $0.56$ & $0.30$ & $0.38$ & $6.11$\\
    \arrayrulecolor{gray!70}\specialrule{0.2pt}{0.2pt}{0.2pt}
    \arrayrulecolor{black}
    \multirow{3}{*}{$\sigma_{k_m}\left(n_{az}^2-n_{bz}^2\right)$} & $1$ & $0.15$ & $0.33$ & $0.09$ & $0.12$ & $4.65$ & $0.50$ & $0.17$ & $0.19$ & $5.69$\\
    & $2$ & $0.22$ & $0.71$ & $0.13$ & $0.16$ & $2.51$ & $0.53$ & $0.22$ & $0.28$ & $6.12$\\
    & $3$ & $0.29$ & $1.42$ & $0.16$ & $0.19$ & $5.49$ & $0.56$ & $0.30$ & $0.38$ & $6.06$ \\
    \arrayrulecolor{gray!70}\specialrule{0.2pt}{0.2pt}{0.2pt}
    \arrayrulecolor{black}
    \multirow{3}{*}{$\sigma_{k_m}\left((n_{az}\cdot\boldsymbol{n_a}^\mathsf{T}\boldsymbol{\tau_a})^2-(n_{bz}\cdot\boldsymbol{n_b}^\mathsf{T}\boldsymbol{\tau_b})^2\right)$} & $1$ & $0.11$ & $0.25$ & $0.08$ & $0.11$ & $4.95$ & $0.51$ & $0.16$ & $0.15$ & $5.38$ \\
    & $2$ & $0.15$ & $0.45$ & $0.10$ & $0.13$ & $2.73$ & $0.52$ & $0.20$ & $0.20$ & $5.40$ \\
    & $3$ & $0.19$ & $1.03$ & $0.13$ & $0.16$ & $2.74$ & $0.55$ & $0.24$ & $0.39$ & $5.61$ \\
    \bottomrule
    \end{tabular}
    }
    \caption{\textbf{Mean absolute depth error (MADE) [$\boldsymbol{\si{mm}}$] on the DiLiGenT benchmark~\cite{Shi2016DiLiGenT} for different choices of $\lambda_m$, where $\boldsymbol{\tau_m} = \boldsymbol{\tau_a} + \lambda_m(\boldsymbol{\tau_b} - \boldsymbol{\tau_a})$.} All the experiments are run for $\num{1200}$ iterations with $\alpha_{b\rightarrow a} =0$. $\sigma_{k_m}$ denotes the sigmoid function $\sigma_{k_m}(x)=1/(1+\exp(-k_m\cdot x))$.}
    \label{tab:ablation_lambda_m}
\end{table*}
\Cref{tab:ablation_lambda_m} shows the results of this ablation, which we perform on the DiLiGenT dataset. For most
objects, introducing a pixel-specific $\lambda_m$ results generally in lower reconstruction accuracy using any of the functions $f(a, b)$ listed above;
larger values of $k_m$ (hence more sharply weighting inclination differences between the two sides) further
decrease
the performance. A noticeable exception is represented by the two objects with larger discontinuities (\texttt{harvest} and \texttt{goblet}), for which specific choices of parameters can lead to improved reconstruction accuracy.

Finally, we highlight that pixel-specific values of $\lambda_m$
find an additional,
critical application in
handling potential
outliers in the input normal map. We discuss this important aspect in detail in \cref{sec_suppl:results_for_noisy_inputs}.

\section{Impact of the discontinuity activation term~\label{sec_suppl:ablation_beta_a_from_b_discont_activation_term}}

In this Section, we provide an ablation analysis on the impact of our discontinuity activation term $\beta_{b\rightarrow a}^{(t)}$ on the reconstruction accuracy. \Cref{tab:ablation_beta_a_from_b} reports the mean absolute depth error on the DiLiGenT dataset as we vary the 
hyperparameters $q$ and $\rho$
(\cf.~\eqref{eq:beta_definition_ours} in the main paper), the effect of which can be visualized in \cref{fig:beta_a_from_b_ablation}.
For $\rho=0.25$, the results show object-specific trends, with some objects achieving higher accuracy for sharper changes of $\beta^{(t)}_{b\rightarrow a}$ (larger $q$, for instance \texttt{harvest}, \texttt{pot1}, \texttt{reading}) and others favoring a smoother discontinuity activation term (smaller $q$, for instance \texttt{bear}, \texttt{pot2}).
For $\rho=0.5$, the method achieves
worse accuracy, in
most
instances also lower than the version without computation of $\alpha_{b\rightarrow a}$ (\cf \cref{tab:diligent_results_main} in the main paper). This performance drop is expected, since for $\rho=0.5$ the discontinuity term
significantly deviates from its designed objective, namely that it should tend smoothly to zero as $w_{b\rightarrow a}^{\mathrm{BiNI}^{(t-1)}}\to0.5^-$ and smoothly to one as $w_{b\rightarrow a}^{\mathrm{BiNI}^{(t-1)}}\to0^+$ (\cf \cref{sec:method_bilateral_formulation} in the main paper for a detailed explanation of this design choice).
\begin{table*}[!ht]
    \centering
    \resizebox{0.7\linewidth}{!}{
    \begin{tabular}{llccccccccc}
    \toprule
    $\rho$ & $q$ & \texttt{bear} & \texttt{buddha} & \texttt{cat} & \texttt{cow} & \texttt{harvest} & \texttt{pot1} & \texttt{pot2} & \texttt{reading} & \texttt{goblet}\\
    \midrule
    \multirow{9}{*}{$0.25$} & $2.5$ & $0.04$ & $0.28$ & $\mathbf{0.06}$ & $0.11$ & $4.35$ & $0.57$ & $0.13$ & $0.17$ & $5.86$\\
    & $5.0$ & $\mathbf{0.02}$ & $\mathbf{0.22}$ & $0.22$ & $0.09$ & $1.11$ & $0.53$ & $\mathbf{0.12}$ & $0.14$ & $2.43$ \\
    & $10.0$ & $\mathbf{0.02}$ & $0.25$ & $\mathbf{0.06}$ & $\mathbf{0.08}$ & $0.93$ & $0.54$ & $\mathbf{0.12}$ & $0.16$ & $1.63$ \\
    & $15.0$ & $0.03$ & $0.24$ & $\mathbf{0.06}$ & $\mathbf{0.08}$ & $0.78$ & $0.55$ & $\mathbf{0.12}$ & $0.16$ & $\mathbf{1.52}$ \\
    & $25.0$ & $0.03$ & $0.25$ & $\mathbf{0.06}$ & $0.10$ & $0.83$ & $0.55$ & $0.13$ & $0.13$ & $5.78$ \\
    & $40.0$ & $0.03$ & $0.23$ & $\mathbf{0.06}$ & $\mathbf{0.08}$ & $\mathbf{0.60}$ & $0.51$ & $0.13$ & $0.18$ & $6.22$ \\
    & $50.0$ & $0.03$ & $0.24$ & $\mathbf{0.06}$ & $\mathbf{0.08}$ & $0.73$ & $0.49$ & $0.13$ & $0.17$ & $4.72$ \\
    & $100.0$ & $0.03$ & $0.23$ & $\mathbf{0.06}$ & $\mathbf{0.08}$ & $4.01$ & $\mathbf{0.48}$ & $0.14$ & $0.17$ & $6.21$ \\
    & $\num{1000.0}$ & $0.03$ & $0.23$ & $\mathbf{0.08}$ & $\mathbf{0.08}$ & $0.64$ & $\mathbf{0.48}$ & $0.14$ & $\mathbf{0.10}$ & $6.10$\\
    \arrayrulecolor{gray!70}\specialrule{0.2pt}{0.2pt}{0.2pt}
    \arrayrulecolor{black}
    \multirow{9}{*}{$0.50$} & $2.5$ & $0.08$ & $0.39$ & $\mathbf{0.06}$ & $0.12$ & $2.20$ & $0.62$ & $0.14$ & $0.20$ & $5.98$\\
    & $5.0$ & $0.09$ & $0.47$ & $0.09$ & $0.12$ & $3.40$ & $0.64$ & $0.13$ & $0.52$ & $6.25$\\
    & $10.0$ & $0.09$ & $0.52$ & $0.09$ & $0.12$ & $1.88$ & $0.58$ & $0.13$ & $0.54$ & $6.18$\\
    & $15.0$ & $0.09$ & $0.57$ & $0.08$ & $0.12$ & $2.52$ & $0.64$ & $0.18$ & $0.55$& $6.14$ \\
    & $25.0$ & $0.09$ & $0.67$ & $0.08$ & $0.12$ & $1.10$ & $0.63$ & $0.17$ & $0.73$ & $4.62$ \\
    & $40.0$ & $0.09$ & $0.40$ & $0.11$ & $0.12$ & $2.13$ & $0.69$ & $0.16$ & $0.59$ & $6.96$ \\
    & $50.0$ & $0.09$ & $0.70$ & $0.12$ & $0.12$ & $2.21$ & $0.61$ & $0.17$ & $0.45$ & $7.23$ \\
    & $100.0$ & $0.10$ & $0.83$ & $0.11$ & $0.11$ & $2.03$ & $0.60$ & $0.16$ & $0.46$ & $7.26$ \\
    & $\num{1000.0}$ & $0.10$ & $0.70$ & $0.14$ & $0.11$ & $2.58$ & $0.87$ & $0.16$ & $0.51$ & $6.94$\\
    \bottomrule
    \end{tabular}
    }
    \caption{\textbf{Mean absolute depth error (MADE) [$\boldsymbol{\si{mm}}$] on the DiLiGenT dataset~\cite{Shi2016DiLiGenT} for }$\rho\in\{0.25, 0.50\}$\textbf{ and different values of }$q$\textbf{.} For each object, \textbf{bold}
    denotes
    the best
    result across the experiments. All the experiments are run for $\num{1200}$ iterations.}
    \label{tab:ablation_beta_a_from_b}
\end{table*}
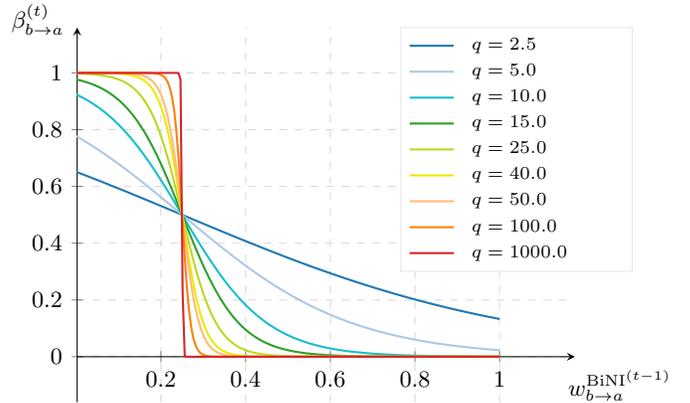
\begin{figure}[t!]
    \centering
    \begin{tikzpicture}[scale=0.98]
        \begin{axis}[
            width=\linewidth,
            height=0.8\linewidth,
            grid=both,
            grid style={dashed, gray!30},
            xlabel={$w_{b\rightarrow a}^{\mathrm{BiNI}^{(t-1)}}$},
            ylabel={$\beta_{b\rightarrow a}^{(t)}$},
            xmax=1.08,
            xmin=0.1,
            ymax=1.05,
            ymin=-0.05,
            xlabel style={below right},
            axis y line=left,
            axis x line=middle,
            every axis x label/.style={at={(current axis.right of origin)},right=7.5mm,below=0.5mm},
            every axis y label/.style={at={(current axis.north west)},above=2mm,left=0.5mm},
            enlargelimits=true,
            ticklabel style={font=\small},
            label style={font=\small},
            title style={font=\small},
            legend pos=north east,
            legend style={
                align=left, 
                column sep=1ex, 
                font=\scriptsize,
                text width=2cm, 
                at={(0.65, 1)}, 
                anchor=north west,
                draw,
                line width=0.1pt,
                draw=gray!20,
                fill=white!50
            }
        ]
            \addplot[domain=0:1, samples=200, thick, abl_color_1] {1. / (1 + exp(2.5*(x - 0.25)))};
            \addlegendentry{$q = 2.5$}
            \addplot[domain=0:1, samples=200, thick, abl_color_2] {1. / (1 + exp(5.0*(x - 0.25)))};
            \addlegendentry{$q = 5.0$}
            \addplot[domain=0:1, samples=200, thick, abl_color_3] {1. / (1 + exp(10.0*(x - 0.25)))};
            \addlegendentry{$q = 10.0$}
            \addplot[domain=0:1, samples=200, thick, abl_color_4] {1. / (1 + exp(15.0*(x - 0.25)))};
            \addlegendentry{$q = 15.0$}
            \addplot[domain=0:1, samples=200, thick, abl_color_5] {1. / (1 + exp(25.0*(x - 0.25)))};
            \addlegendentry{$q = 25.0$}
            \addplot[domain=0:1, samples=200, thick, abl_color_6] {1. / (1 + exp(40.0*(x - 0.25)))};
            \addlegendentry{$q = 40.0$}
            \addplot[domain=0:1, samples=200, thick, abl_color_7] {1. / (1 + exp(50.0*(x - 0.25)))};
            \addlegendentry{$q = 50.0$}
            \addplot[domain=0:1, samples=200, thick, abl_color_8] {1. / (1 + exp(100.0*(x - 0.25)))};
            \addlegendentry{$q = 100.0$}
            \addplot[domain=0:1, samples=200, thick, abl_color_10] {1. / (1 + exp(1000.0*(x - 0.25)))};
            \addlegendentry{$q = 1000.0$}
        \end{axis}
    \end{tikzpicture}
    \caption{\textbf{Discontinuity activation term~\eqref{eq:beta_definition_ours} for} $\rho=0.25$ \textbf{and different values of }$q$\textbf{.} For $\rho=0.5$, the plots are shifted to the right by $0.25$ units along the $w_{b\rightarrow a}^{\mathrm{BiNI}^{(t-1)}}$ axis. \Cf. \cref{tab:ablation_beta_a_from_b} for a quantitative evaluation on the effect of the parameters $\rho$ and $q$.
    }
    \vspace{-10pt}
    \label{fig:beta_a_from_b_ablation}
\end{figure}

\section{Impact of the connectivity~\label{sec_suppl:ablation_connectivity}}
\begin{table*}[!ht]
    \centering
    \resizebox{0.9\linewidth}{!}{
    \begin{tabular}{llccccccccc}
    \toprule
    Method & Connectivity & \texttt{bear} & \texttt{buddha} & \texttt{cat} & \texttt{cow} & \texttt{harvest} & \texttt{pot1} & \texttt{pot2} & \texttt{reading} & \texttt{goblet}\\
    \midrule
    \multirow{3}{*}{Ours w/o $\alpha_{b\rightarrow a}$ computation} & $4$-connectivity & $0.07$ & $\mathbf{0.26}$ & $\mathbf{0.06}$ & $\mathbf{0.08}$ & $4.83$ & $0.50$ & $0.13$ & $\mathbf{0.12}$ & $6.56$ \\
    & $4$-connectivity (diagonal) & $0.26$ & $0.39$ & $0.30$ & $0.09$ & $\mathbf{1.68}$ & $0.47$ & $0.15$ & $0.26$ & $7.31$ \\
    & $8$-connectivity & $\mathbf{0.06}$ & $0.35$ & $0.29$ & $0.09$ & $2.56$ & $\mathbf{0.36}$ & $\mathbf{0.12}$ & $0.39$ & $\mathbf{4.44}$ \\
    \midrule
    \multirow{3}{*}{Ours} & $4$-connectivity & $\mathbf{0.03}$ & $\mathbf{0.24}$ & $\mathbf{0.06}$ & $\mathbf{0.08}$ & $\mathbf{0.73}$ & $0.49$ & $\mathbf{0.13}$ & $\mathbf{0.17}$ & $\mathbf{4.72}$ \\
    & $4$-connectivity (diagonal) & $0.12$ & $0.69$ & $0.28$ & $0.09$ & $1.76$ & $0.50$ & $0.14$ & $0.42$ & $5.56$ \\
    & $8$-connectivity & $0.15$ & $0.35$ & $0.32$ & $\mathbf{0.08}$ & $3.82$ & $\mathbf{0.37}$ & $\mathbf{0.13}$ & $0.50$ & $5.14$ \\
    \bottomrule
    \end{tabular}
    }
    \caption{\textbf{Mean absolute depth error (MADE) [$\boldsymbol{\si{mm}}$] on the DiLiGenT dataset~\cite{Shi2016DiLiGenT} for different connectivities.} For each object and method, \textbf{bold}
    denotes
    the best
    result across the connectivities. All the experiments are run for $\num{1200}$ iterations with $\boldsymbol{\tau_m} = (\boldsymbol{\tau_a} + \boldsymbol{\tau_b}) / 2$. $\mathrm{Ours}$ corresponds to the hyperparameter setting of our main experiments ($q=50.0$ and $\rho=0.25$ in~\eqref{eq:beta_definition_ours}).}
    \label{tab:ablation_connectivity}
\end{table*}
Since our method allows using pixel connectivities not limited to standard $4$-connectivity, in this Section we
investigate
whether using alternative connectivities can yield improved reconstruction accuracy.
\Cref{tab:ablation_connectivity} shows the results of this ablation, where we test our method on the DiLiGenT dataset using standard $4$-connectivity (as in the main paper), $4$-connectivity defined along the diagonals rather than the horizontal and vertical direction, and full $8$-connectivity.
While $4$-connectivity along the diagonals, with very limited exceptions,  generally results in significantly worse performance, we note that, interestingly, full $8$-connectivity produces comparable or slightly better reconstructions than standard $4$-connectivity on some objects (\eg \texttt{cow}, \texttt{pot1}, \texttt{pot2}). However, this improvement is contrasted by reduced accuracy on other objects (\eg \texttt{buddha}, \texttt{cat}, \texttt{reading}) and reduced effect of the $\alpha_{b\rightarrow a}$ computation, leaving standard $4$-connectivity as the most robust and balanced
option.

\section{Additional evaluations of the formulation accuracy~\label{sec_suppl:additional_evaluations_formulation_accuracy}}
In Tables~\ref{tab:comparison_formulation_accuracy_ver3} and~\cref{tab:comparison_formulation_accuracy_ver5}, similarly to \cref{tab:comparison_formulation_accuracy} in the main paper, we provide metrics to evaluate how accurately our formulation approximates the ground-truth relation between depth and surface normals compared to previous methods. In particular, to complement the evaluation of the \emph{absolute} accuracy from the main paper, we report here 
\emph{relative} metrics, specifically
the residual $|(\tilde{z}_a - \tilde{z}_b - \mathrm{RHS}\ /\ \gamma_{b\rightarrow a})\ /\ \tilde{z}_a|$ computed on the ground-truth log-depth map (\cref{tab:comparison_formulation_accuracy_ver3}) and the residual $\left|\left(z_a - \exp\left(\mathrm{RHS}\ /\ \gamma_{b\rightarrow a}\right)\cdot z_b\right)\ /\ z_a\right|$ computed on the ground-truth depth map (\cref{tab:comparison_formulation_accuracy_ver5}), where $\mathrm{RHS}$ denotes the right-hand side of \eqref{eq:bini_formulation} for BiNI and \eqref{eq:ours_formulation_log_with_bini_factor} for Ours. 

The results confirm the findings from the main paper. Namely, while for two objects our method has larger residual standard deviation than BiNI~\cite{Cao2022BiNI} (\texttt{buddha} and \texttt{pot1}), it achieves lower mean residual error by one or two orders of magnitude and lower 
standard deviation for most objects.

\section{Results for noisy inputs~\label{sec_suppl:results_for_noisy_inputs}}
\begin{figure*}[!t]
\centering
\def\colwidth{0.09\textwidth}
\def\colwidthb{0.12\textwidth}
\newcolumntype{M}[1]{>{\centering\arraybackslash}m{#1}}
\addtolength{\tabcolsep}{-4pt}
\begin{tabular}{M{\colwidth} M{\colwidth} M{\colwidth} M{\colwidth}  M{\colwidth}  M{\colwidth} m{\colwidthb} M{\colwidth} }
& \multicolumn{7}{c}{Percentage of outliers, input normal maps, and $|\boldsymbol{n_a}^\mathsf{T} \boldsymbol{\tau_a}|$}\tabularnewline
 & $2\%$ & $4\%$ & $6\%$ & $8\%$ & $10\%$ & $\textrm{ }\textrm{ }\textrm{ }\textrm{ }\textrm{ }\textrm{ }12\%$  \tabularnewline
& 
\includegraphics[width=\linewidth]{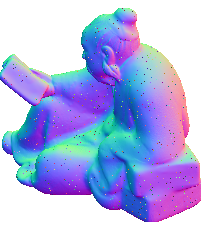} & 
\includegraphics[width=\linewidth]{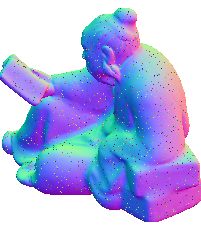} & 
\includegraphics[width=\linewidth]{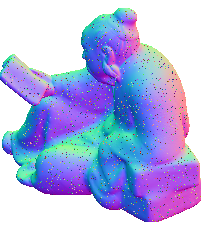} & 
\includegraphics[width=\linewidth]{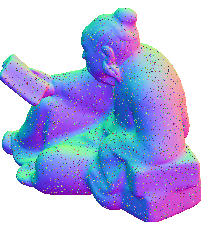} & 
\includegraphics[width=\linewidth]{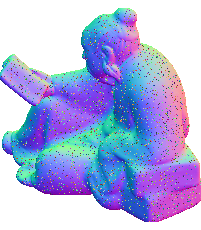} & 
\includegraphics[width=\colwidth]{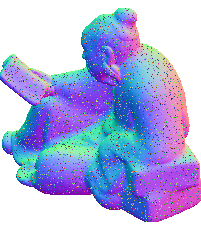}
\\[-1em]
& 
\includegraphics[width=\linewidth]{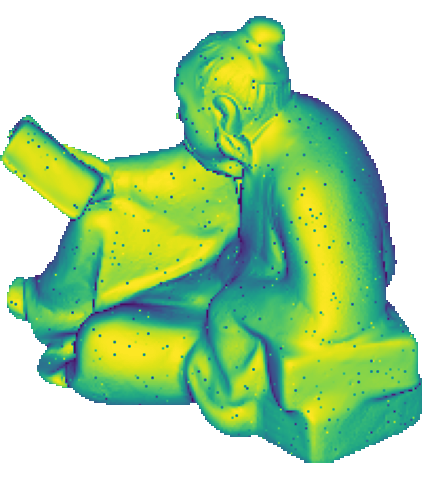} & 
\includegraphics[width=\linewidth]{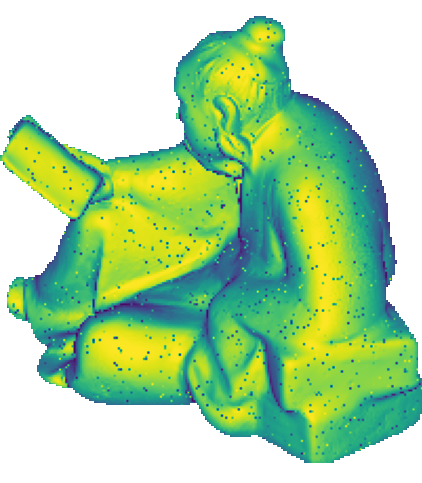} & 
\includegraphics[width=\linewidth]{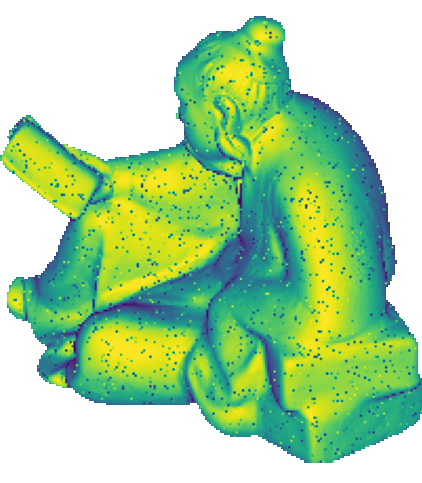} & 
\includegraphics[width=\linewidth]{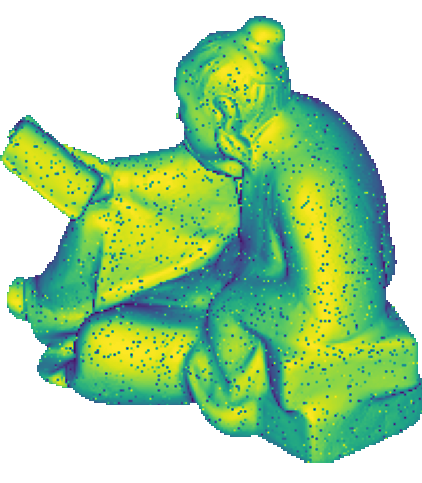} & 
\includegraphics[width=\linewidth]{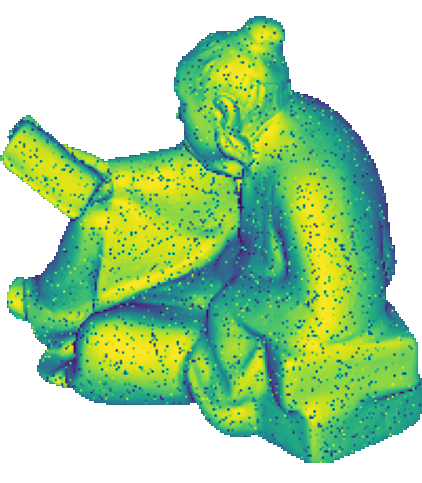} & 
\includegraphics[width=\linewidth]{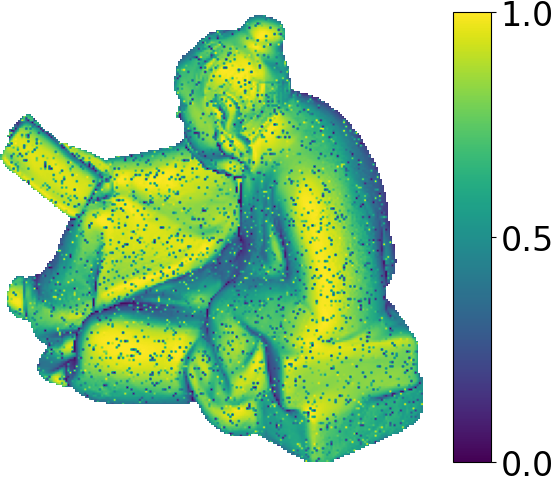}\tabularnewline
\noalign{\vskip -0.2em} 
\noalign{\global\savearrayrulewidth\arrayrulewidth} 
\noalign{\global\arrayrulewidth=1.2pt} 
\cline{2-7}
\noalign{\global\arrayrulewidth\savearrayrulewidth} 
\noalign{\vskip 0.2em}
& \multicolumn{7}{c}{No correction}\tabularnewline
& \includegraphics[width=\linewidth]{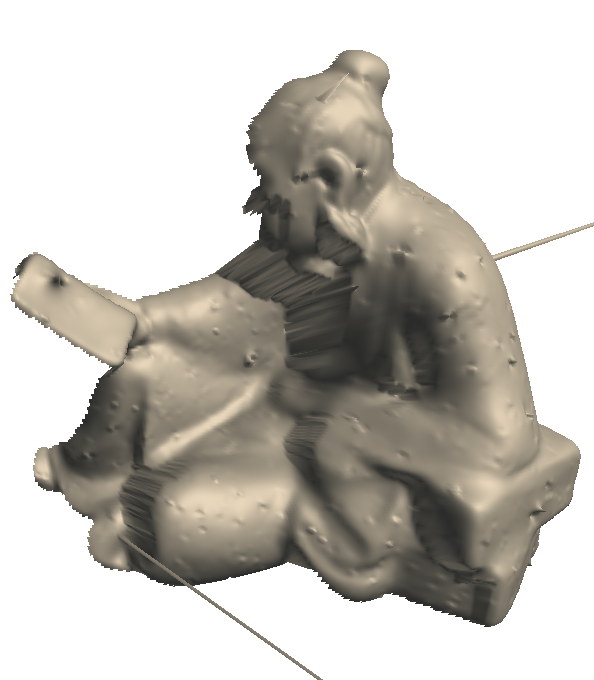}
& \includegraphics[width=\linewidth]{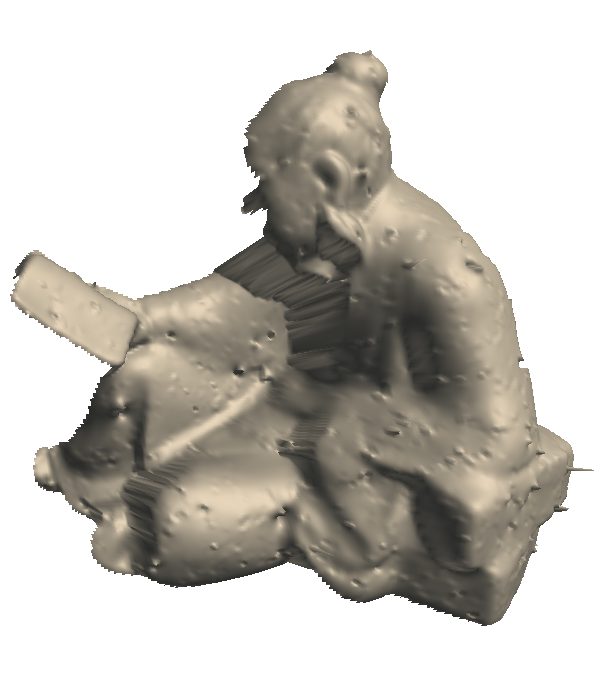}
& \includegraphics[width=\linewidth]{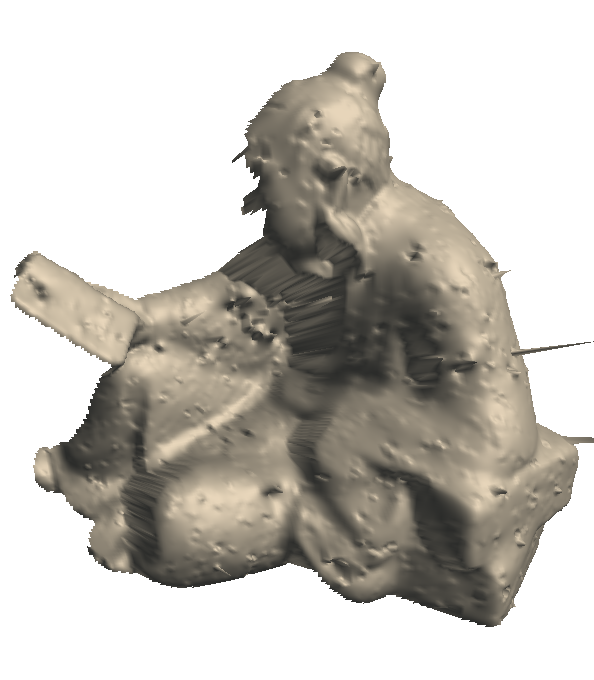}
& \includegraphics[width=\linewidth]{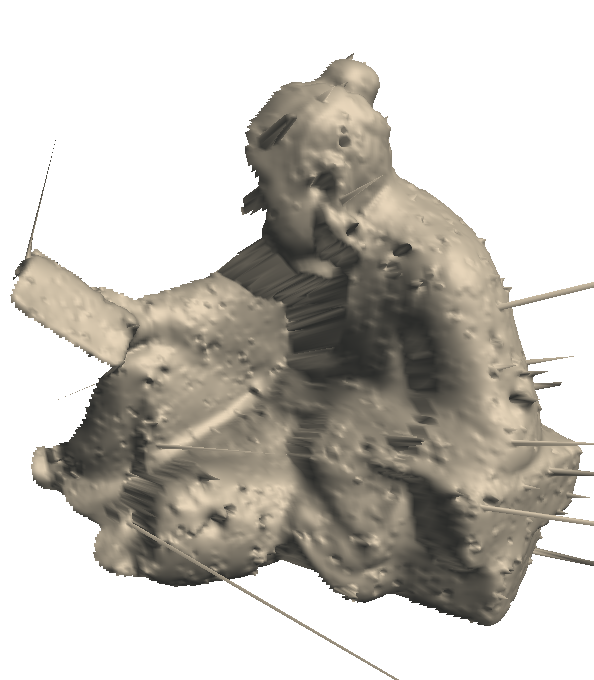}
& \includegraphics[width=\linewidth]{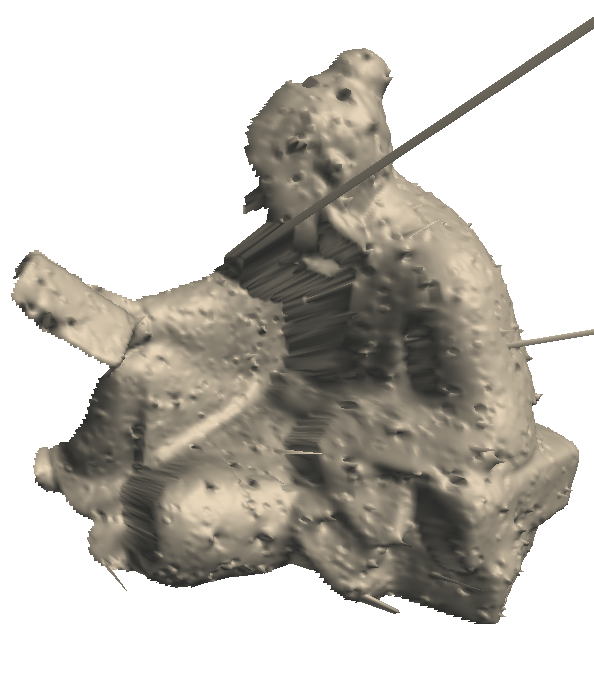}
& \includegraphics[width=\colwidth]{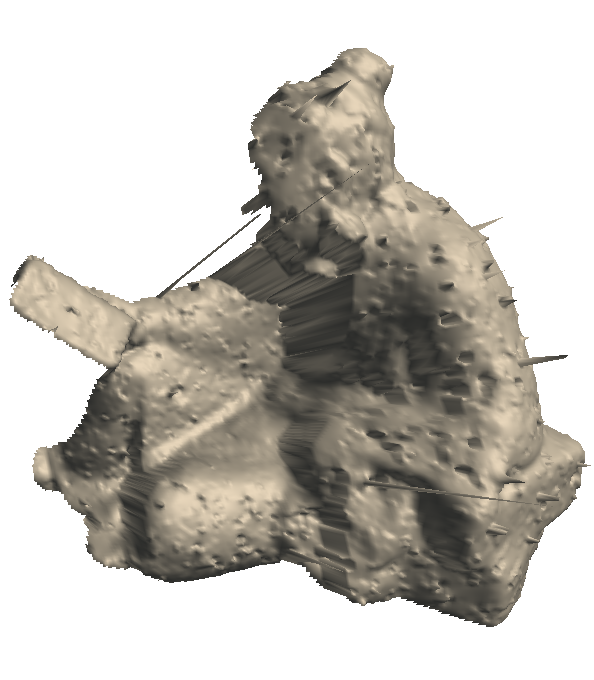}
\\[-1em]
& \includegraphics[width=\linewidth]{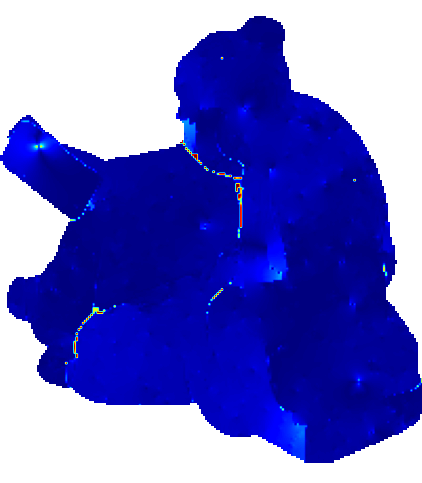}
& \includegraphics[width=\linewidth]{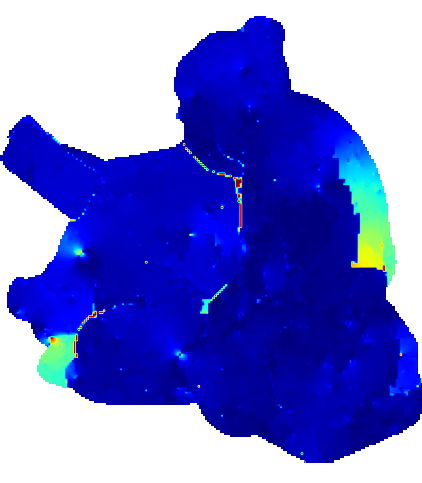}
& \includegraphics[width=\linewidth]{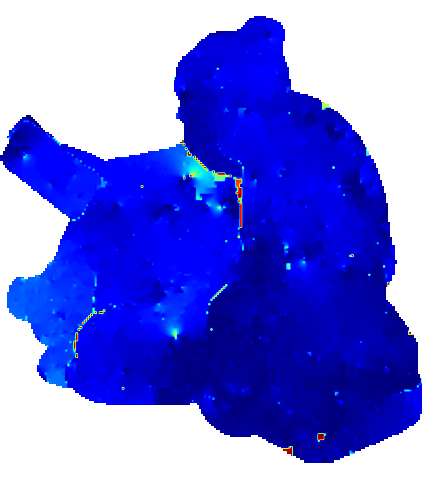}
& \includegraphics[width=\linewidth]{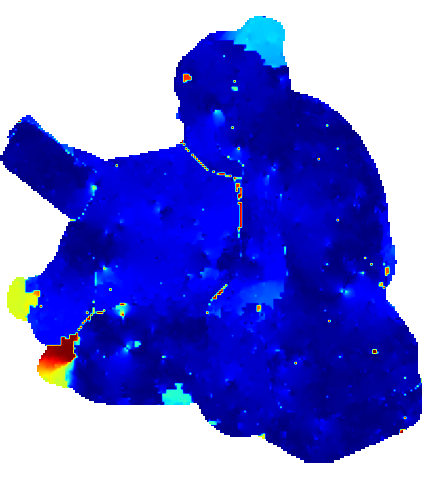}
& \includegraphics[width=\linewidth]{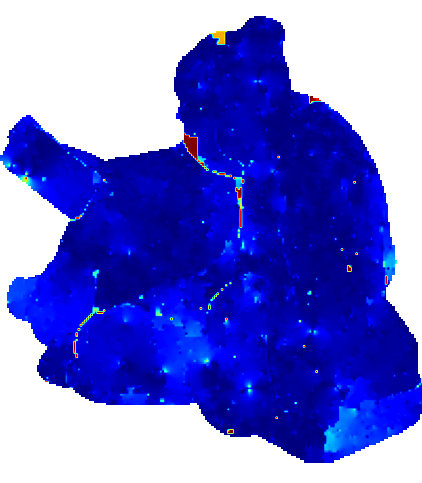}
& \includegraphics[width=\linewidth]{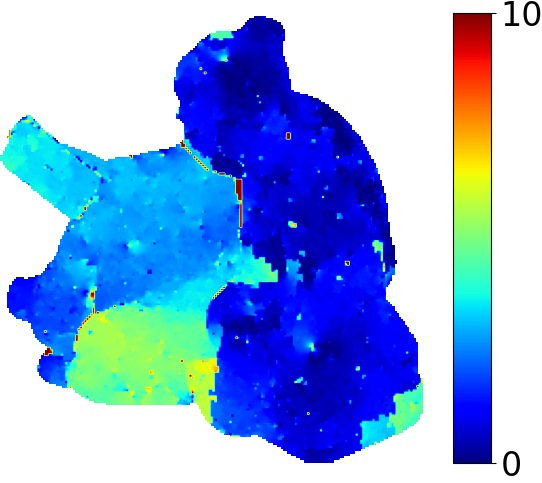}\\[-5pt]
& $\scriptsize{0.38}$ & $\scriptsize{0.72}$ & $\scriptsize{0.85}$ & $\scriptsize{0.91}$ & $\scriptsize{0.70}$ & $\textrm{ }\textrm{ }\textrm{ }\textrm{ }\textrm{ }\textrm{ }\textrm{ }\scriptsize{2.11}$\\
\noalign{\vskip -0.2em} 
\cline{2-7}
\noalign{\vskip 0.2em}
& \multicolumn{7}{c}{Normal averaging where $\boldsymbol{n_a}^\mathsf{T} \boldsymbol{\tau_a} > 0$}\tabularnewline
& \includegraphics[width=\linewidth]{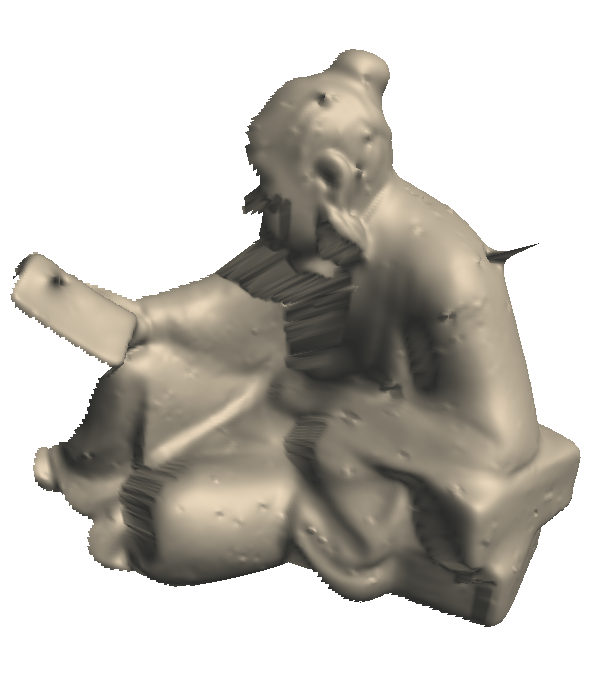}
& \includegraphics[width=\linewidth]{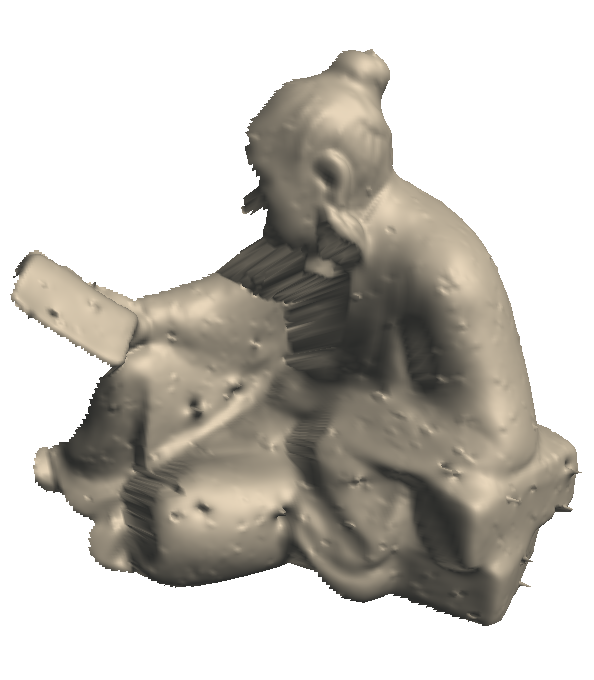}
& \includegraphics[width=\linewidth]{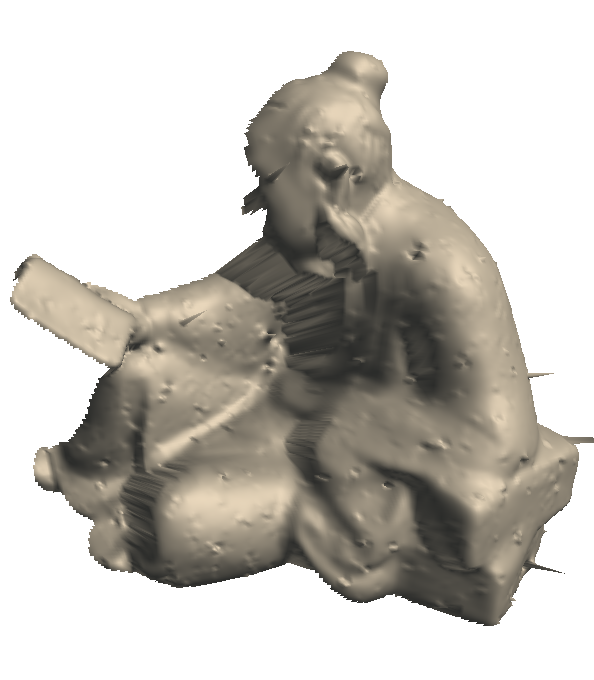}
& \includegraphics[width=\linewidth]{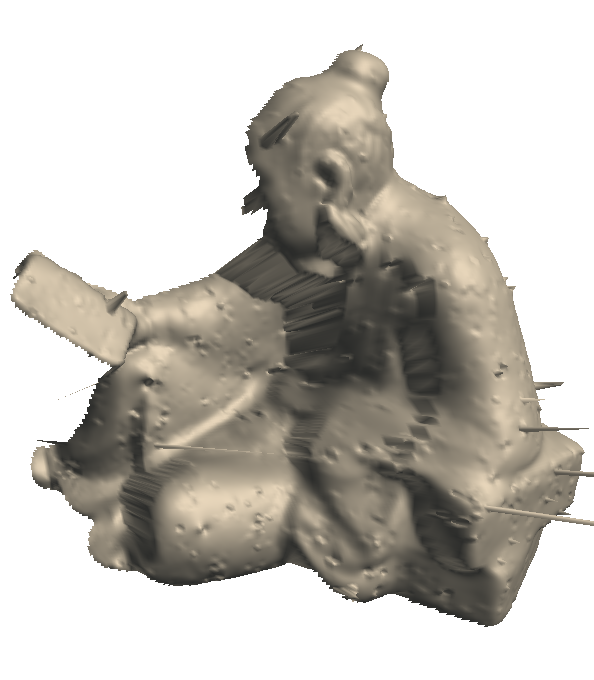}
& \includegraphics[width=\linewidth]{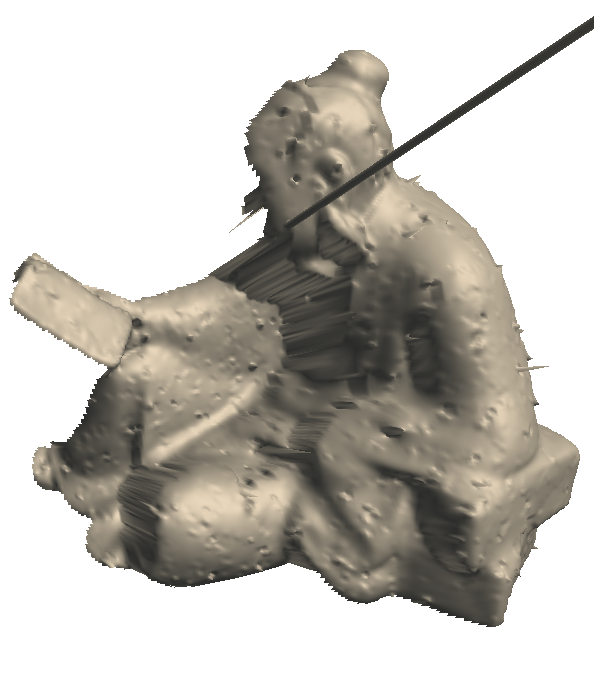}
& \includegraphics[width=\colwidth]{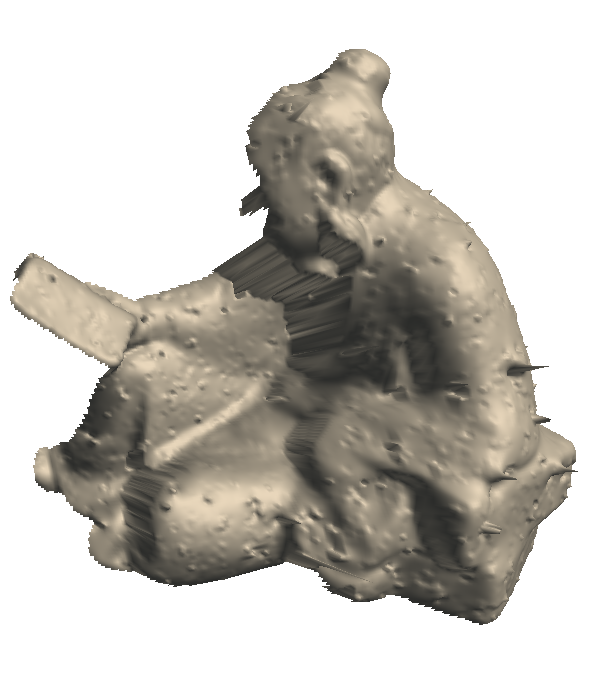}
\\[-1em]
& \includegraphics[width=\linewidth]{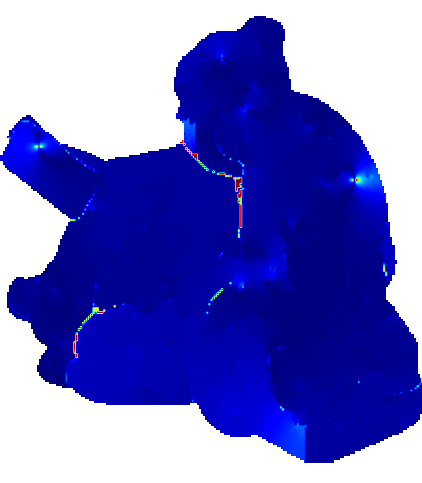}
& \includegraphics[width=\linewidth]{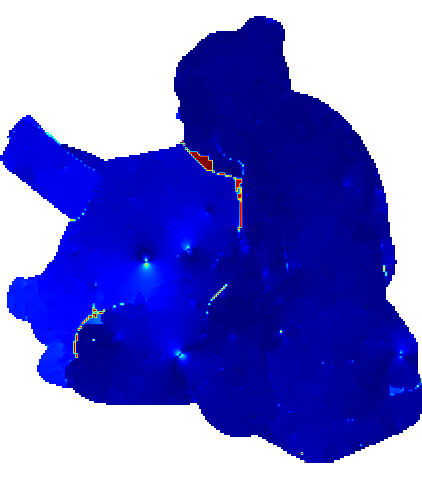}
& \includegraphics[width=\linewidth]{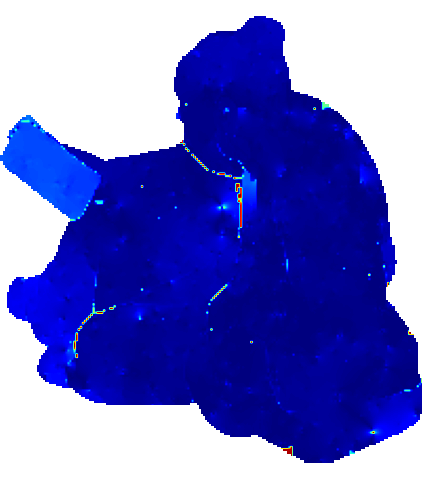}
& \includegraphics[width=\linewidth]{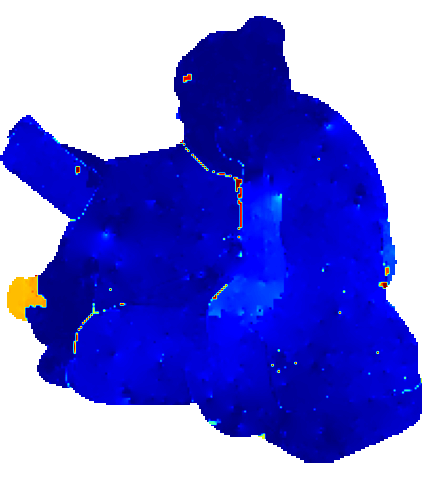}
& \includegraphics[width=\linewidth]{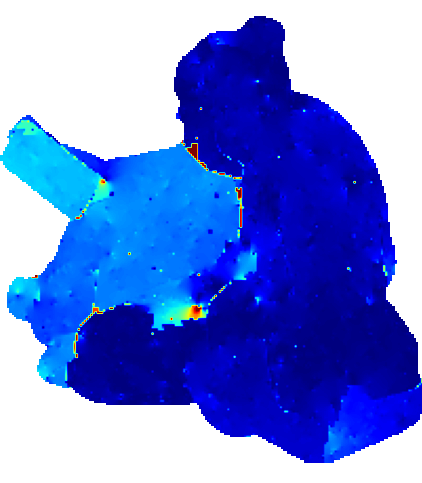}
& \includegraphics[width=\linewidth]{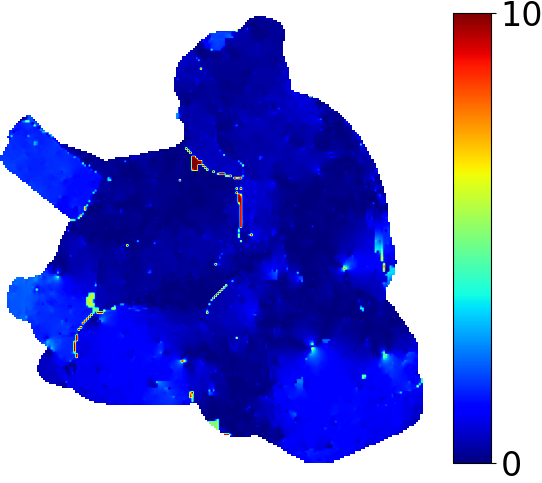}\\[-5pt]
& $\scriptsize{0.36}$ & $\scriptsize{0.50}$ & $\scriptsize{0.42}$ & $\scriptsize{0.81}$ & $\scriptsize{1.17}$ & $\textrm{ }\textrm{ }\textrm{ }\textrm{ }\textrm{ }\textrm{ }\textrm{ }\scriptsize{0.70}$\\
\noalign{\vskip -0.2em} 
\cline{2-7}
\noalign{\vskip 0.2em}
& \multicolumn{7}{c}{Normal averaging where $\boldsymbol{n_a}^\mathsf{T} \boldsymbol{\tau_a} > 0$ or relative change in $|\boldsymbol{n_a}^\mathsf{T} \boldsymbol{\tau_a}|>75\%$}\tabularnewline
& \includegraphics[width=\linewidth]{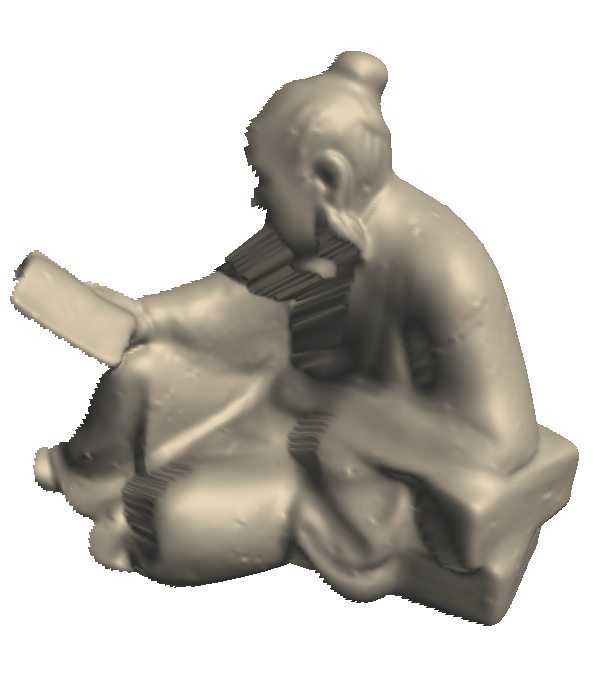}
& \includegraphics[width=\linewidth]{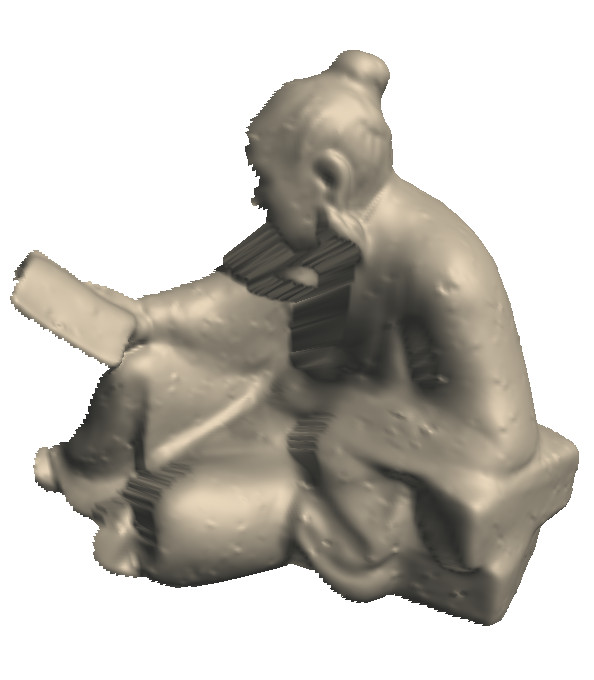}
& \includegraphics[width=\linewidth]{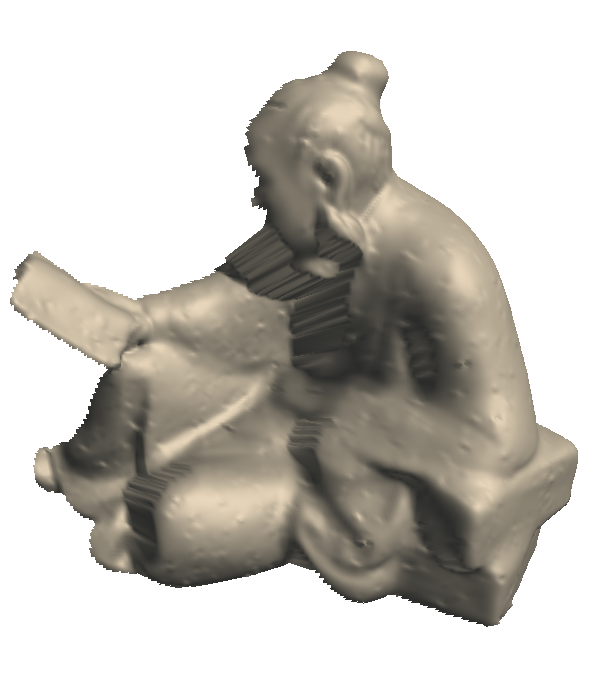}
& \includegraphics[width=\linewidth]{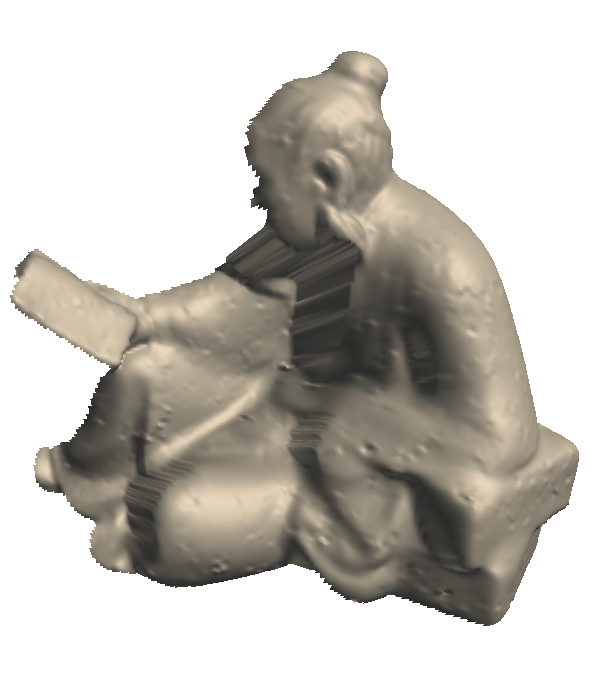}
& \includegraphics[width=\linewidth]{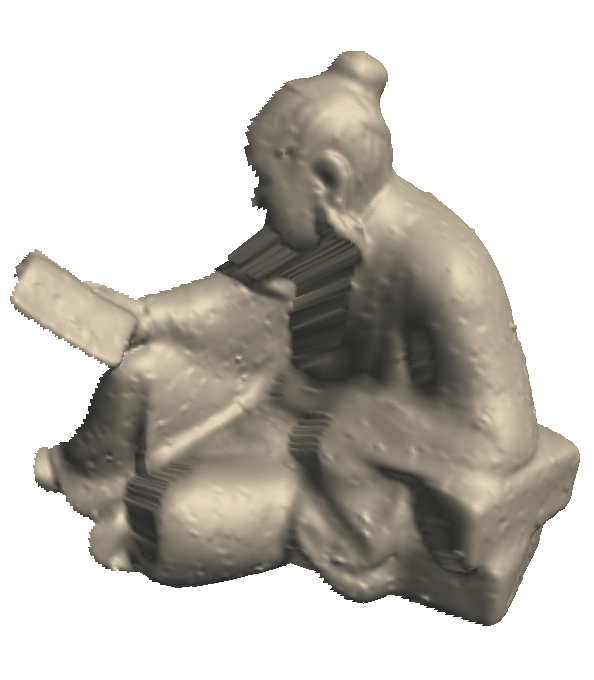}
& \includegraphics[width=\colwidth]{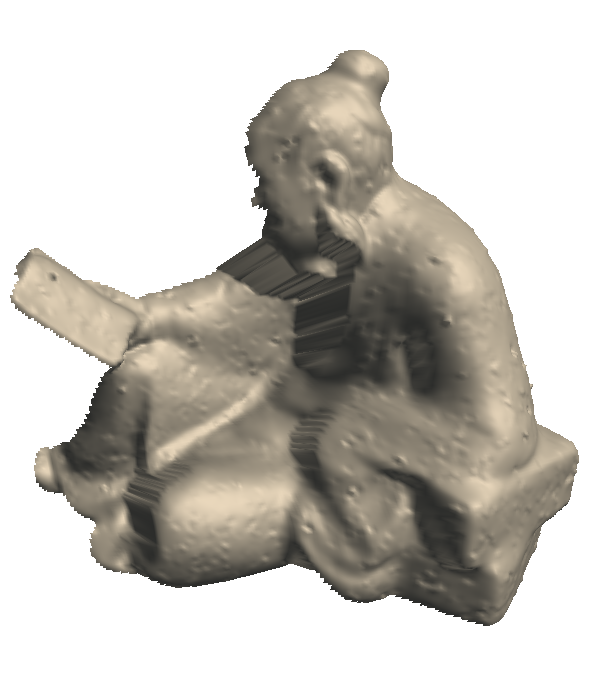}
\\[-1em]
& \includegraphics[width=\linewidth]{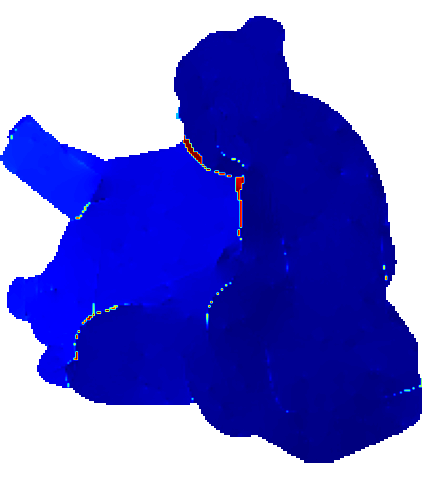}
& \includegraphics[width=\linewidth]{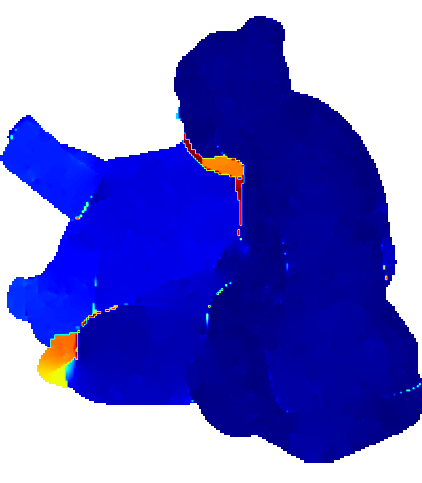}
& \includegraphics[width=\linewidth]{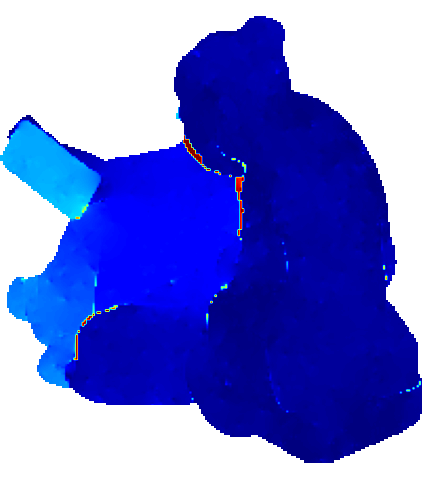}
& \includegraphics[width=\linewidth]{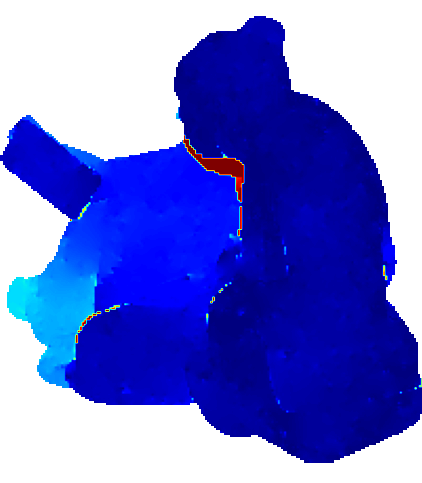}
& \includegraphics[width=\linewidth]{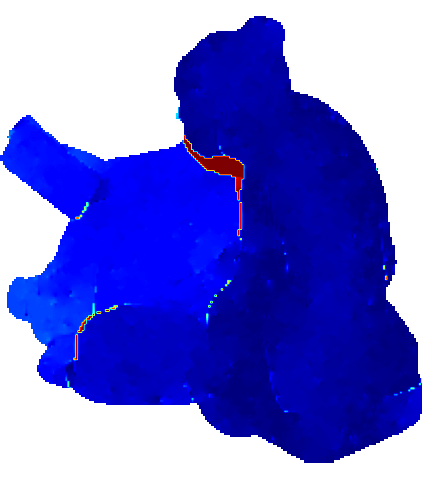}
& \includegraphics[width=\linewidth]{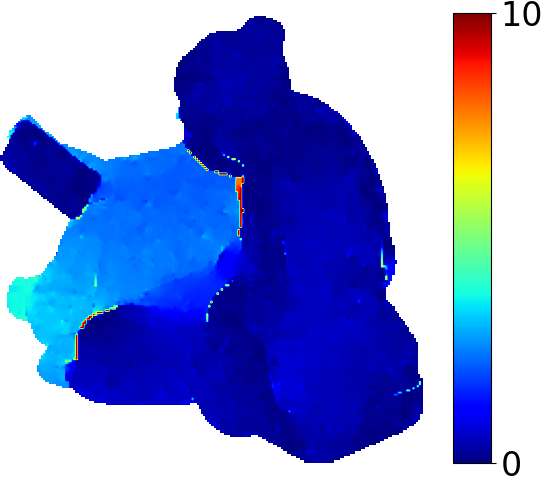}\\[-5pt]
& $\scriptsize{0.54}$ & $\scriptsize{0.67}$ & $\scriptsize{0.75}$ & $\scriptsize{0.84}$ & $\scriptsize{0.77}$ & $\textrm{ }\textrm{ }\textrm{ }\textrm{ }\textrm{ }\textrm{ }\textrm{ }\scriptsize{0.94}$\\
\noalign{\vskip -0.2em} 
\cline{2-7}
\noalign{\vskip 0.2em}
& \multicolumn{7}{c}{Normal averaging where $\boldsymbol{n_a}^\mathsf{T} \boldsymbol{\tau_a} > 0$, outlier filtering through $\boldsymbol{\tau_m}$}\tabularnewline
& \includegraphics[width=\linewidth]{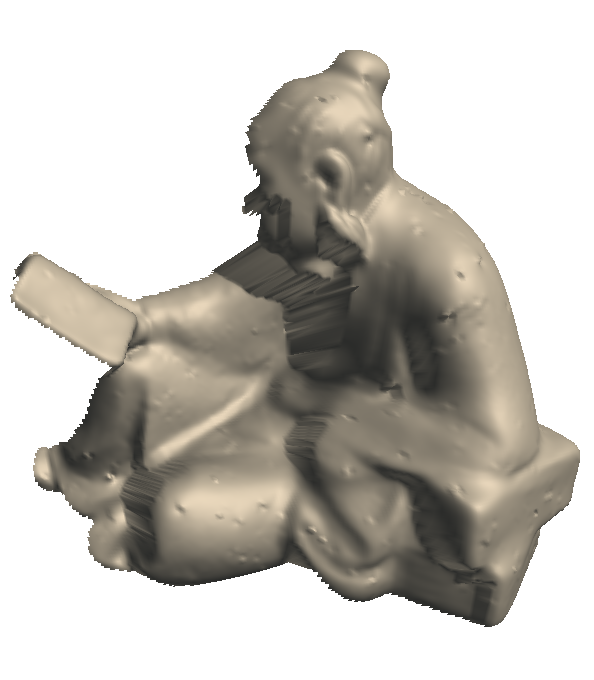}
& \includegraphics[width=\linewidth]{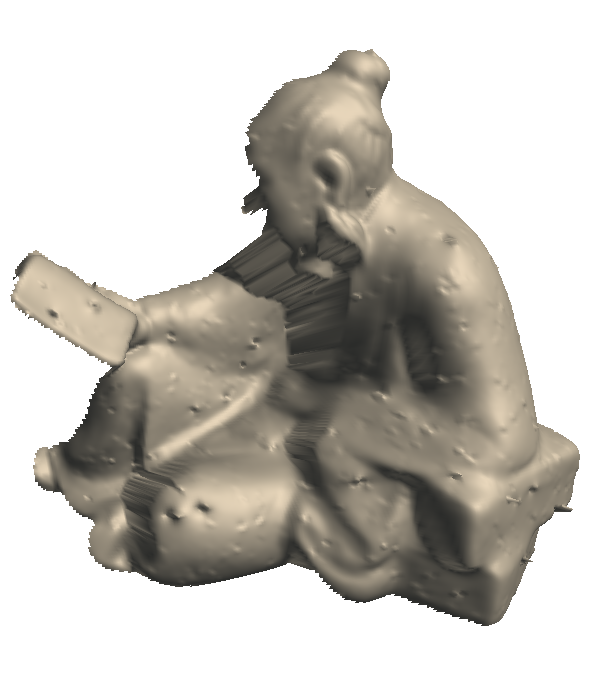}
& \includegraphics[width=\linewidth]{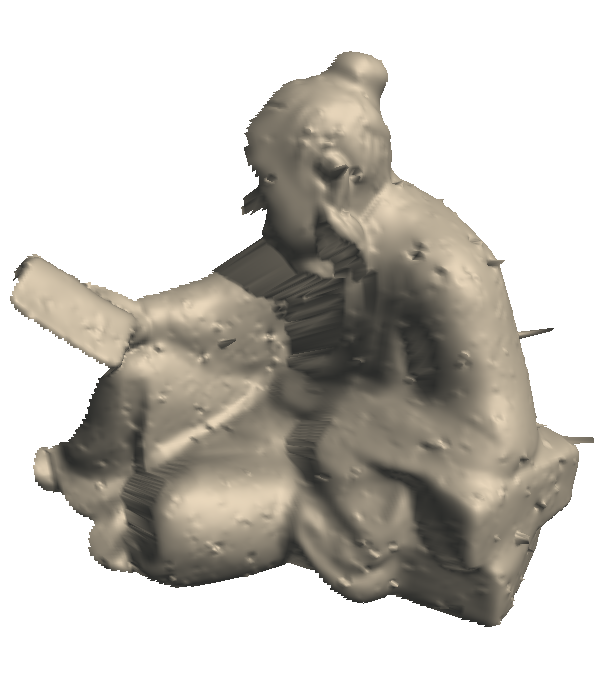}
& \includegraphics[width=\linewidth]{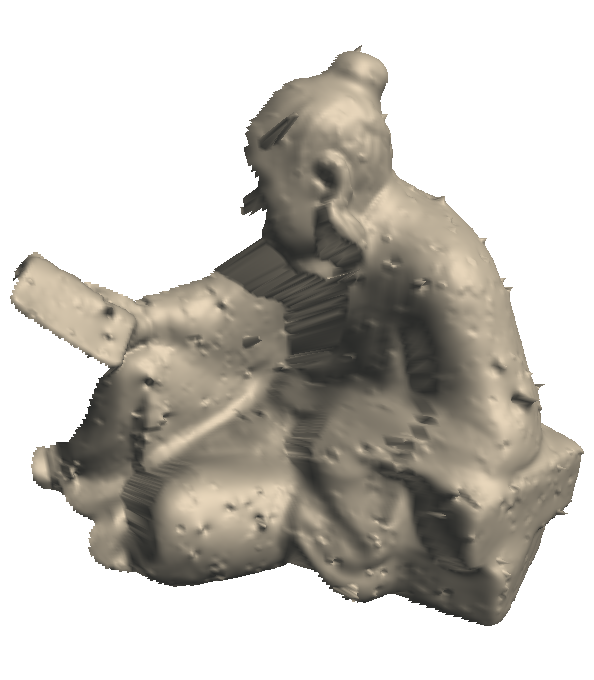}
& \includegraphics[width=\linewidth]{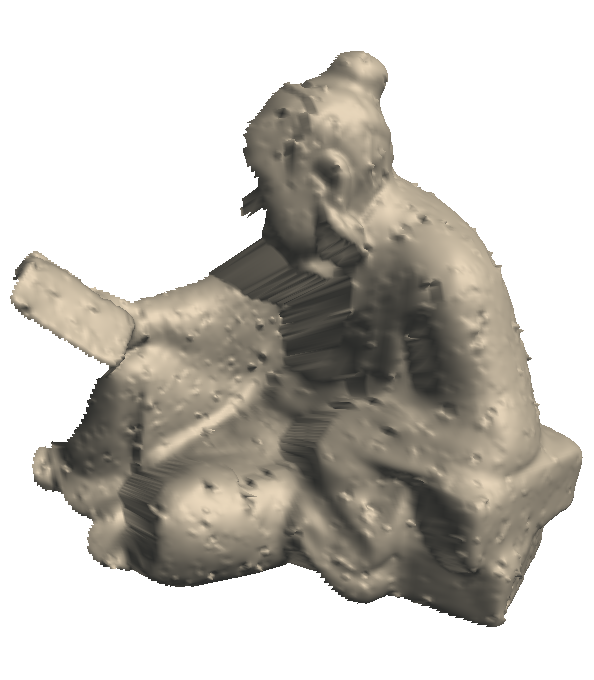}
& \includegraphics[width=\colwidth]{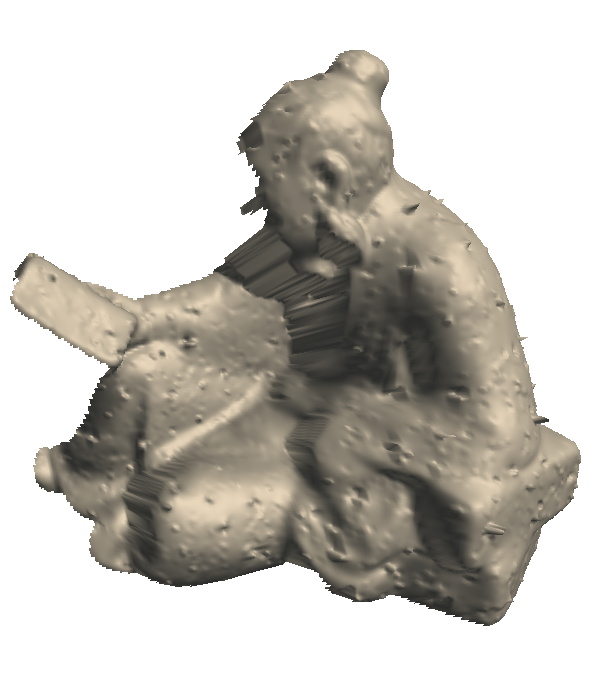}
\\[-1em]
& \includegraphics[width=\linewidth]{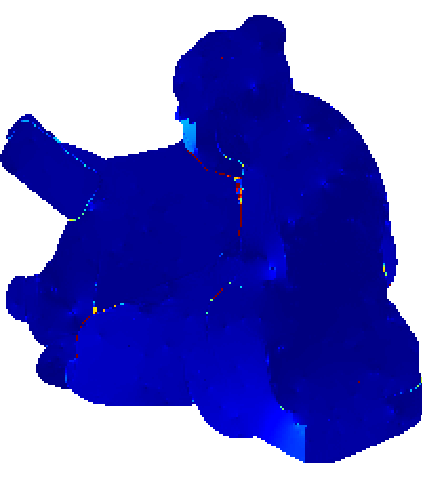}
& \includegraphics[width=\linewidth]{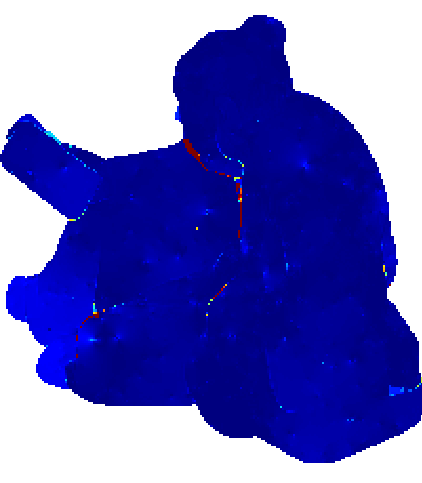}
& \includegraphics[width=\linewidth]{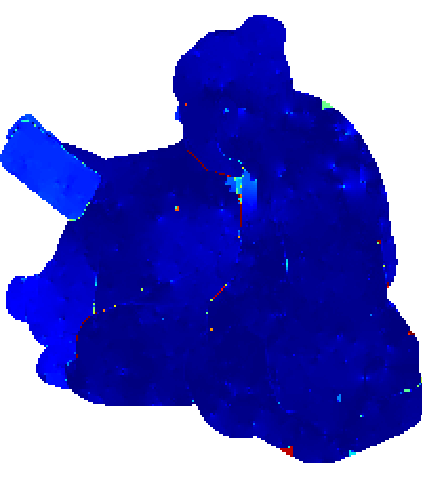}
& \includegraphics[width=\linewidth]{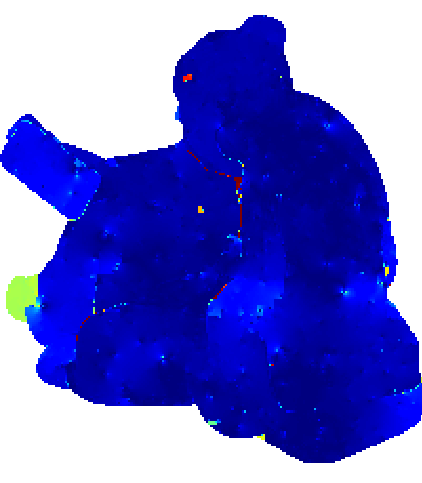}
& \includegraphics[width=\linewidth]{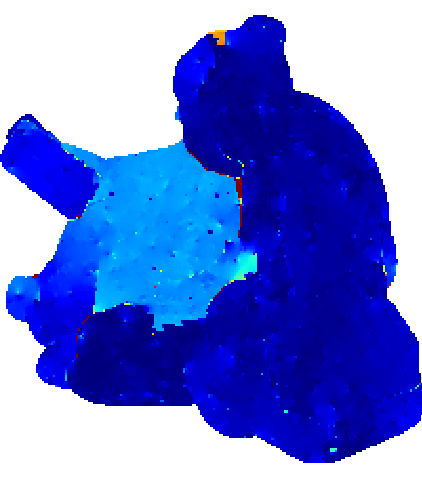}
& \includegraphics[width=\linewidth]{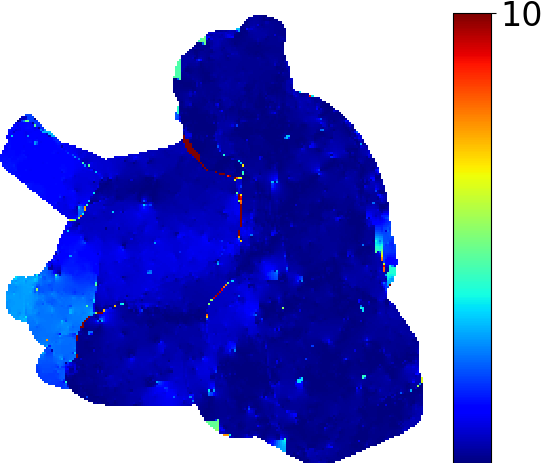}\\[-5pt]
& $\scriptsize{0.32}$ & $\scriptsize{0.29}$ & $\scriptsize{0.39}$ & $\scriptsize{0.54}$ & $\scriptsize{0.99}$ & $\textrm{ }\textrm{ }\textrm{ }\textrm{ }\textrm{ }\textrm{ }\textrm{ }\scriptsize{0.59}$\\
\end{tabular}
\addtolength{\tabcolsep}{4pt}
\caption{\textbf{Ablation on the effect of outliers, object} \texttt{harvest} \textbf{from the DiLiGenT~\cite{Shi2016DiLiGenT} dataset.} We introduce increasing amounts of outliers, for which we replace the surface normal with a randomly sampled unit-norm vector. For each variant, we show the reconstructed surface, the corresponding absolute depth error map, and its mean value (MADE, in $\mathrm{mm}$).}
\label{fig:ablation_noise_outliers}
\end{figure*}
\begin{figure*}[!t]
\centering
\def\colwidth{0.09\textwidth}
\def\colwidthb{0.12\textwidth}
\newcolumntype{M}[1]{>{\centering\arraybackslash}m{#1}}
\addtolength{\tabcolsep}{-4pt}
\begin{tabular}{m{0.7em} M{\colwidth} M{\colwidth} M{\colwidth} M{\colwidth} M{\colwidth}  m{\colwidthb} M{\colwidth} M{\colwidth} }
& \multicolumn{8}{c}{$\sigma$ of the noise (rotational angle), input normal maps, and $|\boldsymbol{n_a}^\mathsf{T} \boldsymbol{\tau_a}|$}\tabularnewline
&  & & $1^\circ$ & $2^\circ$ & $5^\circ$ & $\textrm{ }\textrm{ }\textrm{ }\textrm{ }\textrm{ }\textrm{ }\textrm{ }10^\circ$ \tabularnewline
& & & 
\includegraphics[width=\linewidth]{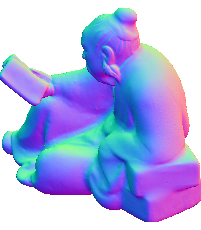} & 
\includegraphics[width=\linewidth]{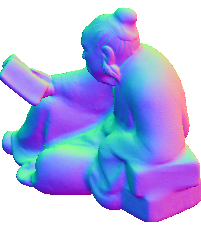} & 
\includegraphics[width=\linewidth]{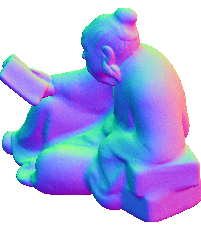} & 
\includegraphics[width=\colwidth]{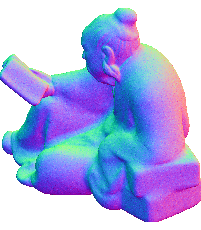} & \rule{\linewidth}{0cm} & \rule{\linewidth}{0cm} 
\\[-1em]
& & & 
\includegraphics[width=\linewidth]{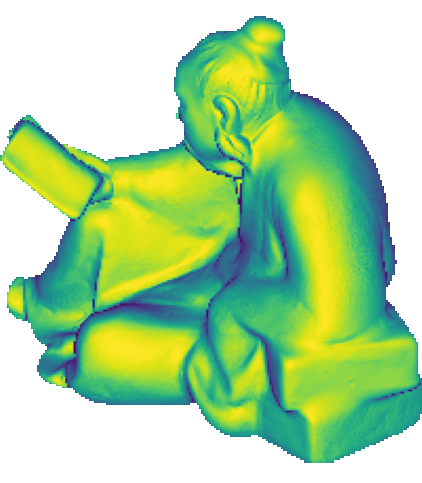} & 
\includegraphics[width=\linewidth]{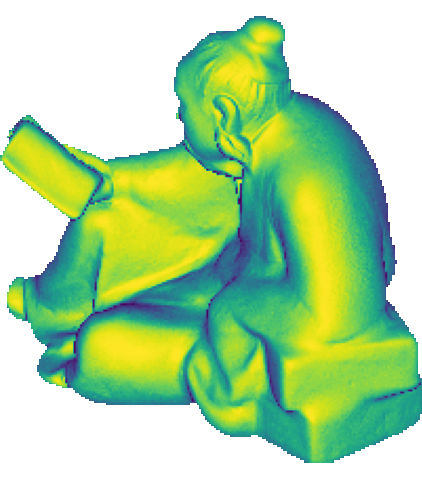} & 
\includegraphics[width=\linewidth]{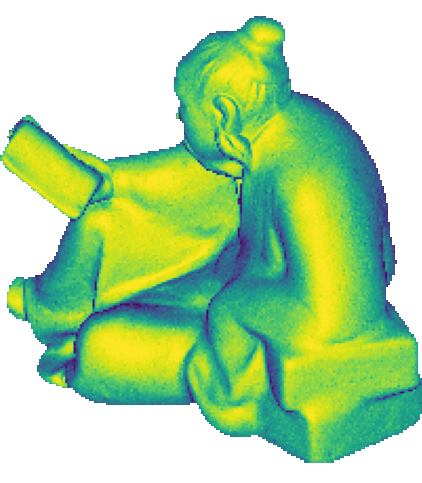} & 
\includegraphics[width=\linewidth]{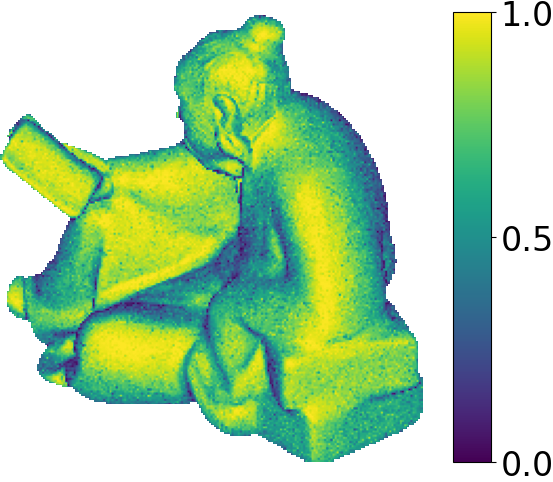} & \rule{\linewidth}{0cm} & \rule{\linewidth}{0cm} \tabularnewline
\noalign{\vskip -0.2em} 
\noalign{\global\savearrayrulewidth\arrayrulewidth} 
\noalign{\global\arrayrulewidth=1.2pt} 
\cline{4-7}
\noalign{\global\arrayrulewidth\savearrayrulewidth} 
\noalign{\vskip 0.2em}
& \multicolumn{8}{c}{No correction}\tabularnewline
& & & \includegraphics[width=\linewidth]{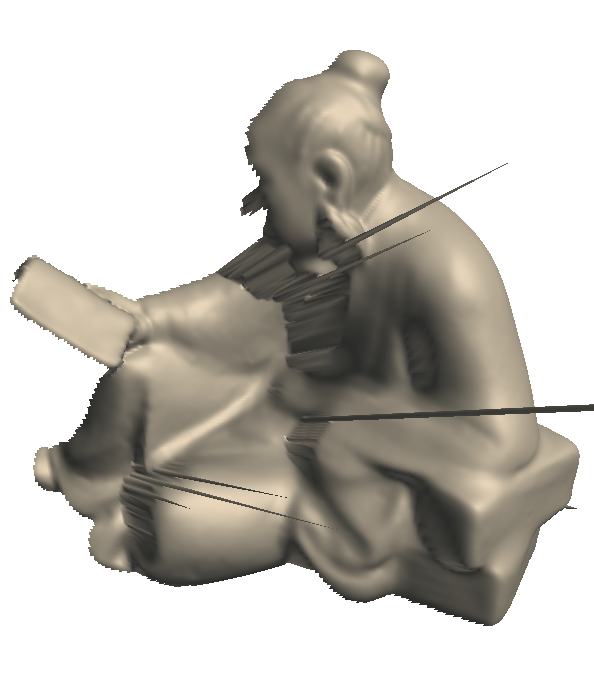}
& \includegraphics[width=\linewidth]{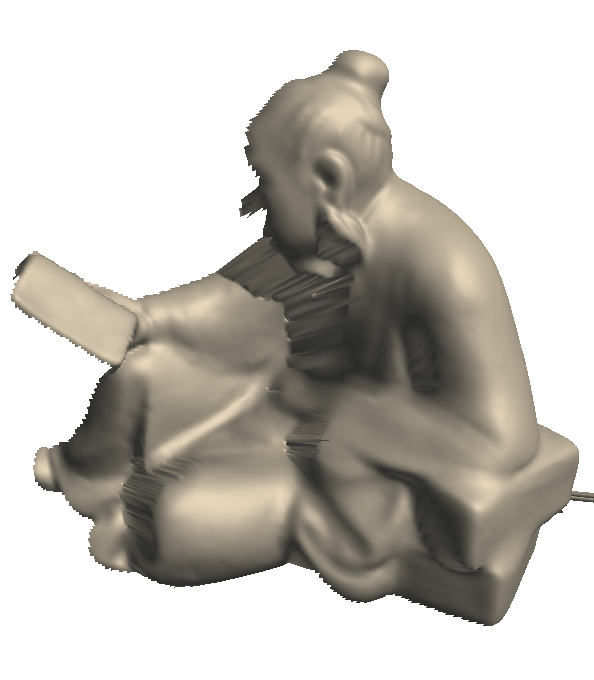}
& \includegraphics[width=\linewidth]{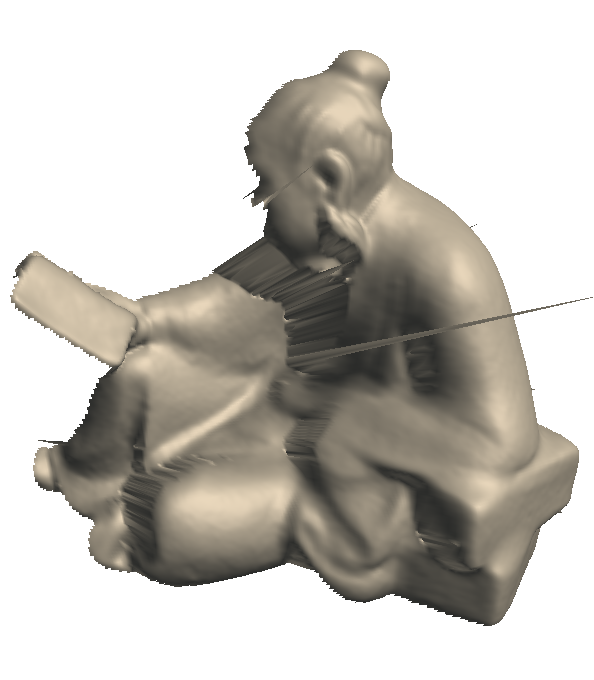}
& \includegraphics[width=\colwidth]{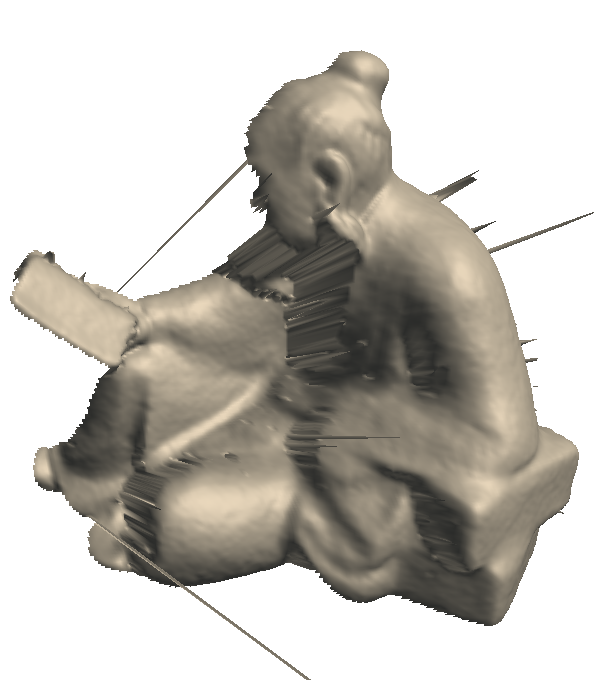}
\\[-1em]
& & & \includegraphics[width=\linewidth]{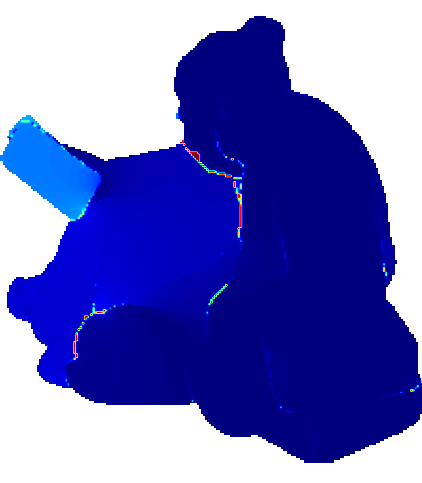}
& \includegraphics[width=\linewidth]{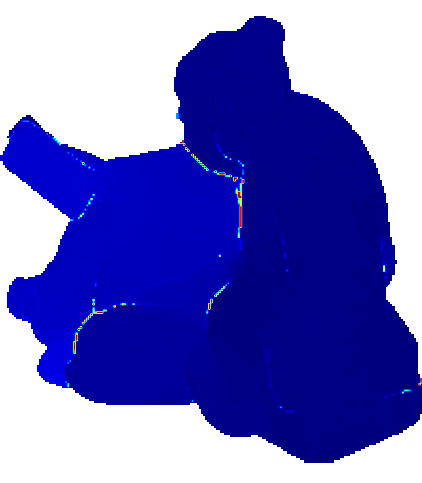}
& \includegraphics[width=\linewidth]{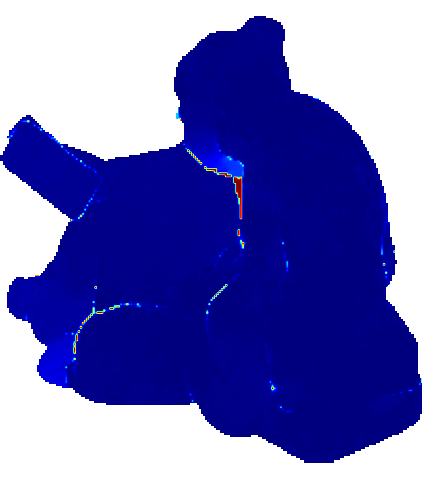}
& \includegraphics[width=\linewidth]{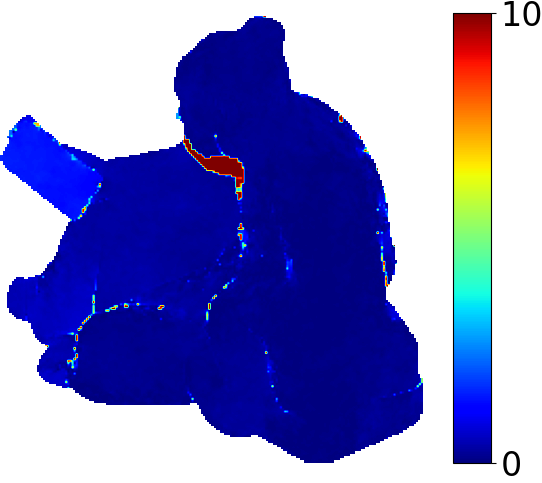}\\[-5pt]
& & & $\scriptsize{0.37}$ & $\scriptsize{0.31}$ & $\scriptsize{0.20}$ & $\textrm{ }\textrm{ }\textrm{ }\textrm{ }\textrm{ }\textrm{ }\textrm{ }\scriptsize{0.51}$\\
\noalign{\vskip -0.2em} 
\cline{4-7}
\noalign{\vskip 0.2em}
& \multicolumn{8}{c}{Normal averaging where $\boldsymbol{n_a}^\mathsf{T} \boldsymbol{\tau_a} > 0$}\tabularnewline
& & & \includegraphics[width=\linewidth]{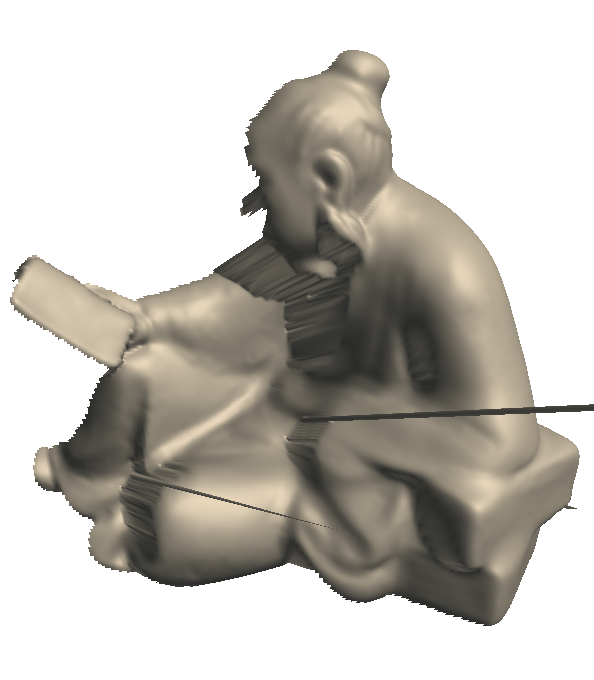}
& \includegraphics[width=\linewidth]{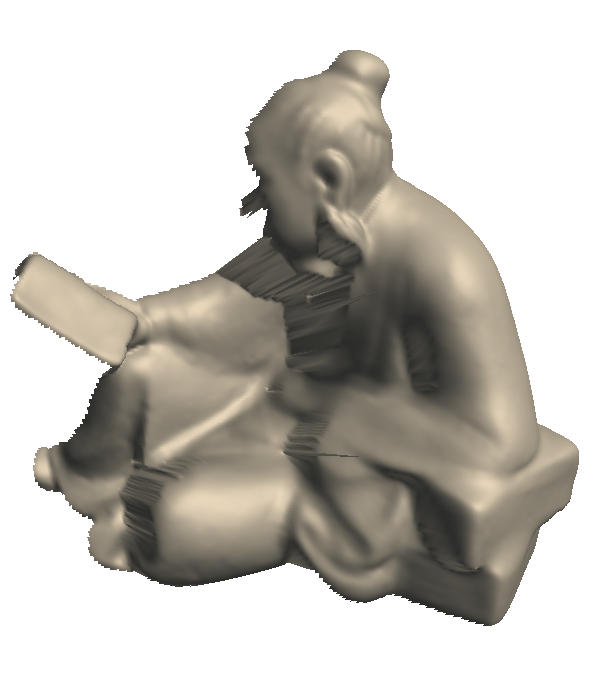}
& \includegraphics[width=\linewidth]{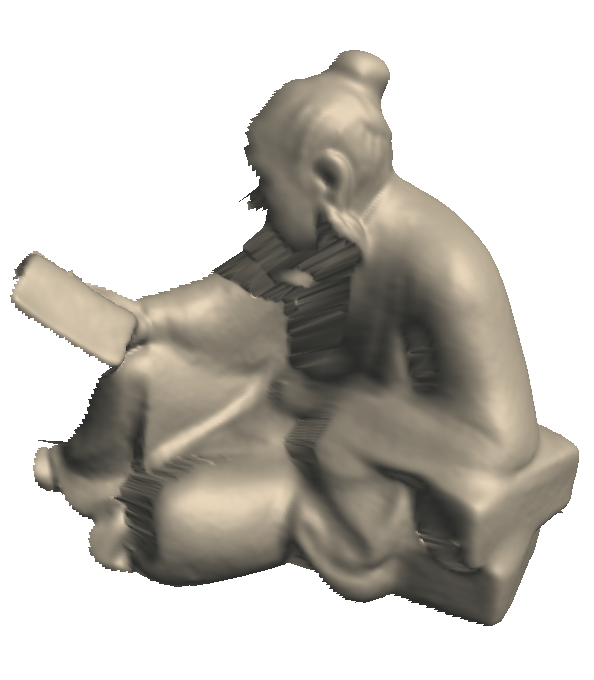}
& \includegraphics[width=\colwidth]{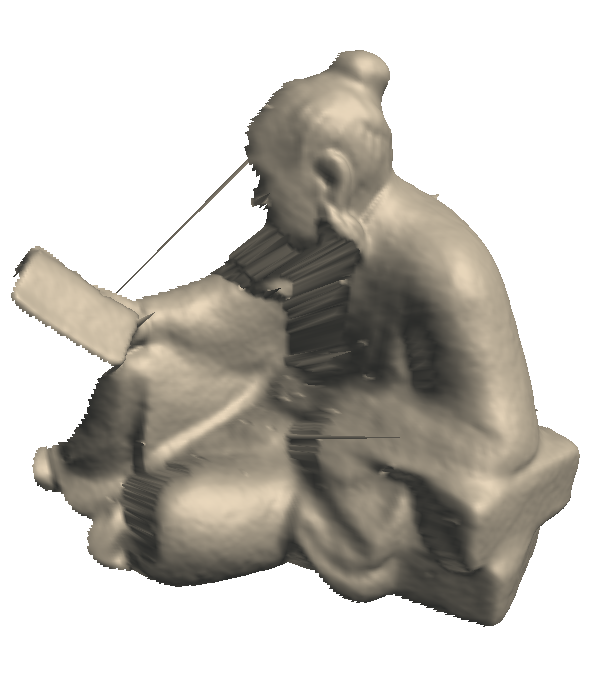}
\\[-1em]
& & & \includegraphics[width=\linewidth]{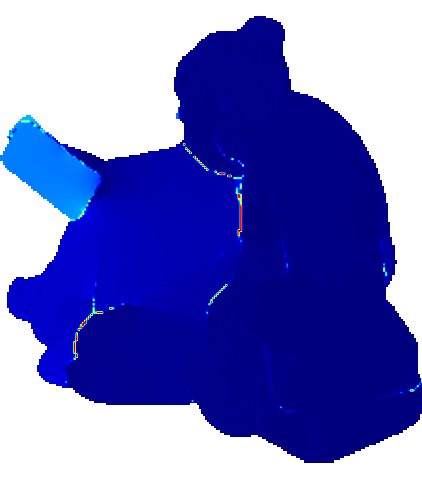}
& \includegraphics[width=\linewidth]{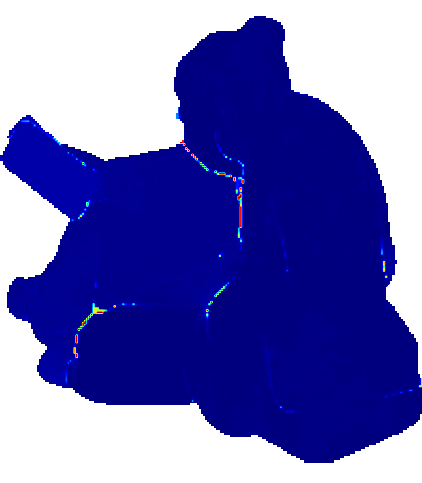}
& \includegraphics[width=\linewidth]{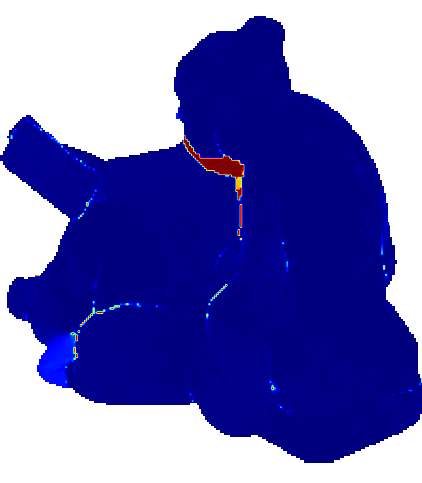}
& \includegraphics[width=\linewidth]{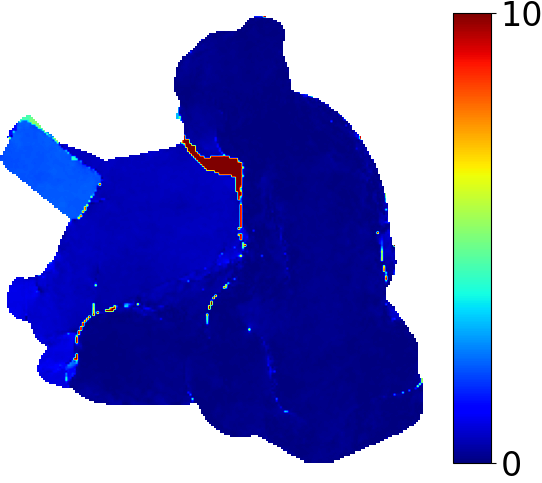}\\[-5pt]
& & & $\scriptsize{0.36}$ & $\scriptsize{0.15}$ & $\scriptsize{0.21}$ & $\textrm{ }\textrm{ }\textrm{ }\textrm{ }\textrm{ }\textrm{ }\textrm{ }\scriptsize{0.54}$\\
\noalign{\vskip -0.2em} 
\cline{4-7}
\noalign{\vskip 0.2em}
&\multicolumn{8}{c}{Normal averaging where $\boldsymbol{n_a}^\mathsf{T} \boldsymbol{\tau_a} > 0$ or relative change in $|\boldsymbol{n_a}^\mathsf{T} \boldsymbol{\tau_a}|>75\%$}\tabularnewline
& & & \includegraphics[width=\linewidth]{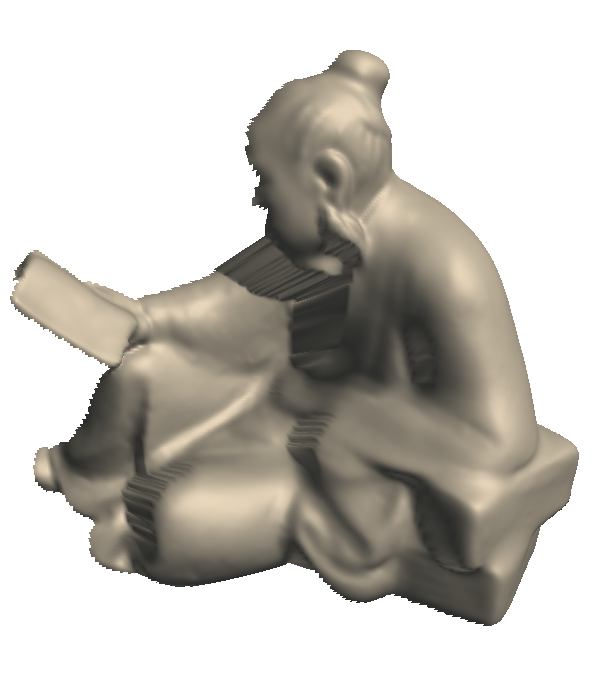}
& \includegraphics[width=\linewidth]{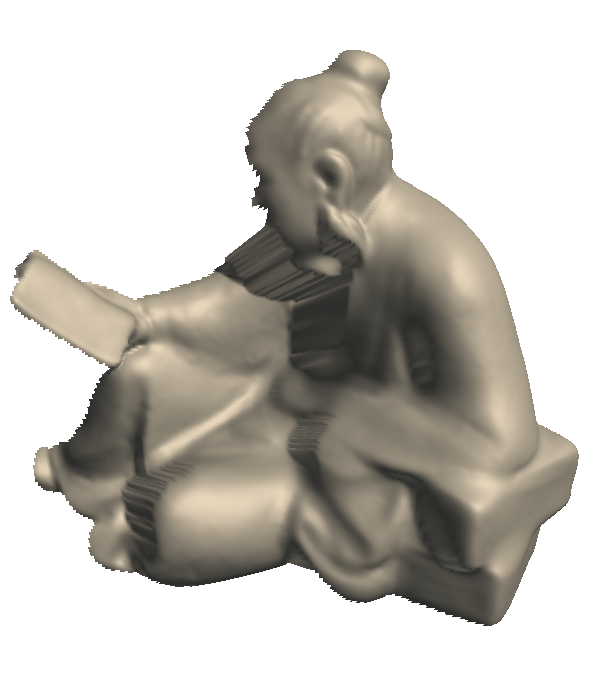}
& \includegraphics[width=\linewidth]{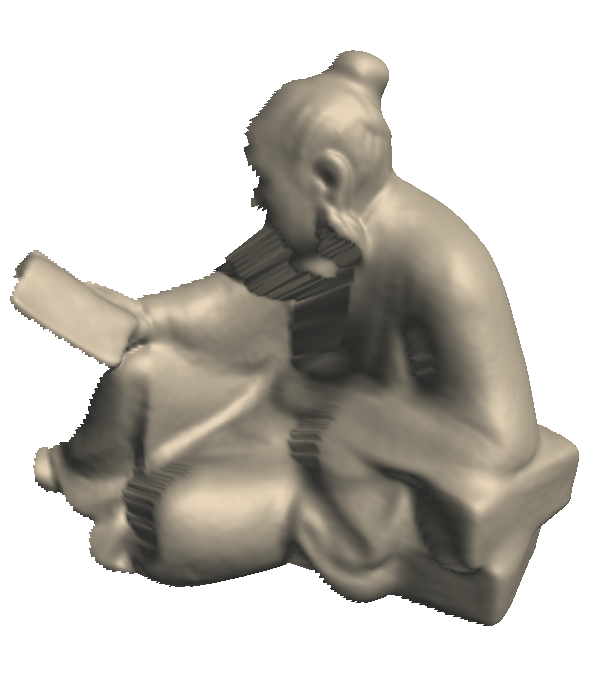}
& \includegraphics[width=\colwidth]{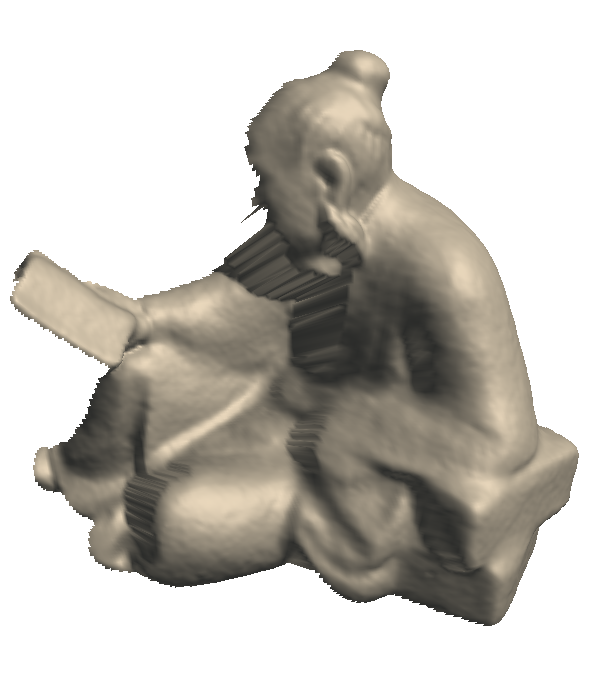}
\\[-1em]
& & & \includegraphics[width=\linewidth]{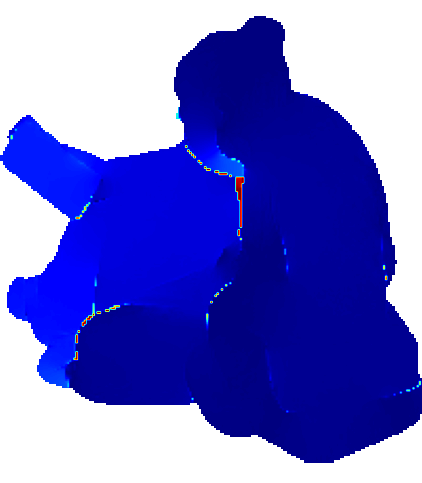}
& \includegraphics[width=\linewidth]{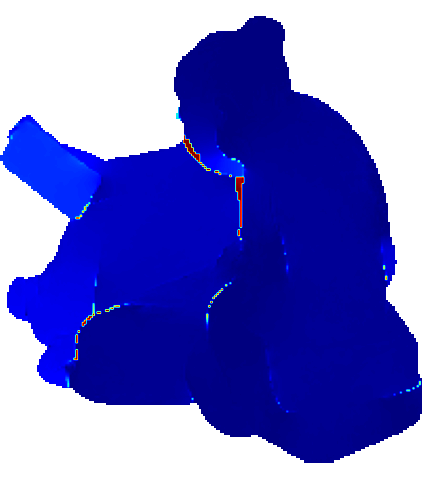}
& \includegraphics[width=\linewidth]{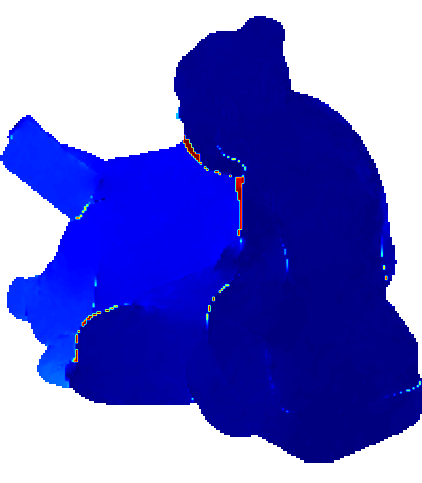}
& \includegraphics[width=\linewidth]{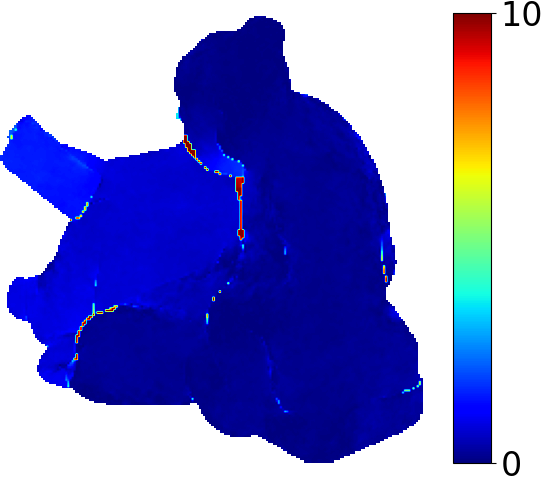}\\[-5pt]
& & & $\scriptsize{0.49}$ & $\scriptsize{0.44}$ & $\scriptsize{0.55}$ & $\textrm{ }\textrm{ }\textrm{ }\textrm{ }\textrm{ }\textrm{ }\textrm{ }\scriptsize{0.47}$\\
\noalign{\vskip -0.2em} 
\cline{4-7}
\noalign{\vskip 0.2em}
&\multicolumn{8}{c}{Normal averaging where $\boldsymbol{n_a}^\mathsf{T} \boldsymbol{\tau_a} > 0$, outlier filtering through $\boldsymbol{\tau_m}$}\tabularnewline
& & & \includegraphics[width=\linewidth]{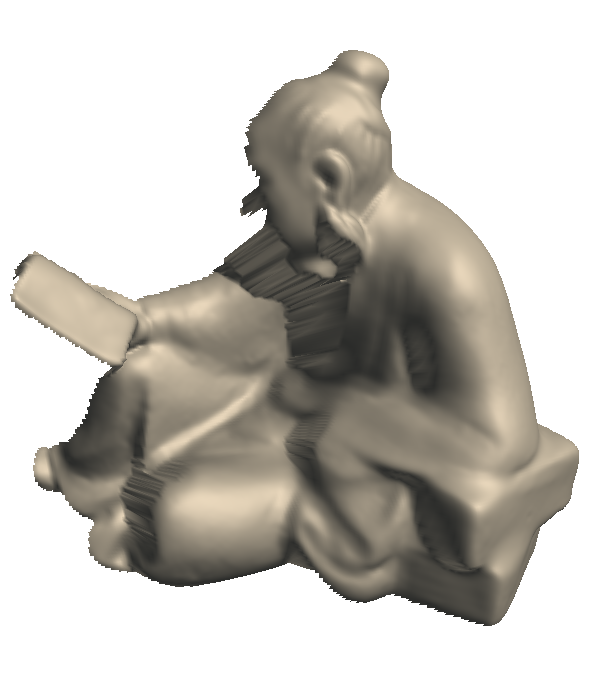}
& \includegraphics[width=\linewidth]{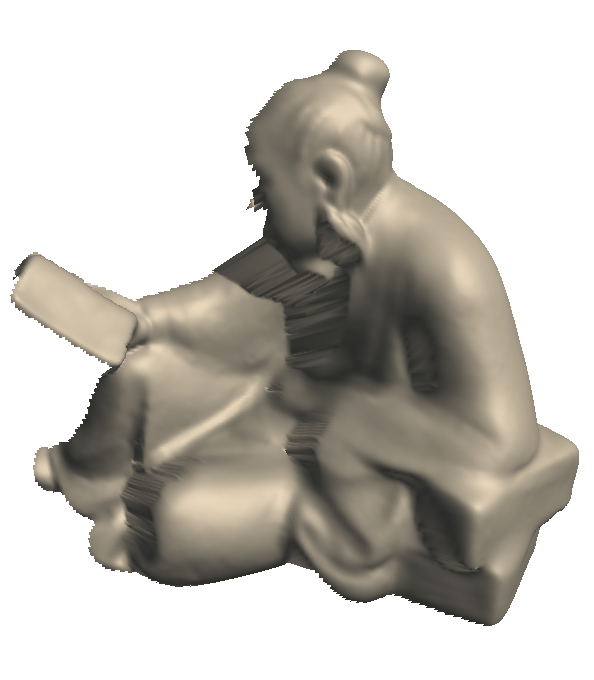}
& \includegraphics[width=\linewidth]{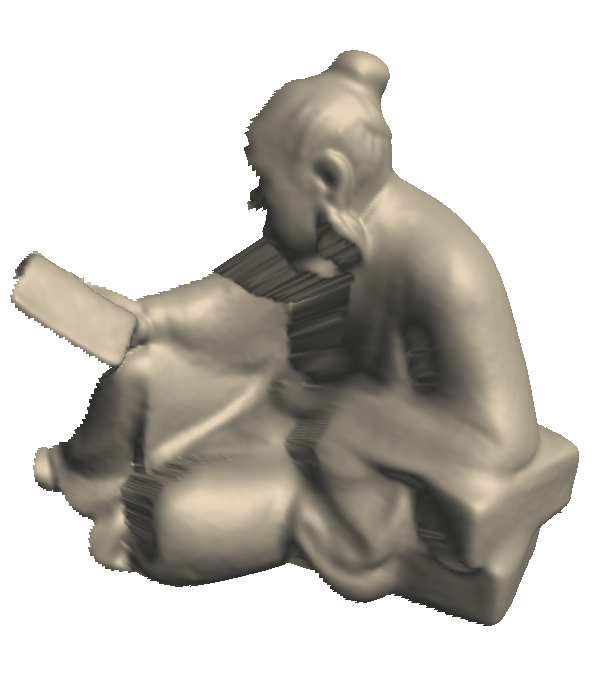}
& \includegraphics[width=\colwidth]{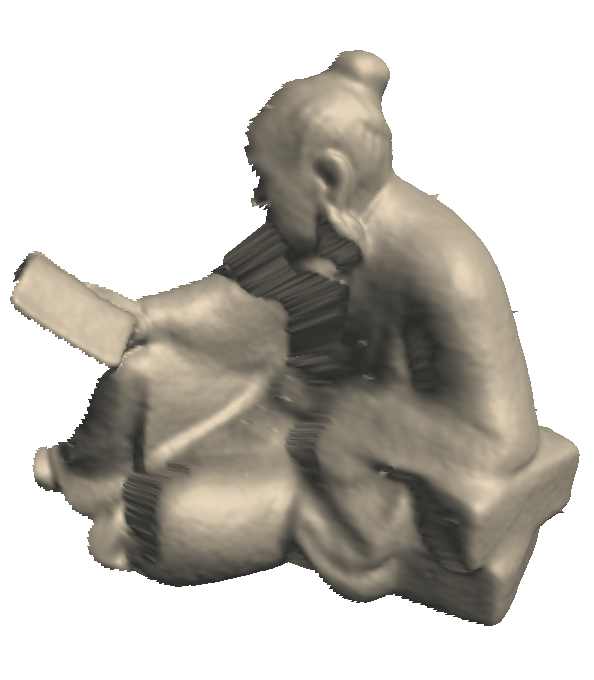}
\\[-1em]
& & & \includegraphics[width=\linewidth]{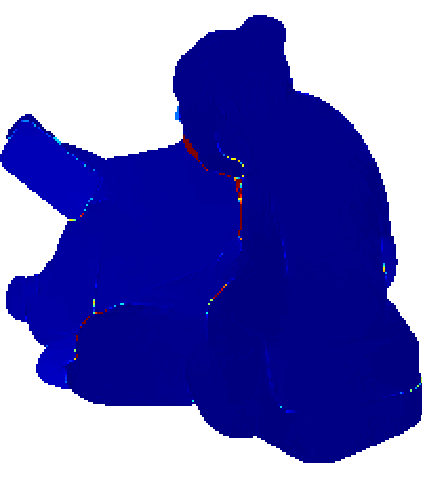}
& \includegraphics[width=\linewidth]{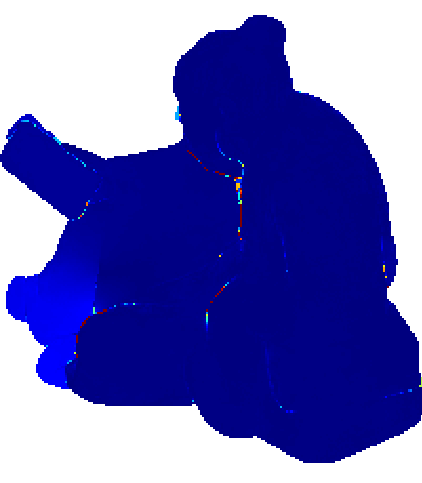}
& \includegraphics[width=\linewidth]{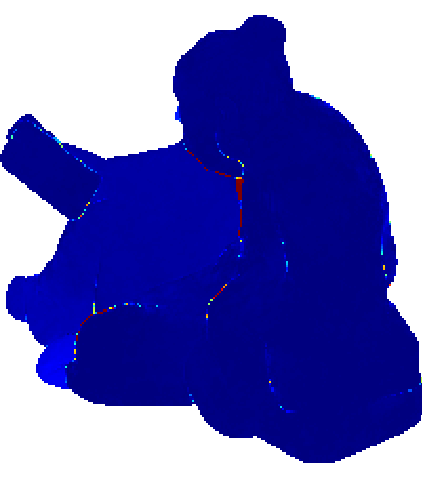}
& \includegraphics[width=\linewidth]{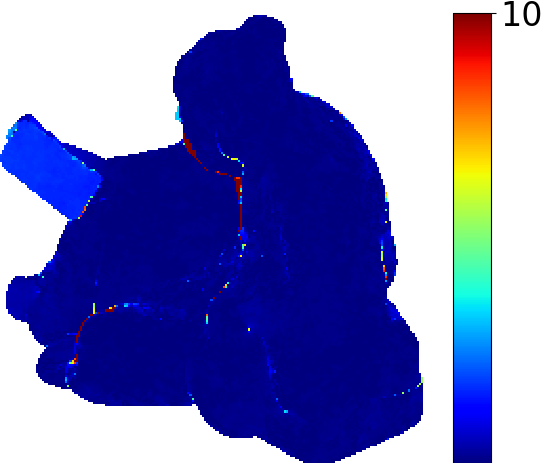}\\[-5pt]
& & & $\scriptsize{0.22}$ & $\scriptsize{0.18}$ & $\scriptsize{0.19}$ & $\textrm{ }\textrm{ }\textrm{ }\textrm{ }\textrm{ }\textrm{ }\textrm{ }\scriptsize{0.25}$\\
\end{tabular}
\addtolength{\tabcolsep}{4pt}
\caption{\textbf{Ablation on the effect of rotational noise, object} \texttt{harvest} \textbf{from the DiLiGenT~\cite{Shi2016DiLiGenT} dataset.} We perturb the surface normals
at each pixel,
rotating them
around
randomly sampled axes by angles sampled from Gaussian distributions with increasingly larger standard deviations. For each variant, we show the reconstructed surface, the corresponding absolute depth error map, and its mean value (MADE, in $\mathrm{mm}$).}
\label{fig:ablation_noise_rotation}
\end{figure*}
In this Section, we investigate the robustness of our method to noise in the input normal map.

Similarly to previous methods~\cite{Cao2022BiNI_supplementary}, we simulate the presence of outlier normals by replacing the original normals with randomly sampled unit vectors, 
with
different
percentages
of
sampled pixels.
\Cref{fig:ablation_noise_outliers} shows that without pre-processing the normal maps, our method can reconstruct most of the underlying surface, but suffers from the presence of spike artifacts and non-smooth effects on the surface (second block from the top in \cref{fig:ablation_noise_outliers}).
We note, however, that a large part of the outliers can and should be detected, because they correspond to physically impossible normals. In particular, as previously observed both in the main paper and in \cref{sec_suppl:analysis_gamma_factor}, a necessary condition for the surface to be observable at one point $\boldsymbol{p_a}$ is that the dot product $\boldsymbol{n_a}^\mathsf{T} \boldsymbol{\tau_a}$ at the corresponding pixel $a$ is negative. We observe that enforcing this condition by applying an averaging filter to the normals at pixels where $\boldsymbol{n_a}^\mathsf{T} \boldsymbol{\tau_a} > 0$ results in a reduction of the amount of spike artifacts (third block from the top in \cref{fig:ablation_noise_outliers}). We additionally note that the presence of outliers can also be detected by inspecting the distribution of
$\boldsymbol{n_a}^\mathsf{T} \boldsymbol{\tau_a}$ or of its absolute value:
while in a natural surface
these quantities vary
continuously
across the surface with the exception of
boundary regions,
for the perturbed normal maps salt-and-pepper noise can be observed in correspondence to the outliers (\cf. second row in the top block of \cref{fig:ablation_noise_outliers}).
We
verify that applying
average filtering
also to pixels where $|\boldsymbol{n_a}^\mathsf{T} \boldsymbol{\tau_a}|$ deviates
significantly
from the mean value in its neighborhood further mitigates the effect of the outliers, removing spike artifacts and
recovering the smoothness of the surface (\cf lowermost block in \cref{fig:ablation_noise_outliers}).

While the above test effectively highlights the impact of outliers on the reconstruction,
we argue that
it
does not fully accurately reflect the statistical characteristics of noise emerging in real-world normal maps, in particular those predicted by learning-based methods. To provide an additional evaluation of the robustness
of our method
under
noise in the input normals,
we perturb the
surface normals
by rotating them around an axis that we randomly sample for each pixel,
with an angle of rotation
that we sample from a Gaussian distribution. \Cref{fig:ablation_noise_rotation} shows the results of this ablation, where we vary the standard deviation of the Gaussian distribution between $1$ and $10$ degrees. Similarly to the experiment with outliers, providing the raw normal map as input to our method results in spike artifacts (second block from the top in \cref{fig:ablation_noise_rotation}). Noticeably, however, most of these artifacts can be corrected by average filtering of the pixels with invalid normals alone (third block from the top in \cref{fig:ablation_noise_rotation}), showing that 
physically impossible normals
constitute the main factor behind these artifacts. As in the case with outliers, additionally filtering pixels where $|\boldsymbol{n_a}^\mathsf{T} \boldsymbol{\tau_a}|$ deviates largely from the mean value in the pixels' neighborhood allows
further reducing
artifacts and removing spikes (lowermost block in \cref{fig:ablation_noise_rotation}).

\noindent\textbf{Outlier filtering through $\boldsymbol{\tau_m}$.}
The spike artifacts resulting from the outlier normals have been identified in the literature as consequences of a type of \textit{Gibbs phenomenon}~\cite{Cao2021NormalIntegrationInversePlaneFitting, Heep2024AdaptiveScreenSpaceMeshingApproach}.
A closer analysis of the terms of our formulation reveals that 
such artifacts arise at outlier pixels where the terms $\boldsymbol{n_i}^\mathsf{T}\boldsymbol{\tau_j}$, for $(i, j)\in\{(a, a), (a, m), (b, b), (b, m)\}$, are either greater than $0$ or have small magnitude, \ie, 
$\boldsymbol{n_i}^\mathsf{T}\boldsymbol{\tau_j} > 0$ or $|\boldsymbol{n_i}^\mathsf{T}\boldsymbol{\tau_j}| \approx 0$.
In the latter case, in particular, the term $\omega_{b\rightarrow a}$, which depends on the multiplication of two such terms both in its numerator and its denominator, 
can significantly deviate from $1$.
This, in turn,
results in
$z_a \gg z_b$ or $z_a \ll z_b$ through
\eqref{eq:ours_formulation}
and thus
introduces
very large discontinuities that
imbalance
the optimization.

Crucially, our method offers a natural way to handle these outliers by controlling the ray direction 
$\boldsymbol{\tau_m} = \boldsymbol{\tau_a} + \lambda_m\cdot(\boldsymbol{\tau_b} - \boldsymbol{\tau_a})$
associated to the mid-point $m$ (see \Cref{sec_suppl:ablation_lambda_m}). 
We find that a simple strategy that results in an effective reduction of the influence of the outliers is to: \textit{(i)} detect
$\omega_{b\rightarrow a}$ terms that are outliers
when
$\lambda_m=0.5$,
evaluated as
$|\log(\omega_{b\rightarrow a})| > \log(1+\epsilon_{\mathrm{out}})$,
where $\epsilon_{\mathrm{out}}$ is a hyperparameter (for instance $\epsilon = 0.1$,
corresponds to a depth variation larger than $10\%$ between $z_a$ and $z_b$, \cf \eqref{eq:ours_formulation});
\textit{(ii)} uniformly sample multiple values of $\lambda_m\in[0, 1]$ for these pixels and select the value of $\lambda_m$ that yields the $\omega_{b\rightarrow a}$ term closest to $1$.
As shown in the last row of \cref{fig:ablation_noise_outliers} and \cref{fig:ablation_noise_rotation},
applying this strategy (here with $\epsilon_{\mathrm{out}} = 0.01$) results in a significant reduction of the spike artifacts, with complete removal of the artifacts in the case of rotational noise (\cref{fig:ablation_noise_rotation}).

\begin{table*}[!ht]
    \centering
    \resizebox{\linewidth}{!}{
    \begin{tabular}{lccccccccc}
    \toprule
    Method & \texttt{bear} & \texttt{buddha} & \texttt{cat} & \texttt{cow} & \texttt{harvest} & \texttt{pot1} & \texttt{pot2} & \texttt{reading} & \texttt{goblet}\\
    \midrule
    BiNI~\cite{Cao2022BiNI} & $(2.37 \pm 3.15) \times 10^{-5}$ & $(3.18 \pm 8.12) \times 10^{-5}$ & $(0.35 \pm 2.28) \times 10^{-4}$ & $(2.65 \pm 4.32) \times 10^{-5}$ & $(0.38 \pm 1.86) \times 10^{-4}$ & $(2.89 \pm 6.75) \times 10^{-5}$ & $(2.59 \pm 4.07) \times 10^{-5}$ & $(0.36 \pm 1.03) \times 10^{-4}$ & $(0.32 \pm 1.01) \times 10^{-4}$\\
    Ours & $(0.08 \pm 1.25) \times 10^{-5}$ & $(0.09 \pm 1.47) \times 10^{-4}$ & $(0.04 \pm 2.48) \times 10^{-4}$ & $(0.18 \pm 2.64) \times 10^{-5}$ & $(0.18 \pm 1.77) \times 10^{-4}$ & $(0.09 \pm 6.52) \times 10^{-4}$ & $(0.33 \pm 3.03) \times 10^{-5}$ & $(0.78 \pm 8.88) \times 10^{-5}$ & $(0.61 \pm 9.10) \times 10^{-5}$\\
    \bottomrule
    \end{tabular}
    }
    \caption{\textbf{Relative formulation accuracy on the ground-truth log-depth map, DiLiGenT dataset~\cite{Shi2016DiLiGenT}}. For both methods, we report mean and standard deviation across the pixels of the relative residual $|(\tilde{z}_a - \tilde{z}_b - \mathrm{RHS}\ /\ \gamma_{b\rightarrow a})\ /\ \tilde{z}_a|$ computed on the ground-truth log-depth map, where $\mathrm{RHS}$ denotes the right-hand side of \eqref{eq:bini_formulation} for BiNI and \eqref{eq:ours_formulation_log_with_bini_factor} for Ours.  We use $\boldsymbol{\tau_m} = (\boldsymbol{\tau_a} + \boldsymbol{\tau_b}) / 2$ and $\alpha_{b\rightarrow a}=0$ for Ours.}
    \label{tab:comparison_formulation_accuracy_ver3}
\end{table*}
\begin{table*}[!ht]
    \centering
    \resizebox{\linewidth}{!}{
    \begin{tabular}{lccccccccc}
    \toprule
    Method & \texttt{bear} & \texttt{buddha} & \texttt{cat} & \texttt{cow} & \texttt{harvest} & \texttt{pot1} & \texttt{pot2} & \texttt{reading} & \texttt{goblet}\\
    \midrule
    BiNI~\cite{Cao2022BiNI} & $(2.46 \pm 2.39) \times 10^{-4}$ & $(2.45 \pm 5.39) \times 10^{-4}$ & $(3.30 \pm 6.30) \times 10^{-4}$ & $(2.35 \pm 2.89) \times 10^{-4}$ & $(0.29 \pm 1.33) \times 10^{-3}$ & $(2.17 \pm 4.46) \times 10^{-4}$ & $(1.89 \pm 3.01) \times 10^{-4}$ & $(3.59 \pm 7.21) \times 10^{-4}$ & $(2.37 \pm 7.45) \times 10^{-4}$\\
    Ours &
    $(0.60 \pm 9.13) \times 10^{-5}$ & $(0.06 \pm 1.10) \times 10^{-3}$ & $(0.03 \pm 1.87) \times 10^{-3}$ & $(0.13 \pm 1.94) \times 10^{-4}$ & $(0.13 \pm 1.30) \times 10^{-3}$ & $(0.07 \pm 8.46) \times 10^{-3}$ & $(0.25 \pm 2.22) \times 10^{-4}$ & $(0.57 \pm 6.51) \times 10^{-4}$ & $(0.45 \pm 6.66) \times 10^{-4}$\\
    \bottomrule
    \end{tabular}
    }
    \caption{\textbf{Relative formulation accuracy on the ground-truth depth map, DiLiGenT dataset~\cite{Shi2016DiLiGenT}}. For both methods, we report mean and standard deviation across the pixels of the relative residual $\left|\left(z_a - \exp\left(\mathrm{RHS}\ /\ \gamma_{b\rightarrow a}\right)\cdot z_b\right)\ /\ z_a\right|$ computed on the ground-truth depth map, where $\mathrm{RHS}$ denotes the right-hand side of \eqref{eq:bini_formulation} for BiNI and \eqref{eq:ours_formulation_log_with_bini_factor} for Ours.  We use $\boldsymbol{\tau_m} = (\boldsymbol{\tau_a} + \boldsymbol{\tau_b}) / 2$ and $\alpha_{b\rightarrow a}=0$ for Ours.}
    \label{tab:comparison_formulation_accuracy_ver5}
\end{table*}
\section{Additional evaluations~\label{sec_suppl:additional_evaluations}}
\begin{table*}[!t]
    \makebox[\linewidth]{%
    \resizebox{0.55\linewidth}{!}{%
    \centering
    \begin{tabular}{lc@{\hspace{6pt}}cc@{\hspace{6pt}}cc@{\hspace{6pt}}cc@{\hspace{6pt}}cc@{\hspace{6pt}}c}
    \toprule
    \multirow{2}{*}{Method} & \multicolumn{2}{c}{\texttt{bear}} & \multicolumn{2}{c}{\texttt{buddha}} & \multicolumn{2}{c}{\texttt{cow}} & \multicolumn{2}{c}{\texttt{pot2}} & \multicolumn{2}{c}{\texttt{reading}}\\
    \cmidrule(lr){2-3} \cmidrule(lr){4-5} \cmidrule(lr){6-7} \cmidrule(lr){8-9} \cmidrule(lr){10-11}
    & GT & PS & GT & PS & GT & PS & GT & PS & GT & PS\\
    \midrule
    BiNI~\cite{Cao2022BiNI} & $0.30$ & $0.45$ & $2.33$ & $1.14$ & $0.26$ & $0.29$ & $0.72$ & $0.90$ & $0.89$ & $1.30$ \\
    Ours w/o $\alpha_{b\rightarrow a}$ & $\mathbf{0.24}$ & $0.45$ & $1.89$ & $1.04$ & $0.23$ & $0.29$ & $0.73$ & $\mathbf{0.83}$ & $0.86$ & $\mathbf{1.14}$ \\
    Ours & $\mathbf{0.24}$ & $\mathbf{0.44}$ & $\mathbf{1.64}$ & $\mathbf{1.02}$ & $\mathbf{0.21}$ & $\mathbf{0.28}$ & $\mathbf{0.66}$ & $\mathbf{0.83}$ & $\mathbf{0.80}$ & $1.24$ \\
    \bottomrule
    \end{tabular}
    }%
    }
    \vspace{-5pt}
    \caption{\textbf{Mean absolute depth error (MADE) [$\boldsymbol{\si{mm}}$] on the DiLiGenT-MV dataset~\cite{Li2020DiLiGenT_MV}, averaged across the $20$ object views.} $\textrm{GT}$: ground-truth normals, $\textrm{PS}$: normals from
    photometric stereo.
    All
    tests
    use
    $\num{1200}$ iterations.
    }
    \label{tab:diligent_mv_results}
\end{table*}
In this Section, we provide additional evaluations of our method and of the baseline of BiNI~\cite{Cao2022BiNI} on the DiLiGenT-MV dataset~\cite{Shi2016DiLiGenT}, which extends the DiLiGenT dataset for a subset of $5$ of its objects (\texttt{bear}, \texttt{buddha}, \texttt{cow}, \texttt{pot2}, \texttt{reading}) by rendering a total of $20$ views per object.
The
dataset contains both ground-truth normals and normals from photometric stereo, which
therefore
allows us to
quantitatively evaluate
the methods also on real normal maps.
We run all methods with the same settings as the main experiments, using $\num{1200}$ iterations, and
apply the outlier filtering strategy described in \cref{sec_suppl:results_for_noisy_inputs} for our method, setting $\epsilon_{\mathrm{out}} = 0.1$.

\Cref{tab:diligent_mv_results}
reports the 
mean absolute
error
(averaged across the $20$ object
views)
against
ground-truth
depth,
which
we render with BlenderProc~\cite{Denninger2023BlenderProc} using
ground-truth meshes and camera parameters.
The results confirm that our
method
performs better
than BiNI also
on
normals
from
photometric stereo,
with
discontinuity estimation
further
increasing
our accuracy.

\section{Limitations~\label{sec_suppl:limitations}}
\textbf{Requirement for physically meaningful normals.}
While effective strategies
for the mitigation of outliers
can be
designed,
as
described
in
\cref{sec_suppl:results_for_noisy_inputs},
our method requires 
that the input normals are physically meaningful, \ie, $\boldsymbol{n_a}^\mathsf{T} \boldsymbol{\tau_a} < 0$. 
As a consequence, an additional preprocessing step on the input normals (\cf \cref{sec_suppl:results_for_noisy_inputs} for example strategies) is required in the presence of outliers, to ensure that the above condition is fulfilled.
\\
\textbf{Non-central camera models.}
Since it
is based on
ray direction vectors, our
formulation
does not allow handling camera models
that are non-central, \ie, that do not assume all camera rays to originate from a single point (such as axial cameras~\cite{Ramalingam2006TheoryCalibrationAxialCameras}).
A particular case of non-central cameras are orthographic cameras, which assume the center of projection to be at an infinite distance from the scene.
As a consequence,
in this model
all ray direction vectors are parallel to each other and perpendicular to the image plane, \ie, $\boldsymbol{\tau_a}=\boldsymbol{\tau_b}=\boldsymbol{\tau_m}=(0, 0, 1)^\mathsf{T}$ for all $a, b, m$. We note that in this case our formulation~\eqref{eq:ours_formulation} reduces to $z_a = \varepsilon_{b\rightarrow a} + z_b$, which, while
correct, does not depend on the surface normals and is
thus
not applicable to normal integration.
\\
\textbf{Run time and input size.} Similarly to previous optimization-based approaches~\cite{Cao2022BiNI, Kim2024DiscontinuityPreserving, Queau2018NormalIntegrationSurvey}, our method is not compatible with real-time deployment, with optimization converging in a time frame in the order of several seconds ($\num{50}$ to $\num{120}$ seconds for input normal maps of size $\num{512}\times\num{612}$).
Additionally,
like for previous approaches,
our system matrix $\mathbf{A}$ (\cf. \eqref{eq:system_equations_ours} in the main paper), albeit sparse, has both
a
number of rows and
a number of
columns
that
scale linearly
with
the number of valid pixels in the input normal map.
This leads to larger processing time and memory usage for
large input
sizes, making it currently unsuitable for high-resolution maps and highly complex scenes.
More optimized implementations
could reduce runtime and memory
usage.
Investigating
more substantial modifications that could move away
completely
from the drawbacks of optimization-based integration is an interesting direction, but falls outside the scope of this study.
\\
\textbf{Hyperparameters.} Our method depends on a number of hyperparameters, namely the parameters $q$ and $\rho$ of our discontinuity activation term $\beta^{(t)}_{b\rightarrow a}$ (\cf~\eqref{eq:beta_definition_ours} in the main paper), the parameter $k$ controlling the sharpness of the bilateral weights $w_{b\rightarrow a}^\textrm{BiNI}$ (\cf.~\eqref{eq:bini_weight_definition} in the main paper), and the ray directions $\boldsymbol{\tau_m}$ that control our planarity assumption (\cf \cref{sec:method_general_formulation} in the main paper). While the default choices $k=2$ and $\boldsymbol{\tau_m} = (\boldsymbol{\tau_a} + \boldsymbol{\tau_b}) / 2$ consistently result in optimal results (\cf \cref{tab:analysis_gamma_a_from_b} and \cref{sec_suppl:ablation_lambda_m}), a certain degree of object specificity can be observed in $\beta^{(t)}_{b\rightarrow a}$, particularly in its parameter $q$ (\cf \cref{sec_suppl:ablation_beta_a_from_b_discont_activation_term}). Therefore, tuning 
the latter parameter
might be desirable to achieve slight improvements in performance.

\end{document}